\documentclass{elsarticle}

\usepackage{lineno,hyperref}
\modulolinenumbers[0]

\journal{arXiv}

\usepackage[a4paper]{geometry}

\usepackage{textcomp}

\newif\iftodo
\todotrue  

\newif\ifcomment
\commenttrue  

\usepackage{times,url,graphicx,latexsym,xspace,%
eurosym,amssymb,amsmath,helvet,courier}

\usepackage[ruled]{algorithm}
\usepackage[noend]{algorithmic}
\usepackage{amsthm}

\usepackage{thmtools, thm-restate}

\renewcommand{\algorithmicrequire}{\textbf{Input:}}
\renewcommand{\algorithmicensure}{\textbf{Output:}}

\renewcommand{\geq}{\geqslant}
\renewcommand{\leq}{\leqslant}

\newcommand{\AU}{\ensuremath{\mathbf{U}}\xspace}

\newcommand{\AB}{\ensuremath{\mathbf{B}}\xspace}
\newcommand{\AL}{\ensuremath{\mathbf{L}}\xspace}

\newcommand{\AUBL}{\ensuremath{\mathbf{UBL}}\xspace}

\newcommand{\AUL}{\ensuremath{\mathbf{UL}}\xspace}

\newcommand{\K}{\ensuremath{\mathcal{K}}\xspace}

\newcommand{\I}{\ensuremath{\mathcal{I}}\xspace}      
\newcommand{\J}{\ensuremath{\mathcal{J}}\xspace}   
      
\newcommand{\R}{\ensuremath{\mathcal{R}}}

\newcommand{\nd}{\noindent}

\newcommand{\tuple}[1]{\langle #1 \rangle }
\newcommand{\triple}[1]{(#1)}
\newcommand{\entail}{\models}

\newcommand{\andc}{\sqcap}

\newcommand{\assign}{\mbox{$\colon\!\!\!\!=$}}

\renewcommand{\int}[1]{{#1}^{\I} }
\newcommand{\intG}[1]{{#1}^{\I_G} }
\newcommand{\intGs}[1]{{#1}^{\I_{G^s}} }

\newcommand{\intP}[1]{{\mathfrak{P}[\![#1]\!]}}
\newcommand{\intC}[1]{{\mathfrak{C}[\![#1]\!]}}

\newcommand{\class}{\ensuremath{\mathsf{class}}}
\newcommand{\spp}{\ensuremath{\mathsf{sp}}}
\newcommand{\subclass}{\ensuremath{\mathsf{sc}}}
\newcommand{\type}{\ensuremath{\mathsf{type}}}
\newcommand{\dom}{\ensuremath{\mathsf{dom}}}
\newcommand{\range}{\ensuremath{\mathsf{range}}}
\newcommand{\eee}{\ensuremath{\mathsf{e}}}

\newcommand{\ii}[1]{\mbox{$(#1)$}}

\newcommand{\subs}{\sqsubseteq}
\newcommand{\dsubs}{\:\raisebox{0.45ex}{\ensuremath{\sqsubset}}\hskip-1.7ex\raisebox{-0.6ex}{\scalebox{0.9}{\ensuremath{\sim}}}\:}

 \newtheorem{theorem}{Theorem}[section]
 \newtheorem{proposition}[theorem]{Proposition}
 \newtheorem{corollary}[theorem]{Corollary}
 \newtheorem{lemma}[theorem]{Lemma}

 \newtheorem{example}{Example}[section]
 \newtheorem{definition}{Definition}[section]
\newtheorem{remark}{Remark}[section]

\newcommand{\ie}{\textit{i.e.},\xspace}
\newcommand{\eg}{\textit{e.g.},\xspace}
\newcommand{\wrt}{w.r.t.\xspace}
\newcommand{\D}{{\mathtt D}}

\renewcommand{\iff}{if and only if\xspace}
\newcommand{\st}{such that\xspace}

\newcommand{\rhodf}{\mbox{$\rho$df}\xspace}
\newcommand{\universe}{\ensuremath{\mathtt{uni}}\xspace}

\def\qed{\hspace*{\fill} \ensuremath{\Box}}

\graphicspath{{../report/images/}}

\usepackage{bussproofs}
\usepackage{stmaryrd}
\newcommand{\subc}{\mathtt{sc}}
\newcommand{\subp}{\mathtt{sp}}
\newcommand{\Dom}{\mathtt{dom}}

\newcommand{\fnl}{\mbox{\\ }}

\newcommand{\disjC}{\ensuremath{\bot_{\sf c}}}

\newcommand{\ssc}{{\sf c}}
\newcommand{\ssp}{{\sf p}}

\newcommand{\disjP}{\ensuremath{\bot_{\sf p}}}

\newcommand{\cmin}{{\sf c}\_\min}
\newcommand{\pmin}{{\sf p}\_\min}

\newcommand{\dtriple}[1]{\langle#1\rangle}
\newcommand{\anytriple}[1]{\llbracket#1\rrbracket}

\newcommand{\fcond}{\rightsquigarrow}

\let\Oldfootnote\footnote
\renewcommand{\footnote}[1]{\Oldfootnote{#1}}

\usepackage{times}
\usepackage{alltt}
\usepackage{algorithmic}
\usepackage{epsfig}
\usepackage{url}
\usepackage{xspace}
\usepackage{latexsym}
\usepackage{rotating} 
\usepackage{subfigure}
\usepackage{fancyvrb}
\usepackage{paralist}
\usepackage{multirow}
\usepackage{amsmath}
\usepackage{mathrsfs}
\usepackage{tikz}
\usetikzlibrary{calc}
\usepackage{algorithm}

\usepackage{hyperref}
\usepackage{url}

\RecustomVerbatimEnvironment{Verbatim}{Verbatim}{fontsize=\small}

\newcommand{\deriv}{\vdash_{\rhodfbot}}

\newcommand{\rdfent}{\vDash_{\rho df_\bot}}
\newcommand{\rdfsat}{\Vdash_{\rho df_\bot}}

\newcommand{\Gclass}{G^{str}}
\newcommand{\Gdef}{G^{def}}

\newcommand{\Gcount}{G^{s}}

\newcommand{\clos}{\ensuremath{{\sf Cl}}}
\newcommand{\rhodfbot}{\ensuremath{\rho df_\bot}}
\newcommand{\closmin}{\ensuremath{{\sf Cl_{\min}}}}

\newcommand{\M}{\mathscr{M}}
\newcommand{\IG}{\mathfrak{I}_G}
\newcommand{\RG}{\mathfrak{R}_G}
\newcommand{\MN}{\M_{\mathbb{N}}}
\newcommand{\CG}{\ensuremath{\R_{\min G}}}

\newcommand{\DR}{\ensuremath{{\sf R}}}
\newcommand{\DP}{\ensuremath{{\sf P}}}
\newcommand{\DC}{\ensuremath{{\sf C}}}
\newcommand{\DL}{\ensuremath{{\sf L}}}

\newcommand{\minentail}{\ensuremath{\entail_{\min}}}
\newcommand{\rnk}{\ensuremath{{\mathtt r}}}
\newcommand{\drnkc}{\ensuremath{{\mathtt D}}}
\newcommand{\drnkp}{\ensuremath{{\mathtt D}}}
\newcommand{\drnk}{\ensuremath{{\mathtt D}}}

\newcommand{\hc}{\ensuremath{h^\ssc_{G}}}
\newcommand{\hp}{\ensuremath{h^\ssp_{G}}}

\newcommand{\boxd}{\Box\hspace*{-1.8mm}\diamond}

\newcommand{\cf}{\textit{cf.}\xspace}

\newcommand{\closinh}{\ensuremath{{\sf Cl}_{in}}}

\begin{document}

\begin{frontmatter}

\title{Defeasible RDFS via Rational Closure}


\author[mymainaddress,mysecondaryaddress]{Giovanni Casini\corref{mycorrespondingauthor}}
\cortext[mycorrespondingauthor]{Corresponding author}
\ead{giovanni.casini@isti.cnr.it}

\author[mymainaddress]{Umberto Straccia
}
\ead{umberto.straccia@isti.cnr.it}

\address[mymainaddress]{CNR - ISTI, Pisa, Italy}
\address[mysecondaryaddress]{CAIR - University of Cape Town, Cape Town, South Africa}

\begin{abstract}
In the field of non-monotonic logics, the notion of Rational Closure (RC) is acknowledged as a notable approach.
In recent years, RC has gained popularity  in the context of Description Logics (DLs), the logic underpinning the standard semantic Web Ontology Language OWL 2, whose main ingredients are classes, the relationship among classes and roles, which are used to describe the properties of classes.\\
\indent In this work, we show instead  how to integrate RC within the triple language RDFS (Resource Description Framework Schema), which together with OWL 2 is a major  standard semantic web ontology language. 

To  do so, we start from \rhodf, a minimal, but significant RDFS fragment that covers the essential features of RDFS,
and then extend it to \rhodfbot, allowing to state that two entities are incompatible/disjoint with each other. Eventually, we propose defeasible \rhodfbot~via a typical RC construction allowing to state default class/property inclusions. \\
\indent Furthermore, to overcome the main weaknesses of RC in our context, \ie~the ``drowning problem'' (\textit{viz.}~the ``inheritance blocking problem''), we further extend our construction by leveraging Defeasible Inheritance Networks (DIN) defining a new non-monotonic inference relation that combines the advantages of both (RC and DIN). To the best of our knowledge this is the first time of such an attempt.\\
\indent In summary, the main features of our approach are: 
(i) the defeasible \rhodfbot~we propose here remains syntactically a triple language by extending it with new predicate symbols with specific semantics; (ii) the logic is defined in such a way that any RDFS reasoner/store may handle the new predicates as  ordinary terms if it does not want to take account of the extra non-monotonic capabilities;
(iii) the defeasible entailment decision procedure is built on top of the \rhodfbot~entailment decision procedure, which in turn is an extension of the one for \rhodf~via some additional inference rules favouring a potential implementation; 
(iv) the computational complexity of deciding entailment in \rhodf~and \rhodfbot~are the same; and 
(v) defeasible entailment can be decided via a polynomial number of calls to an oracle deciding ground triple entailment in \rhodfbot~and, in particular, deciding defeasible entailment can be done in polynomial time.



 \end{abstract}


\end{frontmatter}

%

\section{Introduction}


 \nd \emph{Description Logics} (DLs) \cite{BaaderEtAl2007}  under \emph{Rational Closure} (RC) \cite{Lehmann92b}  is a well-known framework for non-monotonic reasoning in DLs, which has gained rising attention in the last decade~\cite{Bonatti2019,BritzEtAl2015a,Casini10,Casini12,CasiniStraccia13,CasiniEtAl2014,CasiniEtAl19,Giordano16a,Giordano14,Giordano15,Pensel18,Pozzato19}.


We recall that a typical problem that can be addressed using non-monotonic formalisms is reasoning with ontologies in which some classes are exceptional \wrt~some properties of their super classes, as illustrated with the following example.
\vspace*{1em}
\begin{example}\label{ex01} \mbox{ \ }\\
Consider the following facts.\\\\
\nd\begin{tabular}{ll}
(1) & Young people are usually happy. \\
(2) & Young people are usually students. \\
(3) & Drug users are usually unhappy. \\
(4) & Drug users use  drugs. \\
(5) & Drugs are chemical substances. \\
(6) & Drug users are usually young. \\
(7) & Controlled drug users are usually happy.\\\\
\end{tabular} \mbox{ \ }\\ 
\nd We may consider then reasonable to conclude:\\\\
\nd \begin{tabular}{ll}
(8) &Young drug users are usually unhappy. \\
(9) & Controlled young drug users are usually happy.\\
(10) & Drug users are usually students.\\
\end{tabular}
\qed
\end{example}

\nd While DLs provide the logical foundation of formal ontologies of the semantic \emph{Web  Ontology Language} (OWL) family\footnote{\url{https://www.w3.org/TR/owl2-profiles/}} and endowing them with non-monotonic features is still a main issue, as documented by the past 20~years of technical development~\cite{BonattiEtAl2015,Eiter08,Eiter11,Giordano13,Motik10}, addressing non-monotonicity  in the context of  the triple language RDFS (\emph{Resource Description Framework Schema}),\footnote{\url{http://www.w3.org/TR/rdf-schema/}}
~which together with OWL 2 is a major  standard semantic web ontology language, has attracted in comparison little attention so far. Moreover, almost all approaches we are aware of consider a so-called rule-layer on top of RDFS; see \eg~\cite{Analyti15,Antoniou02,Eiter08,Ianni09} and Section~\ref{relw}.

In this paper, we will show instead how to integrate RC directly within the triple language RDFS. To  do so, we start from \rhodf~\cite{Gutierrez11,Munoz09}, a minimal, but significant RDFS fragment that covers the essential features of RDFS, and then extend it to \rhodfbot, allowing to state that two entities are incompatible/disjoint with each other. So, for instance, by referring to Example~\ref{ex01}, we may represent that happy beings ($h$) and unhappy beings ($u$) are incompatible with each other via the  \rhodfbot~triple $\triple{h, \disjC, u}$ (see also Example~\ref{pex} later on), where $\disjC$ is a new predicate added to the \rhodf~vocabulary to model an incompatibility relation. Roughly, you may think of being a triple $\triple{c, \disjC, d}$ as a \rhodfbot~counterpart of a disjointness axiom in OWL (\emph{viz.}~DLs)  of the form $\mathtt{DisjointClasses}(C,D)$ (in DL syntax, $C \andc D \subs \bot$).

\begin{remark}\label{unhappy}
Note that, I'm `unhappy' tells exactly what you are, while I'm `not happy' only tells what you are not, and, thus, I'm `unhappy' does not mean  the same as I'm `not happy'. Nevertheless, both are incompatible with the state of being `happy'.
Of course, the incompatibility/disjointness relation is weaker than the negation/complementing relation in the sense that the latter implies the former, but not vice-versa.
%
We introduce the notion of incompatibility/disjointness as the `weakest' relation we are aware of to address  defeasible non-monotonic reasoning, that requires a notion of conflict between pieces of information.\footnote{Informally, defeasible rules allow for the presence of exceptional cases, that have properties in conflict with respect to the typical situation (\eg in Example \ref{ex01} we say that drug users are usually unhappy (3), but controlled drug users, despite being drug users, are usually happy (7)).} This, however, does not prevent our approach to be extended with stronger notions as well, such as involving forms of `negation' (see,~\eg~\cite{Analyti15,Antoniou02,Eiter08,Ianni09}). However, these have been come so far at the price of an increase of the computational complexity (see Section~\ref{relw}). We will not address such extensions here and leave them for future work. We also refer the reader to~\cite{CasiniEtAl19} for a similar argument related to the use of $\bot$ to express disjointness among DL classes to implement a computationally tractable RC for the DL $\mathcal{ELO}_\bot$.




\end{remark}

\nd Now, based on \rhodfbot, we will then propose  defeasible \rhodfbot~via a typical RC construction, allowing to state, \eg``young people ($y$) are usually happy" via the defeasible triple $\dtriple{y,\subclass,h}$, alongside  classical triples such as  $\triple{du, use, d}$ (``drug users ($du$) use drugs ($d$)")  and $\triple{d,\subclass,c}$ (``drugs are chemical substances ($c$)").\footnote{We recall that according to \rhodf, $\subclass$ stands for ``is subclass of".} 
%
%

However, despite RC having nice behavioural properties~\cite{Lehmann92b}, the main weaknesses of RC in our context is the so-called  ``drowning problem" (also known as the "inheritance blocking problem"), which essentially states that if some defeasible property of a class $C$ contradicts any defeasible property of a superclass $D$ of $C$, then $C$ inherits none of the defeasible properties of $D$, including those that are consistent with the properties of $C$. So, for example, again by referring to  Example~\ref{ex01}, since drug users are atypical young people (they are unhappy, while young people are usually happy), RC wouldn't allow us to associate with drug users any typical property of young people, for example, to conclude  $\dtriple{du,\subclass,s}$ (``drug users ($du$) are usually students ($s$)"). To overcome the ``drowning problem", we further extend our construction by leveraging \emph{Defeasible Inheritance Networks} (DIN)~ \cite{Horty94} defining a new non-monotonic inference relation  combining the advantages of both (RC and DIN)  and, thus, \eg~allowing us eventually to conclude ($10$) in Example~\ref{ex01}. 

In summary, the main features of our approach are: 

\begin{itemize}
\item  
defeasible \rhodfbot~remains syntactically a triple language and is a simple  extension of the \rhodf~vocabulary by introducing some new predicate symbols with specific semantics;

\item any RDFS reasoner/store and SPARQL\footnote{\url{https://www.w3.org/TR/sparql11-query/}}  query answering tool may handle the new types of triples as  ordinary ones if it does not want to take account of the extra non-monotonic capabilities; 

\item the defeasible entailment  decision procedure is built on top of the \rhodfbot~entailment decision procedure, which in turn is an extension of the one for \rhodf~via some additional inference rules favouring a potential implementation; and

\item the computational complexity of deciding entailment in \rhodf~and \rhodfbot~are the same; and


\item defeasible entailment can be decided via a polynomial number of calls to an oracle deciding ground triple entailment in \rhodfbot. In particular, deciding defeasible entailment can be done in polynomial time.

\end{itemize}

\nd In the following, we will proceed as follows. In the next section, we will introduce $\rhodfbot$, by defining its syntax, semantics and entailment decision procedure. In Section~\ref{defrdfs} we extend \rhodfbot~towards defeasible \rhodfbot~via an RC construction, defining syntax, semantics, entailment decision procedure and address its computational complexity.  In Section~\ref{din}, we encapsulate DINs into our framework, define the entailment decision procedure and address its computational complexity. 
All the previous sections address the case of ground graphs only. The non-ground case is addressed in Section~\ref{blanknodes}. Eventually, Section~\ref{relw} addresses related work and Section~\ref{concl} concludes with a brief summary of our contribution and highlights future work. Some poofs of lemmas and theorems are in the appendixes.

\section{$\rhodfbot$ Graphs}
\label{sec:preliminaries}

\nd RC is a popular non-monotonic approach in conditional reasoning: 
given a set of defeasible conditionals $\alpha \fcond\beta$ (if $\alpha$ holds, then presumably $\beta$ holds), we use them like classical monotonic conditionals until we have a conflict in our information, that  triggers the non-monotonic reasoning machinery. Example \ref{ex01} shows  such a case: we have a conflict between classical reasoning, suggesting that drug users, being young, should be happy, and an exceptional more specific information, stating that drug users are unhappy. Faced with such a conflict, the non-monotonic engine solves it giving, in this case, precedence to the more specific information (``drug users are unhappy"). 
%
%
However, in order to implement this kind of defeasible reasoning in the RDFS framework, we need to introduce some notion of informational incompatibility, such as ``someone can not be happy and unhappy at the same time". Such incompatibilities cannot be represented in $\rhodf$: to do so, we extend the $\rhodf$ language to $\rhodfbot$  allowing to express such forms of conflicts (but, see also Remark~\ref{unhappy}). 
We are going to briefly introduce the $\rhodf$ formalism, in order to introduce the $\rhodfbot$ language. We are going to characterise $\rhodfbot$ by appropriately extending  the $\rhodf$ semantics and the $\rhodf$ deductive system.

\subsection{Syntax} \label{sec:rdf-syntax} 

\nd To start with, we rely on a fragment of RDFS, called \emph{minimal} $\rhodf$~\cite[Def.~15]{Munoz09}, that covers all main features of RDFS, is essentially the formal logic behind RDFS and, in fact, suffices to illustrate the main concepts and algorithms we will consider in this work and ease the presentation. 
Specifically, minimal $\rhodf$ has the following \emph{vocabulary} (reserved keywords/predicates) whose meaning will be clear within few paragraphs: 
\begin{equation} \label{vocrdf}
\rhodf = \{ \spp, \subclass, \type, \dom,\range\} \ .
\end{equation}
\nd Moreover, we recall that minimal $\rhodf$ does not consider so-called \emph{blank} nodes (see, \eg~\cite{Hogan14,Munoz09} for more on this theme) and, thus, in what follows, triples and graphs will be \emph{ground}. We will come back to deal with non-ground graphs in Section~\ref{blanknodes} later on. Also, to avoid unnecessary redundancy, we will just drop the term `minimal' in what follows.

So, consider pairwise disjoint alphabets $\AU$ and $\AL$ denoting, 
respectively, \emph{URI references} and \emph{literals}.   With $\AUL$ we will denote the union of these sets. 
Moreover, a \emph{vocabulary} is a subset of $\AUL$. Specifically, we assume that $\AU$ contains the $\rhodf$ vocabulary (see Equation~\ref{vocrdf}).
A literal may be a \emph{plain literal}  (\eg~a string) or a \emph{typed literal} (\eg~a boolean value).\footnote{\url{http://www.w3.org/TR/rdf-primer/}} We call the elements in $\AUL$ \emph{terms}. Terms are denoted with lower case letters $a,b, \ldots$ with optional super/lower script.

A \emph{triple} is of the form $\tau=\triple{s,p,o} \in \AUL \times \AU \times \AUL$,\footnote{As in~\cite{Munoz09}, we allow literals for $s$.}
where $s,o \notin \rhodf$.  We call $s$ the \emph{subject}, $p$ the \emph{predicate}, and $o$ the \emph{object}. A \emph{graph}
$G$ is a set of triples, the \emph{universe} of $G$, denoted $\universe(G)$, is the set of terms in $\AUL$ that occur in the triples of $G$.  

We recall that informally \ii{i} $\triple{p, \spp, q}$ means that property $p$ is a \emph{subproperty} of property $q$;
\ii{ii} $\triple{c, \subclass, d}$ means that class $c$ is a \emph{subclass} of class $d$; \ii{iii} $\triple{a, \type,
  b}$ means that $a$ is of \emph{type} $b$; \ii{iv} $\triple{p, \dom, c}$ means that the \emph{domain} of property $p$
is $c$; and \ii{v} $\triple{p, \range, c}$ means that the \emph{range} of property $p$ is $c$. 

\begin{figure}
\begin{center}
\includegraphics[angle=90,origin=c,scale=0.5]{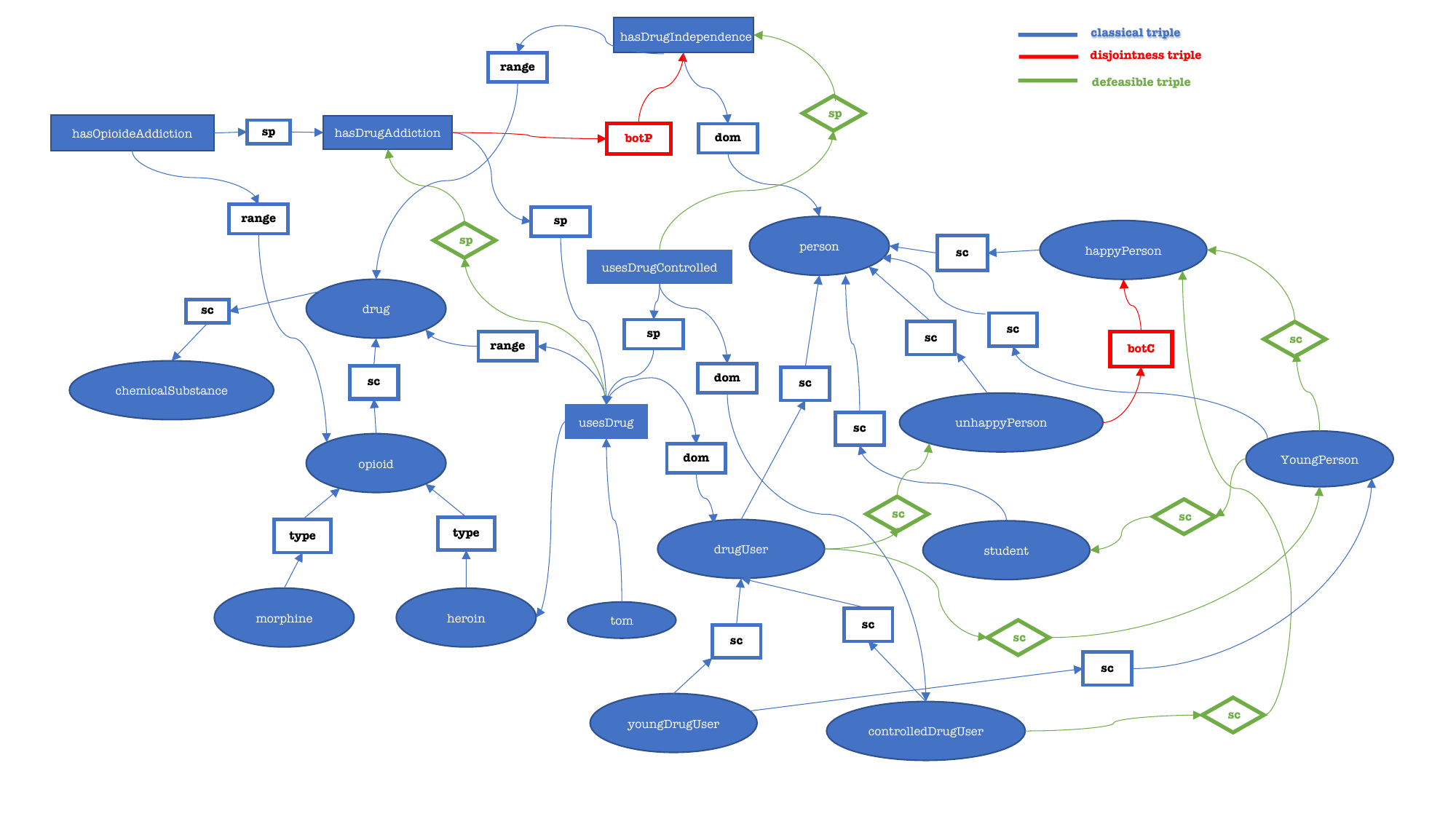}
\end{center}
\caption{The drug ontology graph. Rectangular/diamond nodes are properties/relations. Vocabulary properties are in bold. A triple has the pattern $\star \rightarrow \boxd \rightarrow \bullet$, where $\bullet$ is an oval node, $\boxd$ a rectangular or diamond node, and $\star$ is either an oval or rectangular node. Neither $\star$ nor $\bullet$ can involve vocabulary properties. Blue triples are classical $\rhodf$ triples, red triples are disjointness triples, while green ones are defeasible.} \label{figrdfs}
\end{figure}

\begin{example}[Running example]\label{exrdfs}
The following graph is an excerpt of a drug ontology (see also Figure~\ref{figrdfs} for a pictorial representation, neglecting for the moment the non-blue elements ).
{\small
\begin{eqnarray*}
G & =  \{  & 
\triple{\mathit{morphine}, \type, \mathit{opioid}}, 
\triple{\mathit{heroin}, \type, \mathit{opioid}}, 
\triple{\mathit{opioid}, \subclass, \mathit{drug}}, \\  
&&\triple{\mathit{drug}, \subclass, \mathit{chemicalSubstance}},  
 \triple{\mathit{usesDrug}, \dom, \mathit{drugUser}}, 
 \triple{\mathit{usesDrug}, \range, \mathit{drug}}, \\
&& \triple{\mathit{tom}, \mathit{usesDrug}, \mathit{heroin}},
\triple{\mathit{hasDrugIndependence}, \dom, \mathit{person}}, \\
&& \triple{\mathit{hasDrugIndependence}, \range, \mathit{drug}},  
 \triple{\mathit{hasOpioidAddiction}, \spp, \mathit{hasDrugAddiction}}, \\ 
&& \triple{\mathit{hasDrugAddiction}, \spp, \mathit{usesDrug}}, 
  \triple{\mathit{happyPerson}, \subclass, \mathit{person}},  
  \triple{\mathit{unhappyPerson}, \subclass, \mathit{person}}, \\
&& \triple{\mathit{hasOpioidAddiction}, \range, \mathit{opioid}}, 
  \triple{\mathit{drugUser}, \subclass, \mathit{person}},  
\triple{\mathit{uses}, \dom, \mathit{person}}, \\
&&  \triple{\mathit{controlledDrugUser}, \subclass, \mathit{drugUser}},  
\triple{\mathit{student}, \subclass, \mathit{person}}, 
\triple{\mathit{youngPerson}, \subclass, \mathit{person}}, \\
&&  \triple{\mathit{youngDrugUser}, \subclass, \mathit{drugUser}},   
\triple{\mathit{youngDrugUser}, \subclass, \mathit{youngPerson}}, \\
&&  \triple{\mathit{usesDrugControlled}, \spp, \mathit{usesDrug}}, \triple{\mathit{usesDrugControlled}, \dom, \mathit{controlledDrugUser}} \}
\end{eqnarray*}
} \qed
\end{example}

\begin{remark} \label{remrdfs}
Let us recap that $\rho$df has been introduced in~\cite{Munoz09} to cover the salient features of RDFS from an inference point of view. In \cite[Table A.1]{Munoz09} 
the whole RDFS vocabulary is enumerated.\footnote{See also \url{http://www.w3.org/TR/rdf-schema/}} There you may find, \eg~the keyword {\bf rdfs:Class}, to indicate that a given term is a class.  Besides, in full RDFS one has also to consider many axioms such as $\triple{\subc, \dom, \class}$ indicating that ``the first argument of the the property $\subc$ is a class"  (see \cite[Table A.2]{Munoz09}).\footnote{Instead, we will have no axioms in our setting, see also~\cite[Corollary 14]{Munoz09}.} We believe that, as in~\cite{Munoz09}, such ingredients are of limited importance from a theoretical point of view. Extending our framework to whole RDFS is somewhat straightforward and tedious and, thus, we will not address it here.
\end{remark}

\nd We extend the vocabulary of $\rhodf$ with a new pair of predicates, $\disjC$ and $\disjP$, representing incompatible information: 
\begin{itemize}
    \item $\triple{c,\disjC, d}$ indicates that the classes $c$ and $d$ are disjoint; analogously, 
    \item $\triple{p,\disjP, q}$ indicates that the properties $p$ and $q$ are disjoint.
\end{itemize}
%
%
%
\nd We call $\rhodfbot$ the vocabulary obtained from $\rhodf$ by adding $\disjC$ and $\disjP$, that is, 
\begin{equation}\label{vocrhobot}
\rhodfbot = \{ \spp, \subc, \type, \dom, \range, \disjC, \disjP\} \ .
\end{equation}
\nd  Like for $\rhodf$, we assume that $\AU$ contains the $\rhodfbot$ vocabulary and that all triples $\triple{s,p,o} \in \AUL \times \AU \times \AUL$  are such that $s,o \notin \rhodfbot$. 


\begin{remark}\label{rem1}
Please, note that we allow the $\rhodfbot$ predicates to occur only as second elements of the triples, that is, we allow triples $\triple{p,\spp,q}$, but not triples such as \eg~$\triple{\spp,p,o}$ or $\triple{\disjP,p,o}$, which is in line with the notion of \emph{minimal  $\rhodf$ triple}~\cite[Def. 15]{Munoz09}. 

\end{remark}

\begin{example}[Example~\ref{exrdfs} cont.] \label{exrdfsB}
By referring to Example~\ref{exrdfs}, the graph 
\begin{eqnarray*}
  \Gclass & = G \ \cup \{ & \triple{\mathit{unhappyPerson},\disjC,\mathit{happyPerson}},\\ 
  && \triple{\mathit{hasDrugAddiction},\disjP,\mathit{hasDrugIndependence}} \ \ \}
\end{eqnarray*}
\nd  is a  $\rhodfbot$ graph. 


\qed
\end{example}





\subsection{Semantics} 
\nd Our semantics for $\rhodfbot$ extends the one for $\rhodf$~\cite{Munoz09} in the following way.

An \emph{interpretation} $\I$ over a vocabulary $V$ is a tuple $\I =\tuple{\Delta_{\DR}, \Delta_{\DP}, \Delta_{\DC},
  \Delta_{\DL}, \intP{\cdot}, \intC{\cdot}, \int{\cdot}}$, where $\Delta_{\DR}, \Delta_{\DP}$, $\Delta_{\DC}, \Delta_{\DL}$ are the interpretation domains of $\I$, which are finite non-empty sets, and $\intP{\cdot}, \intC{\cdot}, \int{\cdot}$ are
the interpretation functions of $\I$. In particular:
\begin{enumerate}
\item $\Delta_{\DR}$ are the resources (the domain or universe of $\I$);
\item $\Delta_{\DP}$ are property names (not necessarily disjoint from $\Delta_{\DR}$); 
\item $\Delta_{\DC} \subseteq \Delta_{\DR}$ are the classes; 
\item $\Delta_{\DL} \subseteq \Delta_{\DR}$ are the literal values and contains $\AL \cap V$;
\item  $\intP{\cdot}$ is a function $\intP{\cdot}\colon \Delta_{\DP} \to 2^{\Delta_{\DR} \times \Delta_{\DR}}$;
\item  $\intC{\cdot}$  is a function  $\intC{\cdot}\colon \Delta_{\DC} \to 2^{\Delta_{\DR}}$;
\item $\int{\cdot}$ maps each $t \in \AUL \cap V$ into a value $\int{t} \in \Delta_{\DR} \cup \Delta_{\DP}$, and such that
  $\int{\cdot}$ is the identity for plain literals and assigns an element in $\Delta_{\DR}$ to each element in $\AL$.

\end{enumerate}



\nd As next, we are going to build our semantics starting from the so-called~\emph{reflexive-relaxed} $\rhodf$ semantics~\cite{Munoz09}, in which the predicates $\subclass$ and $\spp$ are \emph{not} assumed to be reflexive in accordance to the notion of minimal graph~\cite[Def. 15]{Munoz09}. Our additional semantic constraints \wrt~\cite[Def.~15]{Munoz09} are those of condition \textbf{Disjointness I} onwards.\footnote{An interesting property of \emph{reflexive-relaxed} $\rhodf$ is that the proof system is axiom-free~\cite[Corollary 14]{Munoz09}.}

\begin{definition}[Satisfaction]\label{satisfaction}
An interpretation $\I$ is a \emph{model} of a  graph $G$, denoted $\I \rdfsat G$, \iff $\I$ is an interpretation over the vocabulary $\rhodfbot \cup \universe(G)$ that satisfies the following conditions:
%
\begin{description}\label{condRDF}
 \item[Simple:] \
 \begin{enumerate}
  \item for each $\triple{s, p, o} \in G$, $\int{p} \in\Delta_{\DP}$ and $(\int{s}, \int{o}) \in \intP{\int{p}}$;
 \end{enumerate}
 \item[Subproperty:] \
 \begin{enumerate}
 \item $\intP{\int{ \spp}}$ is transitive over $\Delta_{\DP}$;
 \item if $(p, q) \in \intP{\int{ \spp}}$ then $p, q \in \Delta_{\DP}$ and $\intP{p} \subseteq \intP{q}$; 
 \end{enumerate}
 \item[Subclass:] \
 \begin{enumerate}
 \item $\intP{\int{\subclass}}$ is transitive over $\Delta_{\DC}$; 
 \item if $(c, d) \in \intP{\int{\subclass}}$ then $c, d \in \Delta_{\DC}$ and $\intC{c} \subseteq \intC{d}$; 
 \end{enumerate}
 \item[Typing I:] \
 \begin{enumerate}
 \item $x \in \intC{c}$ \iff $(x,c) \in \intP{\int{\type}}$;
 \item if $(p, c) \in \intP{\int{\dom}}$ and $(x, y) \in \intP{p}$ then $x \in \intC{c}$;
 \item if $(p, c) \in \intP{\int{\range}}$ and $(x, y) \in \intP{p}$ then $y \in \intC{c}$;
 \end{enumerate}
 \item[Typing II:] \
 \begin{enumerate}
 \item For each $\eee \in \rhodfbot$, $\int{\eee} \in \Delta_{\DP}$;
 \item if $(p, c) \in \intP{\int{\dom}}$ then $p \in \Delta_{\DP}$ and $c \in \Delta_{\DC}$;
 \item if $(p, c) \in \intP{\int{\range}}$ then $p \in \Delta_{\DP}$ and $c \in \Delta_{\DC}$;
 \item if $(x, c) \in \intP{\int{\type}}$ then $c \in \Delta_{\DC}$;
 \end{enumerate}
 
\item[Disjointness I:] \   
  \begin{enumerate}
 \item if $(c, d) \in \intP{\int{\disjC}}$  then $c,d\in\Delta_{\DC}$;
 \item $\intP{\int{\disjC}}$ is symmetric, \emph{sc}-transitive and \textcolor{black}{\emph{c}-exhaustive} over $\Delta_{\DC}$ (see below);
  
 \item if $(p, q) \in \intP{\int{\disjP}}$  then $p,q\in\Delta_{\DP}$;
 \item $\intP{\int{\disjP}}$ is symmetric, \emph{sp}-transitive and  \textcolor{black}{\emph{p}-exhaustive}  over $\Delta_{\DP}$ (see below);
  \end{enumerate}

  \item[Disjointness II:] \   
  \begin{enumerate}
 \item If $(p, c) \in \intP{\int{\dom}}$, $(q, d) \in \intP{\int{\dom}}$, and $(c, d) \in \intP{\int{\disjC}}$, then $(p, q) \in \intP{\int{\disjP}}$;
 
  \item If $(p, c) \in \intP{\int{\range}}$, $(q, d) \in \intP{\int{\range}}$, and $(c, d) \in \intP{\int{\disjC}}$, then $(p, q) \in \intP{\int{\disjP}}$;
 
  \end{enumerate}
%

  

\item[Symmetry:] \
 \begin{enumerate}
\item If $(c,d)\in \intP{\int{\disjC}}$, then $(d,c)\in \intP{\int{\disjC}}$;
\item If $(p,q)\in \intP{\int{\disjP}}$, then $(q,p)\in \intP{\int{\disjP}}$;
\end{enumerate}
\item[sc-Transitivity:] \
 \begin{enumerate}
\item If $(c,d)\in \intP{\int{\disjC}}$ and $(e,c)\in\intP{\int{\subc}}$, then $(e,d)\in \intP{\int{\disjC}}$;
 \end{enumerate}
\item[sp-Transitivity:] \
\begin{enumerate}
\item If $(p,q)\in \intP{\int{\disjP}}$ and $(r,p)\in\intP{\int{\subp}}$, then $(r,q)\in \intP{\int{\disjP}}$ ;
\end{enumerate}
\item[\textcolor{black}{c-Exhaustive}:] \
 \begin{enumerate}
\item If $(c,c)\in \intP{\int{\disjC}}$ and $d\in\Delta_{\DC}$ then $(c,d)\in \intP{\int{\disjC}}$; 
 \end{enumerate}
\item[\textcolor{black}{p-Exhaustive}:] \
\begin{enumerate}
\item If $(p,p)\in \intP{\int{\disjP}}$ and $q\in\Delta_{\DP}$ then $(p,q)\in \intP{\int{\disjP}}$ \ .
\end{enumerate}

\end{description}

\nd A graph $G$ is \emph{satisfiable} if it has a model $\I$ (denoted $\I \rdfsat G$). 
\end{definition}


\begin{remark}
Note that the presence of \eg~$\triple{a,\type,b},\triple{a,\type,c}$ and $ \triple{b,\disjC,c}$ in a graph does not preclude its satisfiability.
In fact, a graph is always satisfiable (see Corollary~\ref{Coroll:sat_bot} later on) avoiding, thus, the possibility of unsatisfiability and the \emph{ex falso quodlibet} principle. This is in line with the $\rhodf$ semantics~\cite{Munoz09}.
\end{remark}

\nd On top of the notion of satisfaction we define a notion of entailment between graphs.





\begin{definition}[Entailment $\rdfent$]\label{entailment}
Given two graphs $G$ and $H$, we say that $G$ \emph{entails} $H$, denoted $G \rdfent H$, \iff every model of $G$ is also a model of $H$.\footnote{For ease of presentation, if a graph is a singleton, we will omit the curly braces.}
\end{definition}

%
%
%
\begin{example}[Example~\ref{exrdfsB} cont.] \label{exrdfsC}
Consider Example~\ref{exrdfsB}. It can be verified that, \eg~the following hold:
\begin{eqnarray*}
\Gclass & \models & \triple{\mathit{tom}, \type, \mathit{person}} \\
\Gclass & \models & \triple{\mathit{hasDrugIndependence}, \disjP, \mathit{hasOpioidAddiction}} \ .
\end{eqnarray*}
\qed
\end{example}

\subsection{Deductive System}\label{sect:rhodfbot_deductive}

\nd In what follows, we provide a sound and complete deductive system for our language. 
Our system extends the classical \emph{minimal} $\rhodf$ system as by~Mu\~noz and others \cite[Proposition 17]{Munoz09}. The rules for $\rhodf$ correspond to rules (1) - (4) below.
The system is arranged in groups of rules that capture the semantic conditions of models. In every rule, $A,B,C,D,E,X$, and $Y$ are meta-variables representing
elements in $\AUL$.  An instantiation of a rule is obtained by replacing those meta-variables with actual terms. The rules are as follows:


%
  \begin{enumerate}
  \item Simple: \\[0.5em]
    \begin{tabular}{llll}
  $\frac{G}{G'}$ for  $G' \subseteq G$ 
    \end{tabular}
  \item Subproperty: \\[0.5em]
    \begin{tabular}{llll}
      $(a)$ & $\frac{\triple{A,  \spp, B},  \triple{B,  \spp, C}}{\triple{A,  \spp, C}}$ & $(b)$ & $\frac{\triple{D,  \spp, E},  \triple{X, D, Y}}{\triple{X, E, Y}}$ 
    \end{tabular}
  \item Subclass: \\[0.5em]
    \begin{tabular}{llll}
      $(a)$ & $\frac{\triple{A, \subclass, B},  \triple{B, \subclass, C}}{\triple{A, \subclass, C}}$ & $(b)$ & $\frac{\triple{A, \subclass, B},  \triple{X, \type, A}}{\triple{X, \type, B}}$ 
    \end{tabular}
  \item Typing: \\[0.5em]
    \begin{tabular}{llll}
      $(a)$ & $\frac{\triple{D, \dom, B},  \triple{X, D, Y}}{\triple{X, \type, B}}$ & $(b)$ & $\frac{\triple{D, \range, B},  \triple{X, D, Y}}{\triple{Y, \type, B}}$ 
    \end{tabular}


      \item Conceptual Disjointness: \\[0.5em]
    \begin{tabular}{llllll}
      $(a)$ & $\frac{\triple{A,\disjC,B}}{\triple{B,\disjC,A}}$  &
      $(b)$ & $\frac{\triple{A,\disjC,B}, \triple{C,\subclass,A}}{\triple{C,\disjC,B}}$ &
      $(c)$ & \textcolor{black}{$\frac{\triple{A,\disjC,A}}{\triple{A,\disjC,B}}$} 
      \end{tabular}  
          
        \item Predicate Disjointness: \\[0.5em]
    \begin{tabular}{llllll}
      $(a)$ & $\frac{\triple{A,\disjP,B}}{\triple{B,\disjP,A}}$  &
      $(b)$ & $\frac{\triple{A,\disjP,B}, \triple{C,\spp,A}}{\triple{C,\disjP,B}}$ &
      $(c)$ & \textcolor{black}{$\frac{\triple{A,\disjP,A}}{\triple{A,\disjP,B}}$}  
          \end{tabular}  
          
      \item Crossed Disjointness: \\[0.5em]
    \begin{tabular}{llll}
      $(a)$ & $\frac{\triple{A,\dom,C}, \triple{B,\dom,D}, \triple{C,\disjC,D}}{\triple{A,\disjP,B}}$  &
      $(b)$ & $\frac{\triple{A,\range,C}, \triple{B,\range,D}, \triple{C,\disjC,D}}{\triple{A,\disjP,B}}$\\
          \end{tabular}    
          


  
    \end{enumerate}
    

\nd As anticipated, please note that the rules that extend minimal $\rhodf$ to $\rhodfbot$ are rules (5) - (7).


Now, using these rules we define a derivation relation in a similar way as in~\cite{Munoz09}.


    
\begin{definition}[Derivation $\deriv$]\label{def:derivation}
Let $G$ and $H$ be $\rhodfbot$-graphs. $ G\deriv H$ iff there exists a sequence of graphs $P_1,P_2,\ldots, P_k$ with $P_1=G$ and $P_k=H$ and for each $j$ ($2 \leq j \leq k$) one of the following cases hold:

\begin{itemize}
\item $P_j \subseteq P_{j-1}$ (rule (1));
\item there is an instantiation $R/R'$ of one of the rules (2)-(
7), such that
$R \subseteq P_{j-1}$ and $P_j = P_{j-1} \cup R'$.
\end{itemize}

\nd Such a sequence of graphs is called a proof of $G\deriv H$. Whenever
$G\deriv H$, we say that the graph $H$ is derived from the graph
$G$. Each pair $(P_{j-1}, P_j)$, $1\leq j \leq k$ is called a step of the proof which
is labeled by the respective instantiation $R/R'$ of the rule applied at
the step.
\end{definition}    

\begin{example}[Example~\ref{exrdfsC} cont.] \label{exrdfsD}
Consider the entailments in Example~\ref{exrdfsC}. The following are the proofs of them.

\begin{description}
\item[Case $\Gclass\deriv \triple{\mathit{tom}, \type, \mathit{person}}$.] \mbox{ \ } \\
{\scriptsize
\begin{prooftree}
\AxiomC{$\triple{\mathit{usesDrug}, \dom, \mathit{drugUser}}$}
\AxiomC{$\triple{\mathit{tom}, \mathit{usesDrug}, \mathit{heroin}}$}
\LeftLabel{(4a)}  \BinaryInfC{$\triple{\mathit{tom}, \type, \mathit{drugUser}}$}
\AxiomC{$\triple{\mathit{drugUser}, \subclass, \mathit{person}}$}
\LeftLabel{(3b)}  \BinaryInfC{$\triple{\mathit{tom}, \type, \mathit{person}}$}
\end{prooftree}
}


\item[Case  $\Gclass\deriv \triple{\mathit{hasDrugIndependence}, \disjP, \mathit{hasOpioidAddiction}}$.] \mbox{ \ } \\
{\scriptsize
\begin{prooftree}
\AxiomC{$\triple{\mathit{hasOpioidAddiction}, \spp, \mathit{hasAddiction}}$}
\AxiomC{$\triple{\mathit{hasAddiction}, \disjP, \mathit{hasDrugIndependence}}$}
\LeftLabel{(6b)}  \BinaryInfC{$\triple{\mathit{hasOpioidAddiction}, \disjP, \mathit{hasDrugIndependence}}$}
\LeftLabel{(5a)} \UnaryInfC{$\triple{\mathit{hasDrugIndependence}, \disjP, \mathit{hasOpioidAddiction}}$}
\end{prooftree}
}
\end{description}
\qed
\end{example}






\nd We are going now to prove the following soundness and completeness theorem (recall that proofs can be found in the appendixes).

\begin{theorem}[Soundness \& Completeness]\label{Th:soundcomplete}
Let $G$ and $H$ be $\rhodfbot$-graphs. 

\vspace{-0.3cm}
\[
G\deriv H\text{ iff }G\rdfent H \ .
\]

\end{theorem}

\nd We divide the proof into  lemmas. The following one is needed for soundness.

\begin{restatable}{lemma}{sound}\label{sound}
Let $G$ and $H$ be $\rhodfbot$-graphs, let $G$ be satisfiable, and let one of the following statements hold:
\begin{itemize}
\item $H\subseteq G$;
\item there is an instantiation $R/R'$ of one of the rules (2)-(7), such that
$R \subseteq G$ and $H = G\cup R'$.
\end{itemize}

\nd Then, $G\rdfent H$.
\end{restatable}

\nd The following lemma defines the construction of the \emph{canonical model} for $\rhodfbot$ graphs. Let $\clos(G)$ be the closure of $G$ under the application of rules $(2)-(7)$.

\begin{restatable}{lemma}{completefirst}\label{completefirst}
Given a $\rhodfbot$-graph $G$,  define  an \emph{interpretation} $\I_G$ as a tuple 
\[
\I_G =\tuple{\Delta_{\DR}, \Delta_{\DP}, \Delta_{\DC},  \Delta_{\DL}, \intP{\cdot}, \intC{\cdot}, \intG{\cdot}}
\]

\nd such that: 
\begin{enumerate}
\item $\Delta_{\DR}:= \universe(G)\cup \rhodfbot$;




\item $\Delta_{\DP}:= \{p\in \universe(G)\mid (s,p,o)\in \clos(G)\}\cup \  \rhodfbot  \ \cup \ \{p\in \universe(G)\mid \text{either }
\triple{p,\spp, q}$, $\triple{q,\spp, p}$, $\triple{p,\Dom, c}$, $\triple{p,\range,d}$, $\triple{p,\disjP,q} \text{ or } \triple{q,\disjP,p}\in  \clos(G) \}$;

\item $\Delta_{\DC}:=\{c\in \universe(G)\mid \triple{x,\type, c}\in \clos(G)\} \cup\{c\in \universe(G)\mid \text{either } \triple{c,\subclass, d}$, $\triple{d,\subclass, c}$, $\triple{p,\Dom, c}$, $\triple{p,\range,c}$, $\triple{c,\disjC, d} \text{ or } \triple{d,\disjC,c}\in \clos(G)\}$; 

\item $\Delta_{\DL} := \universe(G)\cap \AL $;

\item  $\intP{\cdot}$ is an extension function $\intP{\cdot}\colon \Delta_{\DP} \to 2^{\Delta_{\DR} \times \Delta_{\DR}}$  s.t.~$\intP{p}:=\{(s,o)\mid\triple{s,p,o}\in \clos(G)\}$;

%
	
\item $\intC{\cdot}$ is an extension function $\intC{\cdot}\colon \Delta_{\DC} \to 2^{\Delta_{\DR}}$ s.t. $\intC{c}:=\{x\in \universe(G)\mid\triple{x,\type,c}\in \clos(G)\}$;
\item $\intG{\cdot}$ is an identity function over $\Delta_{\DR}$. 


\end{enumerate}

\nd Then, for every $\rhodfbot$-graph $G$, $\I_G\rdfsat G$.

\end{restatable}




\nd Now, we have: 
    
\begin{restatable}{lemma}{completesecond}\label{completesecond}
Let $G$ and $H$ be $\rhodfbot$-graphs. If $G\rdfent H$ then $H\subseteq  \clos(G)$.
\end{restatable}

%
%

\nd Finally, we can prove the main theorem of this section.

\begin{proof}[Proof of Theorem \ref{Th:soundcomplete}]
The proof mirrors the proof of Theorem 8 in \cite{Munoz09}. 
From Lemma \ref{completesecond}, $G\rdfent H$ implies that $H$ can be obtained from $\clos(G)$ using rule (1). Thus, since $G\deriv \clos(G)$, it follows that $G\deriv H$. 
Therefore Theorem \ref{Th:soundcomplete}  follows from Lemmas \ref{sound} and \ref{completesecond}.
\end{proof}

\nd Please note that, like in classical $\rhodf$, \rhodfbot-graphs are always satisfiable.

\begin{corollary}\label{Coroll:sat_bot}
A $\rhodfbot$-graph $G$ is always satisfiable.
\end{corollary}

\begin{proof}
This is an immediate consequence of Lemma \ref{completefirst}.
\end{proof}

\begin{example}\label{ex_can_model}
Consider the simple graph $G=\{\triple{a,\type,c},\triple{a,\type,d},\triple{c,\disjC,d}\}$. The interpretation $\I_G$ is defined as follows:

\begin{itemize}
   \item $\Delta_{\DR}=\{a,c,d\}\cup\rhodfbot$;
   \item $\Delta_{\DP}=\{\type,\disjC\}$;
   \hspace{0.2cm} $\Delta_{\DC}=\{c,d\}$;
   \hspace{0.2cm} $\Delta_{\DL}=\emptyset$;
   \item $\intP{\type}=\{\tuple{a,c},\tuple{a,d}\}$; \hspace{0.2cm} $\intP{\disjC}=\{\tuple{c,d},\tuple{d,c}\}$;
   \item $\intC{c}=\{a\}$;  \hspace{0.2cm}$\intC{d}=\{a\}$;
   \item $\intG{x}= x$, for every $x\in\Delta_{\DR}$.
\end{itemize}
\qed
\end{example}

\subsection{Some Interesting Derived Inference Rules}
\nd In what follows, we illustrate the derivation of some interesting rules of inference. 
To start with, note that the triples $\triple{d,\disjC, d}$ and $\triple{q,\disjP, q}$ are particularly significant: indeed, the intended meaning of \eg~$\triple{d,\disjC,d}$ is that  `concept/class $d$ is empty'. 

%
%
%

Some derived inference rules that will turn out to be useful are the following:
\begin{description}
\item[Empty Subclass:] \fnl
{\footnotesize
\begin{prooftree}
\AxiomC{$\triple{A,\subclass,B}$}
\AxiomC{$\triple{A,\subclass,C}$}
\AxiomC{$\triple{B,\disjC,C}$}
\RightLabel{(EmptySC)}\TrinaryInfC{$\triple{A,\disjC,A}$}
\end{prooftree}
}
\nd Here is the derivation:
{\footnotesize
\begin{prooftree}
\AxiomC{$\triple{A,\subclass,B}$}
\AxiomC{$\triple{B,\disjC,C}$}
\LeftLabel{(5b)}\BinaryInfC{$\triple{A,\disjC,C}$}
\LeftLabel{(5a)}\UnaryInfC{$\triple{C,\disjC,A}$}
\AxiomC{$\triple{A,\subclass,C}$}
\RightLabel{(5b)}\BinaryInfC{$\triple{A,\disjC,A}$}
\end{prooftree}
}
\nd A special case of  rule (EmptySC) is obtained by imposing $B=C$, shows that if a class is empty, also all its subclasses are empty.
{\footnotesize
\begin{prooftree}
\AxiomC{$\triple{A,\subclass,B}$}
\AxiomC{$\triple{B,\disjC,B}$}
\RightLabel{(EmptySC')}\BinaryInfC{$\triple{A,\disjC,A}$}
\end{prooftree}
}
\item[Empty Subpredicate:] \fnl
{\footnotesize
\begin{prooftree}
\AxiomC{$\triple{A,\spp,B}$}
\AxiomC{$\triple{A,\spp,C}$}
\AxiomC{$\triple{B,\disjP,C}$}
\RightLabel{(EmptySP)}\TrinaryInfC{$\triple{A,\disjP,A}$}
\end{prooftree}
}
\nd Here is the derivation:
{\footnotesize
\begin{prooftree}
\AxiomC{$(A,\spp,B)$}
\AxiomC{$(B,\disjP,C)$}
\LeftLabel{(6b)}\BinaryInfC{$(A,\disjP,C)$}
\LeftLabel{(6a)}\UnaryInfC{$(C,\disjP,A)$}
\AxiomC{$(A,\spp,C)$}
\RightLabel{(6b)}\BinaryInfC{$(A,\disjP,A)$}
\end{prooftree}
}
\nd As for (EmptySC), by imposing $B=C$ we obtain a special case of the rule (EmptySP) showing that if a property is empty, so are all its subproperties.
{\footnotesize
\begin{prooftree}
\AxiomC{$\triple{A,\spp,B}$}
\AxiomC{$\triple{B,\disjP,B}$}
\RightLabel{(EmptySP')}\BinaryInfC{$\triple{A,\disjP,A}$}
\end{prooftree}
}
\item[Conflicting Domain:] \fnl
{\footnotesize
\begin{prooftree}
\AxiomC{$\triple{A,\dom,C}$}
\AxiomC{$\triple{A,\dom,X}$}
\AxiomC{$\triple{C,\disjC,X}$}
\RightLabel{(7a')}\TrinaryInfC{$\triple{A,\disjP,A}$}
\end{prooftree}
}
\nd This is a special case of the rule (7a) in which $B=A$.
\item[Conflicting Range:] \fnl
{\footnotesize
\begin{prooftree}
\AxiomC{$\triple{A,\range,C}$}
\AxiomC{$\triple{A,\range,X}$}
\AxiomC{$\triple{C,\disjC,X}$}
\RightLabel{(7b')}\TrinaryInfC{$\triple{A,\disjP,A}$}
\end{prooftree}
}
\nd Similarly as above, this is a special case of the rule (7b) in which $B=A$.
\end{description}

\section{Defeasible $\rhodfbot$} \label{defrdfs}

\nd We now introduce the possibility of modelling defeasible information in the RDFS framework. 

\subsection{Syntax}
\nd We will consider defeasibility  w.r.t. the predicates $\subclass$ and $\spp$ only, and introduce the notion of \emph{defeasible triple} defined next.

\begin{definition}[Defeasible triple]\label{def:defeasible_triple}
A \emph{defeasible triple} is of the form
\[
\delta=\dtriple{s,p,o}\in\AUL\times\{\subclass,\spp\}\times\AUL \  ,
\]
\nd where $s,o \not\in \rhodfbot$.
\end{definition}

\nd The intended meaning of \eg~$\dtriple{c,\subclass, d}$ is ``Typically, an instance of  $c$ is also an instance of  $b$''.  So, for instance, by referring to Example~\ref{ex01},  the defeasible statement $(3)$ can be represented as the defeasible triple $\dtriple{\mathit{drugUser},\subclass,\mathit{unhappyPerson}}$.
Analogously, $\dtriple{p,\spp, q}$ is read as ``Typically, a pair related by $p$ is also related by $q$''. So, for instance, by referring to Figure~\ref{figrdfs}, the 
defeasible triple $\dtriple{\mathit{usesDrug}, \spp, \mathit{hasDrugAddiction}}$ aims at representing the fact that ``typically, if someone uses some drug then it may become addicted to that drug".

A \emph{defeasible graph} is a set  $G=\Gclass\cup \Gdef$, where $\Gclass$ is a \rhodfbot-graph and $\Gdef$ is a set of defeasible triples. 

\begin{example}[Running example cont.]\label{exrdfsDF}
The following set $\Gdef$ is a set of defeasible triples (\cf~green triples within Figure~\ref{figrdfs})
\begin{eqnarray*}
\Gdef & =  \{   &  \dtriple{\mathit{drugUser},\subclass,\mathit{unhappyPerson}}, \dtriple{\mathit{drugUser},\subclass,\mathit{youngPerson}}, \\
&& \dtriple{\mathit{youngPerson}, \subclass, \mathit{student}}, \dtriple{\mathit{controlledDrugUser}, \subclass, \mathit{happyPerson}},  \\
&& \dtriple{\mathit{youngPerson}, \subclass, \mathit{happyPerson}}, \dtriple{\mathit{usesDrug}, \spp, \mathit{hasDrugAddiction}}, \\
&& \dtriple{\mathit{usesDrugControlled}, \spp, \mathit{hasDrugIndependence}} \} \ . 
\end{eqnarray*}
\nd Therefore, the whole graph in Figure~\ref{figrdfs} corresponds to the defeasible graph
\begin{eqnarray*}
G & = &  \Gclass \cup \Gdef \ ,
\end{eqnarray*}
\nd where $\Gclass$ is  defined in Example~\ref{exrdfsB}.
 \qed
\end{example}


\nd Given two defeasible graphs $G$ and $G'$, $G$ is a sub-graph of $G'$ iff $G\subseteq G'$.
In the following, we use the notation $\anytriple{\cdot,\cdot,\cdot}$ to indicate either $\triple{\cdot,\cdot,\cdot}$ or $\dtriple{\cdot,\cdot,\cdot}$ (that is,  $\anytriple{\cdot,\cdot,\cdot} \in\{\triple{\cdot,\cdot,\cdot},\dtriple{\cdot,\cdot,\cdot}\}$).

\begin{remark}
Note that in practice a defeasible triple $\dtriple{c,\subclass, d}$ may be represented as $\triple{c,\subclass_t, d}$, where $\subclass_t$ is a new  symbol indicating defeasible class inclusion. Similarly,  $\dtriple{p,\spp, q}$ may be represented as $\triple{p,\spp_t, q}$, where $\spp_t$ is a new  symbol indicating defeasible property inclusion. Therefore, both defeasible triples could have been represented in the RDF language with vocabulary $\rhodfbot \cup\{\subclass_t, \spp_t\}$. While certainly this is an option for a practical implementation, for ease of presentation, we prefer to stick to the former notation.
\end{remark}


\nd Given a defeasible graph $G=\Gclass\cup \Gdef$, its \emph{strict counterpart} is the graph
\begin{equation}\label{eq:Gs}
\Gcount:=\Gclass\cup\{\triple{s,p,o}\mid \dtriple{s,p,o}\in\Gdef\} \ .
\end{equation}

\nd Generally speaking, given a defeasible graph $G$, we need to define how to reason with it. As mentioned above, in presumptive reasoning it is considered rational to reason classically with defeasible information in case no conflicts arise; on the other hand, if we have to deal with conflictual information we need to introduce  some form of defeasible reasoning to resolve such conflicts. 

To do so, first of all, we need to define the notion of \emph{conflict} in our framework.

\begin{definition}[conflict]
Let $G$ be a defeasible graph. $G$ has a \emph{conflict} if, for some term  $t$, either
$\Gcount\deriv (t,\disjC, t)$ or $\Gcount\deriv (t,\disjP, t)$ holds.
\end{definition}

\nd Informally, we consider that there is a conflict in a defeasible graph if, treating the defeasible triples as classical triples, we have that some term $t$ must be interpreted as being `empty'.\footnote{Note that the concept of conflict that we define here is related to the notion of \emph{incoherence} in OWL formalism \cite{QiHunter2007}: an OWL ontology is incoherent if a concept introduced in the vocabulary turns out to be empty.}

To give a sense of the rationale behind this definition, let's see how it behaves \wrt~our running example.

\begin{example}[Running example cont.]\label{pex}
Consider the defeasible graph $G$ defined in Example~\ref{exrdfsDF}.  Let us prove that $G$ has a conflict. To do so, note that
the strict counterpart of $G$ is
\begin{eqnarray*}
\Gcount & =   & \Gclass \cup \{   \triple{\mathit{drugUser},\subclass,\mathit{unhappyPerson}}, \triple{\mathit{drugUser},\subclass,\mathit{youngPerson}}, \\
&& \triple{\mathit{youngPerson}, \subclass, \mathit{student}}, \triple{\mathit{controlledDrugUser}, \subclass, \mathit{happyPerson}},  \\
&& \triple{\mathit{youngPerson}, \subclass, \mathit{happyPerson}}, \triple{\mathit{usesDrug}, \spp, \mathit{hasDrugAddiction}}, \\
&& \triple{\mathit{usesDrugControlled}, \spp, \mathit{hasDrugIndependence}} \}  \ .
\end{eqnarray*}
%
\nd  Now, let us prove that $\Gcount\deriv\triple{\mathit{drugUser},\disjC, \mathit{drugUser}}$,  \ie~from a ``strict" point of view,  a $\mathit{drugUser}$ cannot exist. 
For ease of presentation, we will use the following abbreviations:
\begin{eqnarray*}
\mathit{unhappyPerson} & \mapsto & uhP \\
\mathit{happyPerson} & \mapsto & hP \\
\mathit{drugUser} & \mapsto & dU \\
\mathit{youngPerson} & \mapsto & yP \ .
\end{eqnarray*}
\nd Now, by rule $(5a)$  applied to 
$\triple{uhP,\disjC,hP}$ we immediately have $\Gcount\deriv\triple{hP,\disjC, uhP}$. Therefore, we have the following proof of 
$\Gcount\deriv\triple{dU,\disjC, dU}$:
%
{\small
\begin{prooftree}
\AxiomC{$\triple{dU,\subclass,yP}$}
\AxiomC{$\triple{yP, \subclass, hP}$}
\LeftLabel{(3a)}\BinaryInfC{$\triple{dU,\subclass,hP}$}
\AxiomC{$\triple{dU,\subclass,uhP}$}
\AxiomC{$\triple{hP,\disjC,uhP}$}
\RightLabel{(EmptySC)}\TrinaryInfC{$\triple{dU,\disjC,dU}$}
\end{prooftree}
}
\nd As a consequence, $G$ has a conflict. Note that in a  similar way we may prove that 
\[
\Gcount\deriv\triple{\mathit{usesDrugControlled},\disjP, \mathit{usesDrugControlled}} \ .
\]
\qed
\end{example}
%
%
%
%
%
%
%
%

\subsection{Semantics}\label{sect_semantics}


\nd An interpretation for a defeasible graph $G$ is composed by a set of $\rhodfbot$ interpretations, ranked accordingly to how much they conform to our expectations.

\begin{definition}[Ranked $\rhodfbot$ Interpretations]\label{defint}
A ranked interpretation is a pair $\R=(\M,r)$, where $\M$ is the set of all $\rhodfbot$ interpretations defined on a fixed set of domains $\Delta_{R}, \Delta_{P}$, $\Delta_{C}, \Delta_{L}$, and $r$ is a ranking function over $\M$~\footnote{We will assume that $0 \in \mathbb{N}$.}
\[
r:\M\mapsto \mathbb{N}\cup\{\infty\}
\]	
\nd satisfying a convexity property:
\begin{itemize}
\item there is an interpretation $\I\in\M$ s.t. $r(\I)=0$;
\item for each $i>0$, if there is an interpretation $\I\in\M$ s.t. $r(\I)=i$, then there is an interpretation $\I'\in\M$ s.t. $r(\I')=(i-1)$.
\end{itemize}
\end{definition}
\nd Informally, the intuition behind ranking interpretations is that, given two $\rhodfbot$ interpretations $\I, \I'\in\M$,  $r(\I)<r(\I')$ indicates that the interpretation $\I$ is more in line with our expectations than the interpretation $\I'$. Given $\R=(\M,r)$, let $\MN$ be the set of elements in $\M$ with rank lower that $\infty$, that is, 
\[
\MN = \{\I\in\M\mid r(\I)\in\mathbb{N}\} \ .
\]
\nd Informally, in $\R$ the $\rhodfbot$-interpretations with rank infinity are simply considered impossible, and the satisfaction relation, which we will define next, is determined referring only to the $\rhodfbot$-interpretations in $\MN$.

\begin{example}\label{ex_ranked_int}
Consider a  vocabulary that allows to talk about birds ($b$), sparrows ($s$), and their flying abilities ($f$). Table \ref{Fig_ex_ranked_int} represents a ranked interpretation $\R=(\M,r)$ in which rank $0$ is populated with the $\rhodfbot$-interpretations that satisfy $\triple{b,\disjC,b}$ and $\triple{s,\subclass,b}$; rank $1$ with the $\rhodfbot$-interpretations that satisfy $\triple{b,\subclass,f}$ and $\triple{s,\subclass,b}$ and that are not already in rank $0$; rank $2$ with the $\rhodfbot$-interpretations  that satisfy $\triple{s,\subclass,b}$ and that are not already in the lower ranks, $0$ and $1$; finally, rank $\infty$ is populated by the $\rhodfbot$-interpretations  that do not satisfy $\triple{s,\subclass,b}$. 

Hence, $\R$ represents the following expectations. In all the conceivable situations (that is, in $\MN$, from rank $0$ to rank $2$), sparrows are birds, since $\triple{s,\subclass,b}$ is satisfied. In the most expected situations, that is, rank $0$, birds do not actually exist ($\triple{b,\disjC,b}$ holds). Then, in rank $1$, there are the situations in which birds exist, and they fly, since $\triple{b,\subclass,f}$ is satisfied. Rank $2$ represents the situations in which birds exist, but they do not necessarily fly ($\triple{b,\subclass,f}$ is not satisfied); such  situations  are conceivable, but less expected. Eventually we have the situations that are considered impossible, represented by rank $\infty$,  in which sparrows are not necessarily birds ($\triple{s,\subclass,b}$ does not hold).

\begin{table}[h]
\caption{A ranked interpretation $\R=(\M,r)$, as described in Example \ref{ex_ranked_int}. $\M$ is the set of all the $\rhodfbot$-interpretations for the given vocabulary. $\llbracket \triple{b,\disjC,b}\cup\triple{s,\subclass,b}\rrbracket$ indicates all the $\rhodfbot$-interpretations in $\M$ that satisfy both $\triple{b,\disjC,b}$ and $\triple{s,\subclass,b}$, $\llbracket \triple{b,\subclass,f}\cup\triple{s,\subclass,b}\rrbracket$ all the $\rhodfbot$-interpretations that satisfy both $\triple{b,\subclass,f}$ and $\triple{s,\subclass,b}$, and so on.}
\label{Fig_ex_ranked_int}
\begin{center}
\renewcommand{\arraystretch}{1.5}
\begin{tabular}{|c|c|}%
\hline
{ rank $\infty$} & {$\M^F\setminus$ (rank $0$ $\cup$ rank $1$ $\cup$ rank $2$)}\\
\hline
{ rank $2$} & {$\llbracket \triple{s,\subclass,b}\rrbracket\setminus$ (rank $0$ $\cup$ rank $1$)}\\
\hline
{ rank $1$} & {$\llbracket \triple{b,\subclass,f}\cup\triple{s,\subclass,b}\rrbracket\setminus$ rank $0$}\\
\hline
{ rank $0$} & {$\llbracket \triple{b,\disjC,b}\cup\triple{s,\subclass,b}\rrbracket$} \\
\hline

\end{tabular}
\end{center}
\end{table}

\end{example}

\nd Given a ranked interpretation $\R=(\M,r)$ and a term $t$, let $\cmin(t,\R)$ be the set of the most expected interpretations in $\M$ in which $t$ (interpreted as class) is not empty, that is, 
\begin{eqnarray*}
\cmin(t,\R) & = &\{\I\in\MN\mid \I\not \rdfsat \triple{t,\disjC, t} \text{ and there is no }\I'\in\MN\text{ s.t. } \\
&& \ \ \I'\not \rdfsat \triple{t,\disjC, t} \text{ and }r(\I')<r(\I)\} \ .
\end{eqnarray*}
\nd Analogously, for a term $t$ interpreted as predicate, we define
\begin{eqnarray*}
\pmin(t,\R) & = &\{\I\in\MN\mid \I\not \rdfsat \triple{t, \disjP, t} \text{ and there is no }\I'\in\MN\text{ s.t. } \\
&& \ \ \I'\not \rdfsat \triple{t,\disjP, t} \text{ and }r(\I')<r(\I)\} \ .
\end{eqnarray*}
\begin{definition}[Ranked satisfaction]\label{defsat}
For every triple $\triple{s,p,o}$, a ranked interpretation $\R=(\M,r)$ \emph{satisfies} $\triple{s,p,o}$ if $\triple{s,p,o}$ is satisfied by every $\rhodfbot$-interpretation in $\M$, that is,
\[
\R\rdfsat\triple{s,p,o}  \text{ iff }  \I\rdfsat\triple{s,p,o} \text{ for every } \I\in\MN \ .
\]
\nd For every  defeasible triple of the form $\dtriple{s,p,o}$, the notion of a ranked interpretation $\R=(\M,r)$ \emph{satisfying} $\dtriple{s,p,o}$, denoted $\R\rdfsat\dtriple{s,p,o}$, is defined as follows:
\begin{eqnarray*}
\R\rdfsat\dtriple{c,\subclass,d} & \text{iff}&  \I\rdfsat\triple{c,\subclass,d} \text{ for every } \I\in\cmin(c,\R) \\
\R\rdfsat\dtriple{p,\spp,q} & \text{iff} &  \I\rdfsat\triple{p,\spp,q} \text{ for every } \I\in \pmin(p,\R) \ .
\end{eqnarray*}

%

\nd Given a defeasible $\rhodfbot$-graph $G=\Gclass\cup\Gdef$, a ranked interpretation $\R=(\M,r)$ is a \emph{model} of $G$ (denoted $\R\rdfsat G$) if $\R\rdfsat \anytriple{s,p,o}$ for every $\anytriple{s,p,o}\in G$.


%
%

\end{definition}

\nd The intuition behind this definition is that the triple $\dtriple{c,\subclass,s}$ holds in a ranked interpretation if $\triple{c,\subclass,s}$ holds in the most expected $\rhodfbot$-interpretations in which $c$ is not an empty class. The intuition follows a similar line of the original propositional construction \cite{Lehmann92b}, and its DL reformulations~\cite{BritzEtAl2015a,Casini10,Giordano15}.

\begin{example}\label{ex_ranked_sat}
Consider the ranked interpretation $\R=(\M,r)$ in Example \ref{ex_ranked_int}. $\MN$ corresponds to all the $\rhodfbot$- interpretations in ranks from $0$ to $2$. Hence we have 

\[\R\rdfsat\triple{s,\subclass,b},\]

\nd since all the $\rhodfbot$-interpretations in $\MN$ satisfy $\triple{s,\subclass,b}$. That is, according to $\R$ sparrows are a subclass of birds. Moreover, we also have

\[\R\rdfsat\dtriple{b,\subclass,f}.\]

\nd That is, typical birds fly. To see this, we have to look at the most expected $\rhodfbot$-interpretations in which birds exist, that is, $\cmin(b,\R)$. Now, $\cmin(b,\R)$ corresponds to the $\rhodfbot$-interpretations in rank $1$, since in rank $0$ all the interpretations satisfy $\triple{b,\disjC,b}$. As all the $\rhodfbot$-interpretations in rank $1$ satisfy $\triple{b,\subclass,f}$, we have that $\R$ satisfies $\dtriple{b,\subclass,f}$.

\end{example}

\nd As in the propositional and DL constructions, once we have defined the notion of ranked interpretation to model defeasible information, the problem is to decide which kind of entailment relation, that is, what kind of defeasible reasoning, we would like to model. Despite it is recognised that there are multiple available options according to the kind of properties we want to satisfy \cite{CasiniStraccia13,CasiniEtAl2014,Giordano15,Kraus90,Lehmann95,Lehmann92b,Pearl90}, it is generally recognised that RC~\cite{Lehmann92b} is the fundamental construction in the area, and most of the other proposed systems can be built as refinements of it. We recall that RC models the so-called \emph{Presumption of Typicality} \cite[p.4]{Lehmann95}, that is the reasoning principle imposing that, if we are not informed about any exceptional property, we presume that we are dealing with a typical situation. The essential behaviour characterising the presumption of typicality is that a subclass that does not show any exceptional property inherits all the typical properties of the superclass. From a semantical point of view the definition of RC can be obtained via various equivalent definitions \cite{HillParis2003,Lehmann92b}: here we opt for the characterisation of RC using the minimal ranked model \cite{Booth98,Pearl90} and, in particular, we consider the characterisation given in~\cite{Giordano15}, which, once adjusted accordingly,  we believe appropriate for our defeasible RDF framework.


In the following, given a defeasible graph $G=\Gclass\cup\Gdef$ and  its strict counterpart $\Gcount$  (see Equation~\ref{eq:Gs}), 
%
%
%
 let the interpretation $\I_{G^s}$ 
\begin{equation} \label{gsdom}
\I_{G^s} :=\tuple{\Delta_{\DR}, \Delta_{\DP}, \Delta_{\DC},  \Delta_{\DL}, \intP{\cdot}, \intC{\cdot}, \intGs{\cdot}}
\end{equation}

\nd be defined from $\Gcount$ as in Lemma~\ref{completefirst}, and consider the domains $\Delta_{\DR}, \Delta_{\DP}, \Delta_{\DC},  \Delta_{\DL}$ in it. With $\M^G$ we denote the set of all $\rhodfbot$-interpretations defined over such domains,  and we indicate a ranked interpretation built over $\M^G$ as $\R^G = (\M^G,r)$. If clear from the context, we may omit the superscript $G$ in $\R^G = (\M^G,r)$.\footnote{Please note that the domains of all interpretations occurring in $\R^G$ are defined as in Equation~\ref{gsdom}.}

Now, given a defeasible graph $G$, let $\IG$ be the set of the ranked interpretations $\R^G$ and let $\RG$ be the elements of $\IG$ that are also models for $G$, that is,
\begin{equation}\label{Eq_rg}
\RG:=\{\R \in \IG\mid \R \rdfsat G\} \ .
\end{equation}

\nd Note that $\RG$ is not empty, since for any defeasible graph $G$ it is easy to define a trivial ranked model: using the construction in Lemma~\ref{completefirst} we define the $\rhodfbot$-interpretation $\I_{G^s}$, that is a model of $G^s$, the strict counterpart of $G$ (see above); consider the ranked interpretation  $\R^{tr} = (\M^G,r)$, where $r(\I_{G^s})=0$ and $r(\I)=\infty$ for any other $\I\in\M^G$. It is easy to check that $\R^{tr}$ is a model of the defeasible graph $G$. 

Moreover, since the domains in Equation~\ref{gsdom} are finite, $\M^G$ is finite as well. Additionally, as a ranked interpretation $\R^G = (\M^G,r)$ built over $\M^G$ has to satisfy the convexity property (see Definition~\ref{defint}), the ranking function $r$ is bounded by $|\M^G|$.\footnote{That is, the maximal rank of any interpretation in $\M^G$ cannot exceed $|\M^G|$.} Therefore, $\IG$ is finite and, thus, so is $\RG$.


Please note that by construction $\R,\R' \in \RG$ differ only \wrt~the involved ranking functions $r,r'$, respectively, which induces the following order over the ranked models in $\RG$.
\begin{definition}[Presumption ordering $\preceq$]\label{Def_presord}
Let $\R=(\M,r)$, $\R'=(\M,r')$, and $\R,\R'\in \RG$.  We define
\begin{enumerate}
\item $\R\preceq\R'$ iff for every $\I\in \M$, $r(\I)\leq r'(\I)$.
\item $\R\prec\R'$ iff $\R\preceq\R'$ and $\R'\not\preceq\R$.
\item $\min_\preceq (\RG):=\{\R\in\RG\mid \text{there is no }\R'\in\RG\text{ s.t. }\R'\prec R\}$.
\end{enumerate}
\end{definition}
%
%
%
\nd The set $\min_\preceq (\RG)$ contains the ranked models of $G$ in which the $\rhodfbot$-interpretations are ``pushed down" as much as possible in the ranking, that is, they are considered as typical as possible. 

We next show that actually there is a unique minimal ranked model for a  defeasible graph.

\begin{restatable}{proposition}{propuniqueminimal}\label{Propuniqueminimal}
For every defeasible graph $G$, $|\min_\preceq (\RG)| = 1$.
\end{restatable}

\nd In the following, the unique $\preceq$-minimal ranked model in $\RG$ is called the \emph{minimal $G$-model} and is denoted with $\CG$, \ie~$\min_\preceq (\RG) = \{\CG\}$. 

Hence, Definition \ref{Def_presord} and Proposition \ref{Propuniqueminimal} together tell us that the only minimal $G$-model $\CG$ of a graph $G$ is the ranked model of $G$ in which every $\rhodfbot$-interpretation is positioned in the lowest possible rank, modulo the satisfaction of the graph $G$. That is, it is the model of $G$ in which we consider every $\rhodfbot$-interpretation as much typical as possible (that is, in the lowest possible rank), given the information at our disposal (the graph $G$).

\begin{definition}[Minimal Entailment]\label{Def_min_entail}
Given a defeasible graph $G$ and the corresponding minimal $G$-model $\CG$ of $G$.
A defeasible graph $G$ \emph{minimally entails} a triple $\anytriple{s,p,o}$, denoted $G\minentail\anytriple{s,p,o}$,
iff $\CG\rdfsat \anytriple{s,p,o}$.
\end{definition}

\nd Using the minimal ranking in $\CG$ we can also define the \emph{height} of a term, indicating at which level of exceptionality a term $t$ is not necessarily empty, as a class or as a predicate. This corresponds to the minimal rank, that is, the rank in $\CG$, in which we encounter a \rhodfbot-interpretation that does not satisfy $\triple{t,\disjC,t}$, or, respectively, $\triple{t,\disjP,t}$.

\begin{definition}[Height]\label{Def_concept_predicate_height}
Let $G=\Gclass\cup\Gdef$ be a defeasible graph, with $\CG=\{\M,r\}$ being its minimal model, and let $t$ be a term in $G$. The \emph{\ssc-height} of $t$, indicated as $\hc(t)$, corresponds to the lowest rank $r(\I)$ of some $\I\in\M$ s.t. $\I\not\rdfsat\triple{t,\disjC,t}$, that is, 
%
%
\begin{equation*}\label{def_hc}
    \hc(t)=\left\{\begin{array}{l}
        \infty  \text{,\hspace{1,2cm} if }\I\rdfsat\triple{t,\disjC,t}\text{ for every }\I\in\MN\\
        \min\{r(\I)\mid \I\in \M\text{ and }\I\not\rdfsat\triple{t,\disjC,t}\}  \text{,\hspace{0.2cm} otherwise.}
    \end{array}\right.
\end{equation*}
\nd Analogously, the \emph{\ssp-height} of $t$, indicated as $\hp(t)$, corresponds to the lowest rank $r(\I)$ of some $\I\in\M$ s.t. $\I\not\rdfsat\triple{t,\disjP,t}$
%
%
\begin{equation*}\label{def_hp}
    \hp(t)=\left\{\begin{array}{l}
        \infty  \text{,\hspace{1,2cm} if }\I\rdfsat\triple{t,\disjP,t}\text{ for every }\I\in\MN\\
        \min\{r(\I)\mid \I\in \M\text{ and }\I\not\rdfsat\triple{t,\disjP,t}\}  \text{,\hspace{0.2cm} otherwise.}
    \end{array}\right .
\end{equation*}
\nd Also, we define the \emph{height $h(\R)$} of a ranked interpretation $\R$ as the highest finite rank of the \rhodfbot-interpretations in it. That is, let $\R=(\M,r)$ be a ranked model, then 
\[
h(\R)\assign \max\{r(\I)\mid \I\in \MN\} \ .
\]
\end{definition}

\nd The \emph{\ssc-height} of a class $c$ indicates how exceptional the objects in it are w.r.t. a graph $G$. We consider the minimal model $\RG$ of the graph $G$, in which every $\rhodfbot$-interpretation is associated to the lowest possible rank. If there is some interpretation $\I$ in rank $0$ such that $\I\not\rdfsat\triple{c,\disjC,c}$, then it is reasonable to expect a situation in which $c$ is a non-empty class. In such a case $\hp(c)=0$, that indicates that $c$ is a class that is compatible with all our expectations (represented by the defeasible part of the graph $G$). If, instead, the first $\rhodfbot$-interpretations not satisfying $\triple{c,\disjC,c}$ appear higher in the ranks, then the class $c$ is exceptional, that is, it can be populated only by objects that do not fully satisfy our expectations. The higher the height, the more exceptional the population of $c$ must be \wrt~the expectations formalised by the graph $G$. Analogously for the \emph{\ssp-height} of a predicate $p$. The following example uses the classical simple penguin scenario to show that the \emph{\ssc-height} of the class of penguins is higher than the  \emph{\ssc-height} of the class of birds, since penguins are exceptional non-flying birds.

\begin{example}[Penguin example]\label{ex_semantics}
Consider the typical \emph{penguin example} from the non-monotonic reasoning literature.
That is, we have a set $F^{str}$ of classical triples and the set $F^{def}$ containing defeasible triples:

\[
F^{str}=\{\triple{p,\subclass,b},\triple{s,\subclass,b},\triple{p,\subclass, e},\triple{e,\disjC,f}\} \text{ and }
F^{def}=\{ \dtriple{b,\subclass,f}\} \ ,
\]

\nd where $p,b,f,e,s$ stand for \emph{penguins, birds, flying creatures, non-flying creatures, sparrow}, respectively. Together they define the graph $F$:

\[
F=F^{str}\cup F^{def}=\{\triple{p,\subclass,b},\triple{s,\subclass,b},\triple{p,\subclass, e},\triple{e,\disjC,f},\dtriple{b,\subclass,f}\} \ . 
\]

\nd  Its strict counterpart is 
\[
F^s=\{\triple{p,\subclass,b},\triple{s,\subclass,b},\triple{p,\subclass, e},\triple{e,\disjC,f}, \triple{b,\subclass,f}\} \ .
\]
\nd Given  $F^s$, we define the domains used for the models in $\mathfrak{R}_F$ following the construction in Lemma \ref{completefirst}. The minimal model $\R_{\min F}\in\mathfrak{R}_F$  will be a model of height $1$: all the $\rhodfbot$-interpretations that satisfy $F^s$ will have rank $0$, all the $\rhodfbot$-interpretations that satisfy $F^{str}\cup\{\triple{b,\disjC,b}\}$ will have rank $0$, all the $\rhodfbot$-interpretations that satisfy $F^{str}$ but neither $\triple{b,\subclass,f}$ nor $\triple{b,\disjC,b}$ will have rank $1$, and all the  $\rhodfbot$-interpretations that do not satisfy $F^{str}$ will have infinite rank. The model is shown in Table~\ref{Fig_ex3.3}.

\begin{table}[h]
\caption{The minimal model of graph $F$ from Example \ref{ex_semantics}. $\llbracket F^{s}\rrbracket$ indicates all the $\rhodfbot$-interpretations in $\M^F$ that satisfy the graph $F^{s}$, $\llbracket F^{str}\cup\triple{b,\disjC,b}\rrbracket$ all the $\rhodfbot$-interpretations in $\M^F$ that satisfy the graph $F^{str}\cup\triple{b,\disjC,b}$, and so on.}
\label{Fig_ex3.3}
\begin{center}
\renewcommand{\arraystretch}{1.5}
\begin{tabular}{|c|c|}%
\hline
{ rank $\infty$} & {$\M^F\setminus$ (rank $0$ $\cup$ rank $1$)}\\
\hline
{ rank $1$} & {$\llbracket F^{str}\rrbracket\setminus$ rank $0$}\\
\hline
{ rank $0$} & {$\llbracket F^{s}\rrbracket$ $\cup$ $\llbracket F^{str}\cup\triple{b,\disjC,b}\rrbracket$}\\
\hline

\end{tabular}
\end{center}
\end{table}

\nd It is easily verified that $F^s$ implies $\triple{p,\disjC,p}$. Also, it is easy to check that $F^s$ does not imply $\triple{b,\disjC,b}$ and does not imply $\triple{s,\disjC,s}$, since we are informed that penguins do not fly, but we are not informed that sparrows do not fly (there is no $\triple{s,\subclass, e}$ in our graph); also, $F^{str}$ does not imply $\triple{p,\disjC,p}$, since $F^{str}$ does not contain  $\triple{b,\subclass,f}$ anymore. The resulting configuration of $\R_{\min F}$ is such that: all the $\rhodfbot$-interpretations with height $0$ satisfy $\triple{p,\disjC,p}$, but not all of them  satisfy $\triple{b,\disjC,b}$ or $\triple{s,\disjC,s}$; not all the $\rhodfbot$-interpretations with height $1$ satisfy $\triple{p,\disjC,p}$.

It can also be verified that $\R_{\min F}$ is a model of $F$: all the $\rhodfbot$-interpretations with finite height satisfy the strict part $F^{str}$. The minimal interpretations that do not satisfy  $\triple{b,\disjC,b}$ have height $0$ and satisfy $F^s$, and consequently satisfy $\triple{b,\subclass, f}$. Therefore, the defeasible triple $\dtriple{b,\subclass, f}$ is satisfied  by $\R_{\min F}$.

Note that there cannot be a  model of $F$ that is preferred to $\R_{\min F}$. If we move any $\rhodfbot$-interpretation from rank $1$ to rank $0$, the resulting model would not satisfy $\dtriple{b,\subclass,f}$ anymore: every interpretation in rank $1$ does not satisfy neither $\triple{b,\disjC,b}$ nor $\triple{b,\subclass,f}$, hence, by moving one of them to rank $0$ we obtain an interpretation $\R$ that would not satisfy $\dtriple{b,\subclass, f}$, since we would have in $\cmin(b,\R)$ a $\rhodfbot$-interpretation not satisfying $\triple{b,\subclass,f}$. Instead, if we move any $\rhodfbot$-interpretation from rank $\infty$ to any finite rank, the resulting model would not satisfy $F^{str}$.

Being a minimal model of $F$, $\R_{\min F}$ satisfies the presumption of typicality. To check this, it suffices to determine what we can derive about sparrows: since sparrows do not have any exceptional property, they should inherit all the typical properties of birds, and we should be able to derive that they presumably fly. We need to check the $\rhodfbot$ interpretations in $\cmin(s,\R_{\min F})$, that must be in rank $0$, since $F^s\not\deriv\triple{s,\disjC,s}$. In particular, the $\rhodfbot$ interpretations at rank $0$ satisfy $F^s$ or $F^{str}\cup\{\triple{b,\disjC,b}\}$, and the interpretations in $\cmin(s,\R_{\min F})$ must be among those satisfying $F^s$, since $F^{str}\cup\{\triple{b,\disjC,b}\}\deriv \triple{s,\disjC,s}$.
This implies that all interpretations in $\cmin(s,\R_{\min F})$ satisfy $\triple{s,\subclass,b}$ and $\triple{b,\subclass,f}$, that is, they satisfy $\triple{s,\subclass,f}$. Consequently, according to Definition \ref{defsat}, $\R_{\min F}\rdfsat \dtriple{s,\subclass,f}$, that is, $F\minentail \dtriple{s,\subclass,f}$, as desired.
\qed
\end{example}

\subsection{Exceptionality} \label{excep}

\nd Minimal entailment defines the semantics. We next define a decision procedure for it. To do so, we define the notion of \emph{exceptionality}, a reformulation in our context of a property that is fundamental for RC \cite{Lehmann92b}. Informally, a class $t$ (or, respectively, a predicate $t$) is exceptional \wrt~a defeasible graph if there is no typical situation in which $t$ can be populated with some instance. Formally it corresponds to saying that in every ranked model of the graph, all the $\rhodfbot$-interpretations with height $0$ satisfy $\triple{t,\disjC, t}$ (respectively, $\triple{t,\disjP, t}$).

\begin{definition}[Exceptionality]\label{Def_exceptionality}
Let $G$ be a defeasible $\rhodfbot$-graph, $\R=(\M,r)$ be a  ranked model in $\RG$ and let $t$ be a term.
\begin{enumerate}
\item We say that $t$ is \emph{\ssc-exceptional} (resp.~\emph{\ssp-exceptional}) \wrt~$\R$ if for every $\I\in\M$ s.t. $r(\I)=0$, we have that $\I\rdfsat \triple{t,\disjC, t}$ (resp.~$\I\rdfsat \triple{t,\disjP, t}$). 

\item We say that $t$ is \emph{\ssc-exceptional} (resp.~\emph{\ssp-exceptional}) \wrt~$G$ if it is \ssc-exceptional (resp.~\emph{\ssp-exceptional}) \wrt~all $\R\in\RG$.



\end{enumerate}
\end{definition}

\nd In Example \ref{ex_semantics} $p$ is \ssc-exceptional \wrt~$\R_{\min F}$.  It turns out that in order to check exceptionality \wrt~a graph it is sufficient to refer to the minimal model of the graph.

\begin{restatable}{proposition}{propexceptionality}\label{Prop_exceptionality}
A term $t$ is \ssc-exceptional (resp. \ssp-exceptional) \wrt~a defeasible graph $G$ iff it is \ssc-exceptional (resp. \ssp-exceptional) \wrt~$\CG$.
\end{restatable}



\nd Hence, in Example \ref{ex_semantics} $p$ is \ssc-exceptional \wrt~the graph $F$, since it is exceptional \wrt~the minimal model $\R_{\min F}$. We now prove that \emph{a term $t$ is \ssc-exceptional (resp., \ssp-exceptional) \wrt~a defeasible graph $G$ iff $\Gcount\deriv \triple{t,\disjC,t}$ (resp., $\Gcount\deriv \triple{t,\disjP,t}$)}, which will provide us a decision procedure to decide exceptionality. 

In order to prove the above claim, we introduce the notion of proof tree and prove some lemmas beforehand. To start with, we reformulate the classical notion of proof tree for $\rhodfbot$. 

\begin{definition}[$\rhodfbot$ proof tree]\label{def:proof_trees}

A $\rhodfbot$ proof tree is a finite tree in which 
\begin{itemize}
\item each node is a $\rhodfbot$~triple;
\item every node is connected to the node(s) immediately above
through one of the inference rules (1)-(7) presented in Section~\ref{sect:rhodfbot_deductive}.
\end{itemize}
\nd Let $T$ be a $\rhodfbot$~proof tree, $H$ be the set of the triples appearing as top nodes (called \emph{leaves}) and let $\triple{s,p,o}$ be the triple appearing in the unique bottom node (called \emph{root}). Then $T$ is a $\rhodfbot$~proof tree from $H$ to $\triple{s,p,o}$.

\end{definition}

\nd For instance, in Example~\ref{pex} we have a $\rhodfbot$~proof tree from 
$\{\triple{dU,\subclass,yP}, \triple{yP, \subclass, hP}, \triple{dU,\subclass,uhP}$, $\triple{hP,\disjC,uhP}\}$ 
to $\triple{dU,\disjC,dU}$.


By Definitions~\ref{def:derivation} and~\ref{def:proof_trees}, the following proposition is immediate to prove.

\begin{proposition}\label{prop:proof_trees}
Let $G$ be a $\rhodfbot$ graph and $\triple{s,p,o}$ be a $\rhodfbot$ triple. Then $G\deriv\triple{s,p,o}$ iff  there is a $\rhodfbot$ proof tree from $H$ to $\triple{s,p,o}$ for some $H\subseteq G$.
\end{proposition}

\nd As next, we define a depth function on the trees.

\begin{definition}[Immediate subtree and depth function $d$]\label{def:depth}
Let $T$ be a $\rhodfbot$ proof tree. The set of the \emph{immediate subtrees} of $T$, $\mathfrak{T}=\{T_1,\ldots,T_n\}$,  are the trees $T_i$ obtained from $T$ by eliminating the root node of $T$.

The \emph{depth} $d(T) \in \mathbb{N}$ of $T$ is defined inductively the following way:

\begin{itemize}
    \item if $T$ is a single node then $d(T)=0$;
    \item else, $d(T)= 1 + \max\{d(T')\mid T'\in\mathfrak{T}\}$.
\end{itemize}
\end{definition}

\nd Now we prove the following lemmas that will be used to establish that a term $t$ is \ssc-exceptional (resp., \ssp-exceptional) \wrt~a defeasible graph $G$ iff $\Gcount\deriv \triple{t,\disjC,t}$ (resp., $\Gcount\deriv \triple{t,\disjP,t}$) (see Proposition~\ref{prop:exceptionality_2} later on).

\begin{restatable}{lemma}{lemmatriplesproofsc}\label{lemma:triples_proof_sc}
Let $T$ be a $\rhodfbot$ proof tree from $H$ to $\triple{p,\subclass,q}$. Then $T$ contains only triples of the form $\triple{A,\subclass,B}$.
\end{restatable}


 


\begin{restatable}{lemma}{lemmatriplesproofbotc}\label{lemma:triples_proof_bot_c}
Let $T$ be a $\rhodfbot$~proof tree from $H$ to $\triple{p,\disjC,q}$. Then $T$ contains only triples of the form $\triple{A,\subclass,B}$ or $\triple{A,\disjC,B}$.
\end{restatable}

\nd Note that a tree proving $\triple{t,\disjC,t}$ is just a particular case of Lemma~\ref{lemma:triples_proof_bot_c}.

\begin{restatable}{lemma}{lemmaminimalinterpretations}\label{lemma:minimal_interpretations}
Let $G$ be a defeasible $\rhodfbot$-graph, $\R=(\M,r)$ be a  ranked model in $\RG$ and $\dtriple{p,\subclass,q}\in \Gdef$. For every $\I\in\M$ s.t. $r(\I)=0$, either $\I\rdfsat\triple{p,\subclass,q}$ or $\I\rdfsat\triple{p,\disjC,p}$.

Analogously, for every $\dtriple{p,\spp,q}\in \Gdef$ and every $\I\in\M$ s.t. $r(\I)=0$, either $\I\rdfsat\triple{p,\spp,q}$  or $\I\rdfsat\triple{p,\disjP,p}$.
\end{restatable}




\nd We can extend the above lemma to derived subclass and subproperty triples.

\begin{restatable}{lemma}{lemmaproofsubclass}\label{lemma:proof_subclass}
Let $G$ be a defeasible $\rhodfbot$-graph, $\R=(\M,r)$ be a  ranked model in $\RG$ and let $\Gcount \deriv \triple{p,\subclass,q}$ for some terms $p,q$. For every $\I\in\M$ s.t. $r(\I)=0$, either $\I\rdfsat\triple{p,\subclass,q}$ or $\I\rdfsat\triple{p,\disjC,p}$.

Analogously, if $\Gcount \deriv \triple{p,\spp,q}$  for some terms $p,q$, then for every $\I\in\M$ s.t. $r(\I)=0$, either $\I\rdfsat\triple{p,\spp,q}$ or $\I\rdfsat\triple{p,\disjP,p}$.
\end{restatable}

\begin{restatable}{lemma}{lemmacexceptionality}\label{lemma:c-exceptionality}
Let  $G$ be a defeasible graph, $p$ and $q$ terms, and let $\CG=(\M,r)$ be $G$'s minimal model. If $\Gcount\deriv \triple{p,\disjC,q}$ then $\I\rdfsat \triple{p,\disjC,q}$ for every $\I\in\M$ s.t. $r(\I)=0$.
\end{restatable}

\nd The following is an immediate corollary of Lemma \ref{lemma:c-exceptionality}.

\begin{corollary}\label{coroll:c-exceptionality}
For any defeasible graph $G$ and any term $t$, if $\Gcount\deriv \triple{t,\disjC,t}$ then  $t$ is \ssc-exceptional w.r.t. $G$.
\end{corollary}

\nd Now we prove the analogous result for \ssp-exceptionality.

\begin{restatable}{lemma}{lemmapexceptionality}\label{lemma:p-exceptionality}
Let  $G$ be a defeasible graph, $p,q$ be any pair of terms, and let $\CG=(\M,r)$ be $G$'s minimal model. If $\Gcount\deriv \triple{p,\disjP,q}$ then $\I\rdfsat \triple{p,\disjP,q}$ for every $\I\in\M$ s.t. $r(\I)=0$.
\end{restatable}

\nd An immediate corollary of Lemma \ref{lemma:p-exceptionality} is the following.

\begin{corollary}\label{coroll:p-exceptionality}
For any defeasible graph $G$ and any term $t$, if $\Gcount\deriv \triple{t,\disjP,t}$ then  $t$ is \ssp-exceptional w.r.t. $G$.
\end{corollary}

\nd Now we are ready to prove (see appendix) the main proposition of this section.

\begin{restatable}{proposition}{propexceptionalitysecond}\label{prop:exceptionality_2}
A term $t$ is \ssc-exceptional (resp., \ssp-exceptional) w.r.t. a defeasible graph $G$ iff $\Gcount\deriv \triple{t,\disjC,t}$ (resp., $\Gcount\deriv \triple{t,\disjP,t}$).
\end{restatable}

\nd Proposition~\ref{prop:exceptionality_2} gives us a correct and complete correspondence between the semantic notions of \ssc-exceptionality and \ssp-exceptionality 
with the $\rhodfbot$ decision procedure $\deriv$. This correspondence allows us then to compute all the  \ssc-exceptional triples and all the \ssp-exceptional ones, as illustrated by the procedures $\mathtt{ExceptionalC}(G)$ and $\mathtt{ExceptionalP}(G)$, respectively, where the notion of exceptional triple is defined in the obvious way:

\begin{definition}[Exceptional triple]\label{Def_exceptionalitytriple}
Let $G$ be a defeasible $\rhodfbot$-graph. 
We say that a defeasible triple $\dtriple{p,\subc,q}\in \Gdef$ (resp.~$\dtriple{p,\subp,q}\in \Gdef$) is \emph{\ssc-exceptional} (resp.~\emph{\ssp-exceptional}) \wrt~$G$ if $p$ is \ssc-exceptional  (resp.~\emph{\ssp-exceptional}) \wrt~$G$.
\end{definition}

\floatname{algorithm}{Procedure}
\renewcommand{\thealgorithm}{}

\begin{algorithm}[t]\caption{$\mathtt{ExceptionalC}(G)$}\label{proc:excC}

\algorithmicrequire\  Defeasible graph $G=\Gclass\cup\Gdef$ 

\algorithmicensure\ Set $\epsilon^\ssc(G)$ of \ssc-exceptional triples \wrt~$G$

\begin{algorithmic}[1]
\STATE $\epsilon^\ssc(G)\assign\emptyset$
\FORALL{$\dtriple{p,\subc,q}\in \Gdef$}
    \IF{$\Gcount\deriv\triple{p,\disjC,p}$}
        \STATE $\epsilon^\ssc(G)\assign\epsilon^\ssc(G)\cup\{\dtriple{p,\subc,q}\}$
    \ENDIF    
\ENDFOR
\STATE \textbf{return}\ $\epsilon^\ssc(G)$

\end{algorithmic}
\end{algorithm}

\begin{algorithm}\caption{$\mathtt{ExceptionalP}(G)$}\label{proc:excP}

\algorithmicrequire\  Defeasible graph $G=\Gclass\cup\Gdef$ 

\algorithmicensure\ Set $\epsilon^\ssp(G)$ of \ssp-exceptional triples \wrt~$G$
\begin{algorithmic}[1]
\STATE $\epsilon^\ssp(G)\assign\emptyset$
\FORALL{$\dtriple{p,\subp,q}\in \Gdef$}
    \IF{$\Gcount\deriv\triple{p,\disjP,p}$}
        \STATE $\epsilon^\ssp(G)\assign\epsilon^\ssc(G)\cup\{\dtriple{p,\subp,q}\}$
    \ENDIF    
\ENDFOR
\STATE \textbf{return}\ $\epsilon^\ssp(G)$

\end{algorithmic}
\end{algorithm}

\nd Procedures $\mathtt{ExceptionalC}(G)$ and $\mathtt{ExceptionalP}(G)$ correctly model exceptionality, as proved by the following  immediate corollary of Proposition \ref{prop:exceptionality_2}.

\begin{corollary}\label{coroll:exceptionality_funct}
Given a defeasible graph $G$ and a defeasible triple $\dtriple{p,\subc,q}\in\Gdef$ (resp., $\dtriple{p,\subp,q}\in\Gdef$), $\dtriple{p,\subc,q}\in\epsilon^\ssc(G)$ (resp., $\dtriple{p,\subp,q}\in\epsilon^\ssp(G)$) iff it is \ssc-exceptional (resp. \ssp-exceptional) \wrt $G$.
\end{corollary}

\begin{example}[Running example cont.]\label{pexA}
Like in Example~\ref{pex}, it is easily verified that, besides $\Gcount\deriv\triple{dU,\disjC, dU}$ and 
$\Gcount\deriv\triple{uDC,\disjP, uDC}$, we also have  $\Gcount \deriv  \triple{cDU,\disjC, cDU}$.~\footnote{$uDC \mapsto \mathit{usesDrugControlled}$, $cDU \mapsto \mathit{controlledDrugUser}$ and $hDI \mapsto \mathit{hasDrugIndependence}$.}
As a consequence, 
\begin{eqnarray*}
\mathtt{ExceptionalC}(G) & = & \{ \dtriple{dU,\subclass,uhP}, \dtriple{dU,\subclass, yP}, \dtriple{cDU, \subclass, hP} \} \\
\mathtt{ExceptionalP}(G) & = & \{ \dtriple{uDC,\spp,hDI} \} \ .
\end{eqnarray*}
\qed
\end{example}

\subsection{The Ranking Procedure} \label{ranking}

\nd  Iteratively applied, the notions of \ssc-exceptionality and \ssp-exceptionality allow us to associate to every term, \ie to every defeasible triple,  a rank value \wrt~a defeasible graph $G$. Specifically, we  introduce a ranking procedure, called $\mathtt{Ranking}(G)$,  that orders  the defeasible information in $\Gdef$ into a sequence $\drnk_0,\ldots,\drnk_n,\drnk_\infty$ of sets $\drnk_i$ of defeasible triples, with $n\geq 0$ and $\drnk_\infty$ possibly empty. The procedure is shown below.
 \begin{algorithm}\caption{$\mathtt{Ranking}(G)$}\label{proc_rank}

\algorithmicrequire\  Defeasible graph $G=\Gclass\cup\Gdef$ 

\algorithmicensure\ ranking $\rnk(G)=\{\drnk_0,\ldots,\drnk_n,\drnk_\infty\}$

\begin{algorithmic}[1]
\STATE $\drnk_0\assign\Gdef$
\STATE $i\assign 0$
\REPEAT
    \STATE $\drnk_{i+1}\assign\mathtt{ExceptionalC}(\Gclass\cup\drnk_{i})\cup\mathtt{ExceptionalP}(\Gclass\cup\drnk_{i})$

    \STATE $i\assign i+1$
\UNTIL{$\drnk_i =\drnk_{i+1}$}
\STATE $\drnk_{\infty}\assign\drnk_{i}$
\STATE $\rnk(G)\assign \{\drnk_0,\ldots,\drnk_{i-1},\drnk_\infty\}$
\RETURN $\rnk(G)$

\end{algorithmic}
\end{algorithm}

\nd The ranking procedure is built on top of the $\mathtt{ExceptionalC}$ and $\mathtt{ExceptionalP}$ procedures, using  \rhodfbot decision steps only.

\begin{example}[Running example cont.]\label{pexB}
Consider Example~\ref{pexA}. Let us order $\Gdef$ by running $\mathtt{Ranking}(G)$. So, consider
$\drnk_0\assign\Gdef$. From Example~\ref{pexA} we immediately have
\begin{eqnarray*}
\mathtt{ExceptionalC}(\Gclass\cup \drnk_0) & \assign &  \mathtt{ExceptionalC}(G)\\
\mathtt{ExceptionalP}(\Gclass\cup \drnk_0) & \assign & \mathtt{ExceptionalP}(G)  \\
\drnk_1 & \assign &  \{ \dtriple{dU,\subclass,uhP}, \dtriple{dU,\subclass, yP}, \dtriple{cDU, \subclass, hP}, \dtriple{uDC,\spp,hDI} \} \ .
\end{eqnarray*}
\nd Now, it can be verified that \wrt~$\Gclass\cup \drnk_1$, $dU$ and $uDC$ are not exceptional, while $cDU$ is, as
\begin{eqnarray*}
\Gclass\cup (\drnk_1)^s & \not \deriv & \triple{dU,\disjC, dU} \\
\Gclass\cup (\drnk_1)^s & \not \deriv & \triple{uDC,\disjP, uDC} \\
\Gclass\cup (\drnk_1)^s & \deriv & \triple{cDU,\disjC, cDU} \ .
\end{eqnarray*}
\nd Therefore,
\begin{eqnarray*}
\drnk_2& \assign &  \{\dtriple{cDU, \subclass, hP} \} \ .
\end{eqnarray*}
\nd Finally,  it is easy to check that
\begin{eqnarray*}
\drnk_3 & \assign &  \emptyset 
\end{eqnarray*}
\nd that is, 
\begin{eqnarray*}
\drnk_\infty & \assign &  \emptyset 
\end{eqnarray*}
\nd and, thus,
\begin{eqnarray*}
\rnk(G) & \assign & \{\drnk_0, \drnk_1, \drnk_2, \drnk_\infty\} \ .
\end{eqnarray*}
\qed
\end{example}


\nd The following example shows an example of a triple that turns out to have infinite rank.

 \begin{example}\label{ex_inf}
 Let $L$ be the graph $\{\triple{p,\dom,r}, \triple{q,\dom,t}, \triple{r,\disjC, t}, \dtriple{q,\subc,p}\}$. The only defeasible triple is $\dtriple{q,\subc,p}$, hence $\drnk_0=\{\dtriple{q,\subc,p}\}$. Applying the procedure $\mathtt{ExceptionalC(L)}$, from the graph $L^s$ we conclude the triple $\triple{q,\disjP,q}$ with the following derivation:

 \begin{prooftree}
 \AxiomC{$\triple{p,\dom,r}$}
 \AxiomC{$\triple{q,\dom,t}$}
 \AxiomC{$\triple{r,\disjC,t}$}
 \RightLabel{(7a')}\TrinaryInfC{$\triple{p,\disjP,q}$}
 \AxiomC{$\triple{q,\subp,p}$}
 \RightLabel{(6b)}\BinaryInfC{$(q,\disjP,q)$}
 \end{prooftree}

\nd  Hence we have $\drnk_{1}=\drnk_0$, that is, we have a fixed point, and we can conclude $\drnk_{\infty}=\drnk_0=\{\dtriple{q,\subc,p}\}$. 

\qed
\end{example}


\nd Now we prove that the ranking procedure $\mathtt{Ranking}(G)$ correctly mirrors the ranking of the defeasible information  \wrt~the height functions $\hc$ and $\hp$. Specifically, we want to show  (see Proposition~\ref{Prop_ranking_main} later on) that
 \begin{itemize}
     \item for $i<n$, $\dtriple{p,\subc,q}\in\D_i\setminus \D_{i+1}$ iff $\hc(p)=i$;
     \item for $i=n$, $\dtriple{p,\subc,q}\in\D_n\setminus \D_{\infty}$ iff $\hc(p)=n$,
 \end{itemize}
 
 \nd and an analogous result \wrt~$\hp$.
 

To do so, we need to introduce some preliminary constructions and lemmas. 
To start with, let us note that the information in a graph $G=\Gclass\cup\Gdef$ is ranked from the semantical point of view by the minimal model $\CG=(\M,r)$ and the height functions $\hc$ and $\hp$ defined on it. Let $\CG^i=(\M^i,r^i)$ be the submodel of $\CG$ obtained by eliminating all the $\rhodfbot$ interpretations in $\M$ whose height is strictly less than $i$ ($i\geq 0$). Specifically, given a graph $G=\Gclass\cup\Gdef$ and its minimal model $\CG=(\M,r)$, $\CG^i=(\M^i,r^i)$ is defined as follows:
\begin{itemize}
    \item $\M^i\assign\{\I\in\M\mid r(\I)\geq i\}$;
    \item $r^i(\I)\assign r(\I)-i$, for every $\I\in\M^i$.
\end{itemize}
\nd That is, $\CG^i$ is obtained from $\CG$ by eliminating all the interpretations that have a rank of $i-1$ or less.  We next show that the interpretation $\CG^i$ models the defeasible information in the graph that has a higher level of exceptionality: the higher the value of $i$, the higher the level of exceptionality. That is, given a defeasible graph $G=\Gclass\cup\Gdef$, let 
\begin{equation}\label{Eq_Gi}
  G^i\assign\Gclass\cup\Gdef_i \ ,
\end{equation}
\nd where
\begin{equation}\label{Eq_GdefCi}
  \Gdef_i\assign \{\dtriple{p,\subc,q}\in\Gdef\mid \hc(p)\geq i\} \cup \{\dtriple{p,\subp,q}\in\Gdef\mid \hp(p)\geq i\} \ .
\end{equation}
\nd Then the following  proves that $\CG^i$ is indeed a model of $G^i$.

\begin{restatable}{lemma}{LemmaCGi}\label{Lemma_CGi}
Let $G=\Gclass\cup\Gdef$ be a defeasible graph and $\CG=(\M,r)$ its minimal ranked model. Then $\CG^i=(\M^i,r^i)$ is a model of the subgraph $G^i=\Gclass\cup\Gdef_i$.
\end{restatable}




\nd Now, it turns out that the minimal model for $G^i$, that is built using the set of $\rhodfbot$-interpretations $\M$, can easily be defined by extending $\CG^i$. That is, let 
$\R^*_i=(\M,r^*_i)$ be a ranked interpretation where $r^*_i$ is defined as 
\[
    r^*_i(\I)=\left\{
    \begin{array}{ll}
        r^i(\I)  & \text{if } \I\in \M^i\\
         0  & \text{otherwise} \ .
    \end{array}\right.
\]

\nd The following holds.
\begin{restatable}{lemma}{LemmaminimalmodelCGi}\label{Lemma_minimal_model_CGi}
Given a defeasible graph $G$, $\R^*_i$ is the minimal model of the subgraph $G^i$.
\end{restatable}

\nd The following lemma connects the height of a term with the notion of exceptionality in the models $\R^*_i$.


\begin{restatable}{lemma}{Lemmarenkexceptionality}\label{Lemma_renk_exceptionality} Let $G=\Gclass\cup\Gdef$ be a defeasible graph and let $\CG$ be its minimal model, with $h(\CG)=n$.  
For every $i\leq n$ and 
term $p$ s.t. $\hc(p)\geq i$ (resp., $\hp(p)\geq i$) $p$ is \ssc-exceptional (resp., \ssp-exceptional) w.r.t.~$\CG^i=(\M^i,r^i)$ iff it is \ssc-exceptional (resp., \ssp-exceptional) w.r.t.~$\R^*_i=(\M,r^*_i)$.
\end{restatable}

\nd The following lemma connects the height to the computation of exceptionality.

\begin{restatable}{lemma}{Lemmarankexcept}\label{Lemma_rank_except}
Let $G=\Gclass\cup\Gdef$ be a defeasible graph, and $\CG=(\M,r)$ its minimal model, with $h(\CG)=n$,  and $\dtriple{p,\subc,q}\in\Gdef$. Then, 

\begin{itemize}
    \item for every $i< n$, $\hc(p)\geq i+1$ iff $\dtriple{p,\subc,q} \in \mathtt{ExceptionalC}(G^i)$;
    \item for $i= n$, $\hc(p)=\infty$ iff $\dtriple{p,\subc,q}  \in \mathtt{ExceptionalC}(G^i)$.
\end{itemize}
\nd Analogously, let $\dtriple{p,\subp,q}\in\Gdef$. Then, 
\begin{itemize}
    \item for every $i< n$, $\hp(p)\geq i+1$ iff $ \dtriple{p,\subp,q}\in \mathtt{ExceptionalP}(G^i)$;
    \item for $i= n$, $\hp(p)=\infty$ iff $\dtriple{p,\subp,q} \in \mathtt{ExceptionalP}(G^i)$.
\end{itemize}
\end{restatable}

\nd We now move on to prove the correspondence between the ranking procedure and the height (the semantic ranking of the defeasible information). 

Given a defeasible graph $G=\Gclass\cup\Gdef$, $G^\D_i$ ($i\geq 0$) is the subgraph of $G$ defined as follows:
\begin{equation}\label{Def_Gssci}
    G^\D_i\assign \Gclass\cup \drnk_i \ ,
\end{equation}
\nd where $\drnk_i$ is in the output of $\mathtt{Ranking}(G)$, \ie~$\drnk_i \in \rnk(G)$. Our objective now is to prove that $G^i=G^\D_i$.
To prove that for every $i$, $G^i=G^\D_i$, it is sufficient to prove that for every rank value $i\neq\infty$ a term $t$ is exceptional w.r.t. $G^\D_i$ iff it is exceptional w.r.t. $\CG^i$.







The following can be shown.

\begin{restatable}{lemma}{Lemmaexceptranking}\label{Lemma_except_ranking}
Let $G$ be a defeasible graph, with $n$ being the height of its minimal model.  Then, for every $i\leq n$, $G^i=G^\D_i$.
\end{restatable}




\nd Now we can state the main proposition for our ranking procedure.
 
 \begin{proposition}\label{Prop_ranking_main}
 Let $G=\Gclass\cup\Gdef$ be a defeasible graph, $\dtriple{p,\subc,q}$ (resp., $\dtriple{p,\subp,q}$) be in $\Gdef$, and $\rnk(G)= \{\drnk_0,\ldots,\drnk_{n},\drnk_\infty\}$ be the ranking obtained by $\mathtt{Ranking}(G)$. The following statements hold:
 \begin{itemize}
     \item for $i<n$, $\dtriple{p,\subc,q}\in\D_i\setminus \D_{i+1}$ iff $\hc(p)=i$;
     \item for $i=n$, $\dtriple{p,\subc,q}\in\D_n\setminus \D_{\infty}$ iff $\hc(p)=n$.
 \end{itemize}
 \end{proposition}
 \begin{proof}
Immediate consequence of Lemma \ref{Lemma_except_ranking}.
 \end{proof}

\nd Another consequence of Lemma \ref{Lemma_except_ranking} is the following corollary, that will be useful later on.

 \begin{corollary}\label{Coroll_ranking_main}
  Let $G=\Gclass\cup\Gdef$ be a defeasible graph, with $\rnk(G)= \{\drnk_0,\ldots,\drnk_{n},\drnk_\infty\}$ being the ranking obtained by $\mathtt{Ranking}(G)$, and let $p$ be a term. The following statements hold:
  \begin{itemize}
      \item $\hc(p)=0$ \iff $G^\D_0\not\deriv\triple{p,\disjC,p}$;
      \item for every $i$, $0<i\leq n$, $\hc(p)=i$ \iff $G^\D_{i-1}\deriv\triple{p,\disjC,p}$ and  $G^\D_i\not\deriv\triple{p,\disjC,p}$;
      \item $\hc(p)=\infty$ \iff $G^\D_n\deriv\triple{p,\disjC,p}$.
  \end{itemize}
\nd Analogously,
  \begin{itemize}
      \item $\hp(p)=0$ \iff $G^\D_0\not\deriv\triple{p,\disjP,p}$;
      \item for every $i$, $0<i\leq n$, $\hp(p)=i$ \iff $G^\D_{i-1}\deriv\triple{p,\disjP,p}$ and  $G^\D_i\not\deriv\triple{p,\disjP,p}$;
      \item $\hc(p)=\infty$ \iff $G^\D_n\deriv\triple{p,\disjP,p}$.
  \end{itemize}
 \end{corollary}
\begin{proof}
Immediate from Proposition \ref{prop:exceptionality_2}, Lemma \ref{Lemma_rank_except} and Lemma \ref{Lemma_except_ranking}.
\end{proof}

\subsection{Decision Procedures for Defeasible $\rhodf$} \label{decproc}


\nd In this section we present our main decision procedures. That is, given  triples $\dtriple{p,\subc,q}$, $\dtriple{p,\subp,q}$ and $\triple{s,p,o}$ as queries, the procedures below  decide whether they are or not minimally entailed by a defeasible graph $G$.


\begin{remark}\label{rnkpre}
Given a fixed defeasible graph $G$,  we  assume that its ranking 
$\rnk(G) =\{\drnk_0,\ldots,\drnk_n,\drnk_\infty\} = \mathtt{Ranking}(G)$ has already been computed .
\end{remark}

\nd We start with the case of \rhodfbot-triples of the form $\triple{s,p,o}$. The procedure $\mathtt{StrictMinEntailment}$ below decides whether 
$G\minentail\triple{s,p,o}$ holds. Essentially, given $G$, we take the strict part $\Gclass$ to which we add  disjointness triples related to defeasible ones with infinite rank and then test for $\rhodfbot$ entailment.
 
\begin{algorithm}\caption{$\mathtt{StrictMinEntailment}(G, \rnk(G), \triple{s,p,o})$}\label{proc_decision_class}

\algorithmicrequire\  Graph $G=\Gclass\cup\Gdef$,  ranking $\rnk(G)=\{\drnk_0,\ldots,\drnk_n,\drnk_\infty\}$, a \rhodfbot-triple $\triple{s,p,o}$

\algorithmicensure\ $\mathtt{true}$ if $G\minentail\triple{s,p,o}$; $\mathtt{false}$ otherwise

\begin{algorithmic}[1]

\STATE $G'\assign \Gclass\cup\{\triple{p,\disjC,p}\mid \dtriple{p,\subc,q}\in \drnk_\infty\}\cup\{\triple{p,\disjP,p}\mid \dtriple{p,\subp,q}\in \drnk_\infty\}$

\RETURN $G'\deriv \triple{s,p,o}$

\end{algorithmic}
\end{algorithm}

%
%
%
%
%


\nd The following lemma can be proved, which motivates the construction of graph $G'$ in the procedure $\mathtt{StrictMinEntailment}$.

\begin{restatable}{lemma}{propdecisionclass}\label{prop_decision_class}
Let $G=\Gclass\cup\Gdef$ be a defeasible graph, $\rnk(G)=\{\drnk_0,\ldots,\drnk_n,\drnk_\infty\}$ its ranking, and $\CG=(\M,r)$ its minimal model, with $\I\in\M$. Then $\I\in\MN$ iff $\I\rdfsat \Gclass\cup\{\triple{p,\disjC,p}\mid \dtriple{p,\subc,q}\in \drnk_\infty\}\cup\{\triple{p,\disjP,p}\mid \dtriple{p,\subp,q}\in \drnk_\infty\}$.
\end{restatable}

\nd The following theorem establishes correctness and completeness of the $\mathtt{StrictMinEntailment}$ procedure.

\begin{theorem}\label{coroll_decision_class}
Consider a defeasible graph $G$ and a \rhodfbot-triple $\triple{s,p,o}$. Then 
\[
G\minentail\triple{s,p,o} \mbox{ iff }  \mathtt{StrictMinEntailment}(G, \rnk(G), \triple{s,p,o}) \ .
\]
\end{theorem}

\begin{proof}
By Lemma~\ref{prop_decision_class} and  Theorem~\ref{Th:soundcomplete} we have that $G\minentail\triple{s,p,o}$ \iff  $\Gclass\cup\{\triple{p,\disjC,p}\mid \dtriple{p,\subc,q}\in \drnk_\infty\}\cup\{\triple{p,\disjP,p}\mid \dtriple{p,\subp,q}\in \drnk_\infty\}\rdfent\triple{s,p,o}$. Therefore, the procedure 
\\ $\mathtt{StrictMinEntailment}$ is correct and complete.
\end{proof}

\begin{example}[Running example cont.]\label{pexC}
Consider Example~\ref{pexB}. We have seen that $\rnk(G) = \{\drnk_0, \drnk_1$, $\drnk_2, \drnk_\infty\}$, with $\drnk_\infty =  \emptyset$. Let us show now that ``tom is a person". To do so, let us show that
\[
\mathtt{StrictMinEntailment}(G, \rnk(G), \triple{\mathit{tom}, \type, \mathit{person}}) = \mathtt{true} \ .
\]
As $\drnk_\infty = \emptyset$, we have that   $G' = \Gclass$. Now, in Example~\ref{exrdfsD} we have seen that $\Gclass\deriv \triple{\mathit{tom}, \type, person}$. Therefore, $\mathtt{StrictMinEntailment}(G, \rnk(G), \triple{\mathit{tom}, \type, \mathit{person}})$ returns $\mathtt{true}$ and by Theorem~\ref{coroll_decision_class} we can conclude that $G\minentail\triple{\mathit{tom}, \type, \mathit{person}}$.
\qed
\end{example}

\begin{algorithm}\caption{$\mathtt{DefMinEntailmentC}(G, \rnk(G), \dtriple{p,\subc,q})$}\label{proc_decision_def_C}

\algorithmicrequire\  Graph $G=\Gclass\cup\Gdef$,  ranking $\rnk(G)=\{\drnk_0,\ldots,\drnk_n,\drnk_\infty\}$, defeasible triple $\dtriple{p,\subc,q}$

\algorithmicensure\ $\mathtt{true}$ if $G\minentail\dtriple{p,\subc,q}$; $\mathtt{false}$ otherwise

\begin{algorithmic}[1]

\STATE $i\assign 0$
\STATE $\drnk_{n+1}\assign \drnk_{\infty}$
\REPEAT
\IF{$i\leq n$}
    \STATE $G'\assign \Gclass \cup (\drnk_{i})^s$ 
    \STATE $j\assign i$
    \STATE $i\assign i+1$

\ELSE 
    \RETURN{$\mathtt{true}$}
\ENDIF 
\UNTIL{$G'\not\deriv\triple{p,\disjC,p}$}
\STATE $\drnkc^p\assign \{\dtriple{r,\subc, s}\mid \dtriple{r,\subc, s}\in \drnk_j\setminus \drnk_{j+1}\}$
\RETURN{$\Gclass\cup (\drnkc^p)^s\deriv \triple{p,\subc,q}$}

\end{algorithmic}
\end{algorithm}
 
\nd We next consider triples of the form $\dtriple{p,\subc, q}$. The decision procedure $\mathtt{DefMinEntailmentC}$  decides whether $G\minentail\dtriple{p,\subc,q}$ holds  (we will later consider also an analogous procedure for  triples of form $\dtriple{p,\subp, q}$).
Roughly, given a graph $G=\Gclass\cup\Gdef$, the ranking $\rnk(G)=\{\drnk_0,\ldots,\drnk_n,\drnk_\infty\}$ of defeasible triples in $\Gdef$ and a query triple $\dtriple{p,\subc, q}$, we first search for the lowest ranked $\drnk_j$ (smallest $j$) such that $p$ is not  \ssc-exceptional \wrt~$\Gclass \cup (\drnk_{j})^s$, \ie~$\Gclass \cup (\drnk_{j})^s\not\deriv\triple{p,\disjC,p}$. Then, from such a set $\drnk_j$ we remove all higher ranked defeasible triples, and we call $\drnkc^p$ the resulting set. We conclude by checking whether the query triple $\dtriple{p,\subc,q}$ is entailed by  $\Gclass\cup (\drnkc^p)^s$, \ie~$\Gclass\cup (\drnkc^p)^s\deriv \triple{p,\subc,q}$.
 
To establish the correctness and completeness of procedure $\mathtt{DefMinEntailmentC}$ we prove some lemmas beforehand. Let us recall that, by Lemma \ref{lemma:triples_proof_sc}, every proof tree with a triple $\triple{p,\subc,q}$ as root contains only triples of the form $\triple{A,\subc,B}$. 
Using such a lemma, the following can be proven.

\begin{restatable}{lemma}{lemmatreescdisjC}\label{lemma_tree_sc_disjC}
Let $T$ be a proof tree from a graph $H$ to a triple $\triple{p,\subc,q}$, and let $\I$ be a model of $H$. If\ $\I\rdfsat \triple{s,\disjC,s}$ for some triple $\triple{s,\subc,o}\in H$, then $\I\rdfsat \triple{p,\disjC,p}$.
\end{restatable}

\begin{restatable}{lemma}{lemmaalgminentCcorrect}\label{lemma_alg_minentC_correct}
Let $G=\Gclass\cup\Gdef$ be a defeasible graph, and let $\rnk(G)=\{\drnk_0,\ldots,\drnk_n,\drnk_\infty\}$ be its ranking. For any pair of terms $p,q$ s.t. $\hc(p)\leq n$, 
\[
G\minentail\dtriple{p,\subc,q}\text{ iff } \Gclass\cup(\drnkc^p)^s\deriv \triple{p,\subc,q} \ .
\]
\nd where $\drnkc^p$ is defined as in the  $\mathtt{DefMinEntailmentC}$ procedure.
\end{restatable}

\nd The following theorem establishes correctness and completeness of the $\mathtt{DefMinEntailmentC}$ procedure.

\begin{restatable}{theorem}{theoralgminentCcorrect}\label{theor_alg_minentC_correct}
Let $G=\Gclass\cup\Gdef$ be a defeasible graph and let $\dtriple{p,\subc,q}$ be a defeasible triple. Then,   
\[
G\minentail\dtriple{p,\subc,q} \text{ iff } \mathtt{DefMinEntailmentC}(G, \rnk(G), \dtriple{p,\subc,q}) \ .
\]
\end{restatable}



\begin{example}[Running example cont.]\label{pexD}
Consider again Example~\ref{pexB}. 
Let us show that ``young drug users are usually unhappy" (\cf~statement (8) in Example~\ref{ex01}). To do so, let us show that
\[
\mathtt{DefMinEntailmentC}(G, \rnk(G), \dtriple{yDU, \subclass, uhP}) = \mathtt{true} \ .
\]

\nd For $i=0$, we have $G'=\Gcount$, and $\Gcount\deriv \triple{yDU, \disjC, yDU}$, since $\Gcount\deriv \triple{dU, \disjC, dU}$, that together with $\triple{yDU, \subclass, dU}$ implies 
$\triple{yDU, \disjC, yDU}$ by the rule $(EmptySP')$.

For $i=1$, we have $G'=G\cup(\drnk_1)^s$, and $G'\not\deriv \triple{yDU, \disjC, yDU}$, since $G'\not\deriv \triple{dU, \disjC, dU}$. Therefore, according to line 11 we have
\begin{eqnarray*}
\drnkc^{p} & \assign & \drnk_1\setminus\drnk_2 = \{\dtriple{dU, \subclass, uhP}, \dtriple{dU, \subclass, yP}, \dtriple{uDC, \spp, hDI}\} \ .
\end{eqnarray*}

\nd Finally, it is easy to check that:
\[
\Gclass\cup (\drnkc^{p})^s\deriv \triple{yDU, \subclass, uhP}
\]
\nd since $\triple{dU, \subclass, uhP}\cup \triple{yDU, \subclass, dU}\subseteq \Gclass\cup (\drnkc^{yDU})^s$.
Therefore, 
\[
\mathtt{DefMinEntailmentC}(G, \rnk(G), \dtriple{yDU, \subclass, uhP}) = \mathtt{true}
\]
\nd and by Theorem~\ref{theor_alg_minentC_correct} we can conclude that $G\minentail\dtriple{yDU, \subclass, uhP}$.

Please note that $G\not\minentail\dtriple{yDU, \subclass, hP}$ as $\triple{yP, \subclass, hP}\notin \Gclass\cup (\drnkc^p)^s$.
\qed
\end{example}
 
\begin{algorithm}\caption{$\mathtt{DefMinEntailmentP}(G, \rnk(G), \dtriple{p,\subp,q})$}\label{proc_decision_def_P}

\algorithmicrequire\  Graph $G=\Gclass\cup\Gdef$,  ranking $\rnk(G)=\{\drnk_0,\ldots,\drnk_n,\drnk_\infty\}$, defeasible triple $\dtriple{p,\subp,q}$

\algorithmicensure\ $\mathtt{true}$ if $G\minentail\dtriple{p,\subp,q}$; $\mathtt{false}$ otherwise

\begin{algorithmic}[1]

\STATE $i\assign 0$
\STATE $\drnk_{n+1}\assign \drnk_{\infty}$
\REPEAT
\IF{$i\leq n$}
    \STATE $G'\assign \Gclass\cup  (\drnk_{i})^s$ 
        \STATE $j\assign i$
    \STATE $i\assign i+1$
\ELSE 
    \RETURN{$\mathtt{true}$}
\ENDIF 
\UNTIL{$G'\not\deriv\triple{p,\disjP,p}$}
\STATE $\drnkp^p\assign \{\dtriple{r,\subp, s}\mid \dtriple{r,\subp, s}\in \drnk_j\setminus \drnk_{j+1}\}$
\RETURN{$\Gclass\cup (\drnkc^p)^s \deriv \triple{p,\subp,q}$}

\end{algorithmic}
\end{algorithm}

\begin{remark}[Drowning problem]\label{drowning}
\nd We take the opportunity, using Example \ref{pexD}, to show that our entailment relation presents the same inferential limit that characterises  Rational Closure, and that motivates the entailment relation we are going to present in Section \ref{din} later on.  By referring to Example~\ref{pexD}, please note that indeed we also have
\[
G\not\minentail\dtriple{dU, \subclass, s}
\]
\nd as $\mathtt{DefMinEntailmentC}(G, \rnk(G), \dtriple{dU, \subclass, s})$ returns $\mathtt{false}$, 
\ie~we cannot derive that ``drug users are usually students" (\cf~statement (10) in Example~\ref{ex01}).

This latter example shows a well-known behaviour of RC: in the case a class is exceptional \wrt~a super-class, it does not inherit any of the typical properties of the super-class. In this specific case, since drug users are exceptional young people (they are not happy), in RC they do not inherit any of the typical properties of young people: in particular, we cannot conclude that drug users are students. This behaviour, called the  \emph{drowning problem} \cite{BenferhatEtAl93}, may not be desirable in some applications and various  RC extensions have been developed to overcome this 
problem~\cite{CasiniStraccia13,CasiniEtAl2014,Lehmann95}. In this work, we will address it in Section~\ref{din} via DINs.
\end{remark}

\nd Eventually, an analogous procedure to $\mathtt{DefMinEntailmentC}$ can be defined for the case of defeasible triples of the form $ \dtriple{p,\subp,q}$, as illustrated by the $\mathtt{DefMinEntailmentP}$ procedure.

The proof that procedure $\mathtt{DefMinEntailmentP}$ is correct and complete \wrt~$\minentail$ proceeds similarly to the one for $\mathtt{DefMinEntailmentC}$. 

Specifically, we first prove the analogous of Lemma~\ref{lemma:triples_proof_sc}.

\begin{lemma}\label{lemma:triples_proof_sp}
Let $T$ be a $\rhodfbot$ proof tree from $H$ to $\triple{p,\subp,q}$. Then $T$ contains only triples of the form $\triple{A,\subp,B}$.
\end{lemma}
\begin{proof}
The proof is similar to the proof of Lemma \ref{lemma:triples_proof_sc}, we just need to refer to rule (2a) instead of (3a).
\end{proof}

\nd Also for the other propositions the proof is analogous to the correspondent propositions for procedure $\mathtt{DefMinEntailmentC}$. It suffices to change every instance of $\subc$ with $\subp$, $\disjC$ with $\disjP$, every reference to Lemma \ref{lemma:triples_proof_sc} with Lemma \ref{lemma:triples_proof_sp}, and so on. Specifically, we have 

\begin{lemma}\label{lemma_tree_sc_disjP}
Let $T$ be a proof tree from a graph $H$ to a triple $\triple{p,\subp,q}$, and let $\I$ be a model of $H$. If $\I\rdfsat \triple{s,\disjP,s}$ for some triple $\triple{s,\subp,o}\in H$, 
then $\I\rdfsat \triple{p,\disjP,p}$.
\end{lemma}

\begin{lemma}\label{lemma_alg_minentP_correct}
Let $G=\Gclass\cup\Gdef$ be a defeasible graph, and let $\rnk(G)=\{\drnk_0,\ldots,\drnk_n,\drnk_\infty\}$ be its ranking. For any pair of terms $p,q$ s.t. $\hp(p)\leq n$, 
\[
G\minentail\dtriple{p,\subp,q}\text{ iff } \Gclass\cup (\drnkp^p)^s\deriv \triple{p,\subp,q} \ ,
\]
\nd where $\drnkp^p$ is defined as in the $\mathtt{DefMinEntailmentP}$ procedure.
\end{lemma}

\nd Eventually, we conclude with the following theorem establishing correctness and completeness of the $\mathtt{DefMinEntailmentP}$ procedure.

\begin{theorem}\label{theor_alg_minentP_correct}
Let $G=\Gclass\cup\Gdef$ be a defeasible graph and let $\dtriple{p,\subp,q}$ be a defeasible triple. Then,
\[
G\minentail\dtriple{p,\subp,q} \text{ iff }\mathtt{DefMinEntailmentP}(G, \rnk(G), \dtriple{p,\subp,q}) \ .
\]
\end{theorem}


\begin{example}[Running example cont.]\label{pexE}
Consider again Example~\ref{pexB}. Likewise Example~\ref{pexD}, it is not difficult to show that indeed
\[
\mathtt{DefMinEntailmentP}(G, \rnk(G), \dtriple{uDC, \spp, hDI}) = \mathtt{true} \ .
\]
\nd Therefore, by Theorem~\ref{theor_alg_minentP_correct} we can conclude that $G\minentail\dtriple{uDC, \spp, hDI})$.
That is, ``someone that uses  some drug in a controlled way has usually drug independence to that drug".
However, please also note that $G\not \minentail\dtriple{uDC, \spp, hDA})$ instead\footnote{$hDA \mapsto \mathit{hasDrugAddiction}$.} as one may check that
\[
\mathtt{DefMinEntailmentP}(G, \rnk(G), \dtriple{uDC, \spp, hDA}) = \mathtt{false} \ .
\]
\qed
\end{example}

\nd For completeness, we conclude this section with some examples relying on the ``penguin" Example~\ref{ex_semantics}.

\begin{example}\label{ex_decision_1}
Consider the graph $H$, similar to the graph $F$ from Example \ref{ex_semantics}, but with the following changes:
\begin{itemize} 
    \item the triple $\dtriple{b,\subclass,hf}$ has been added, where $hf$ is read as ``having feathers";
    \item the information that penguins do not fly has been made defeasible, that is, the triple $\triple{p,\subclass, e}$ has been substituted by $\dtriple{p,\subclass, e}$;
    \item the triples $\dtriple{pj,\subclass,f}$ and $\triple{pj,\subclass,p}$ have been added, where $pj$ is read as ``penguins with jet-packs".
\end{itemize}

\nd That is, $H=H^{str}\cup H^{def}$, with
\begin{eqnarray*}
H^{str} & = &\{\triple{p,\subclass,b},\triple{s,\subclass,b}, \triple{e,\disjC,f},\triple{pj,\subclass,p}\} \\
H^{def} & = & \{\dtriple{b,\subclass,f},\dtriple{p,\subclass, e}, \dtriple{b,\subclass,hf}, \dtriple{pj,\subclass,f}\} \ .
\end{eqnarray*}
\nd Given $H^s=\{\triple{p,\subclass,b},\triple{s,\subclass,b}, \triple{b,\subclass,f},\triple{p,\subclass, e},\triple{e,\disjC,f}, \triple{b,\subclass,hf}, \triple{pj,\subclass,f},\triple{pj,\subclass,p}\}$, it is easy to check that:
\begin{eqnarray*}
H^s & \not \deriv & \triple{b,\disjC, b} \\
H^s & \not \deriv  & \triple{s,\disjC, s} \\
H^s & \deriv & \triple{p,\disjC, p} \\
H^s & \deriv & \triple{pj,\disjC, pj} \ .
\end{eqnarray*}
\nd Therefore, within procedure $\mathtt{Ranking}(H)$, we have 
\[
\drnk_1=\{\dtriple{p,\subclass, e},  \dtriple{pj,\subclass,f}\} \ .
\]
\nd Since
\begin{eqnarray*}
H^{str}\cup ({\drnk_1)}^s & \not \deriv & \triple{p,\disjC, p} \\
H^{str}\cup ({\drnk_1})^s & \deriv &  \triple{pj,\disjC, pj} \\
\end{eqnarray*}
\nd the procedure $\mathtt{Ranking}(H)$ determines
\[
\drnk_2=\{  \dtriple{pj,\subclass,f}\} \ ,
\]
\nd and terminates with  
\[
\drnk_{\infty}=\emptyset \ ,
\]
\nd as $H^{str}\cup ({\drnk_2})^s \not \deriv \triple{pj,\disjC, pj}$.

Having computed $\mathtt{Ranking}(H)$, we can now check what is minimally entailed. 

At first, we can check that penguins do not fly, that is, we can make the following queries:
\begin{itemize}
    \item[{\bf Triple $\dtriple{p,\subclass,e}$.}]
    We apply the procedure $\mathtt{DefMinEntailmentC}(H,r(H),\dtriple{p,\subclass,e})$ and have:
    $G'\not\deriv\triple{p,\disjC, p}$ for $i=1$, that implies  $\drnk^p=\{\dtriple{p,\subclass,e}\}$.
    $G^{str}\cup (\drnk^p)^s\deriv \triple{p,\subclass,e}$, since $(\drnk^p)^s=\{\triple{p,\subclass,e}\}$, and $\mathtt{DefMinEntailmentC}(H,r(H),\dtriple{p,\subclass,e})$ returns $\mathtt{true}$.
    
    \item[{\bf Triple $\dtriple{p,\subclass,f}$.}]
    We apply the procedure $\mathtt{DefMinEntailmentC}(H,r(H),\dtriple{p,\subclass,f})$ and have:
    $G^{str}\cup (\drnk^p)^s\not \deriv \triple{p,\subclass,f}$, and $\mathtt{DefMinEntailmentC}(H,r(H),\dtriple{p,\subclass,f})$ returns $\mathtt{false}$.
\end{itemize}
\nd The procedure $\mathtt{DefMinEntailmentC}$ allows us to correctly derive that ``penguins do not fly", while it avoids to derive that ``penguins fly". What about the other typical property of birds in our graph, that is, having feathers?
\begin{itemize}
    \item[{\bf Triple $\dtriple{p,\subclass,hf}$.}]
    We apply the procedure $\mathtt{DefMinEntailmentC}(H,r(H),\dtriple{p,\subclass,hf})$  and have:
    $G^{str}\cup (\drnk^p)^s\not \deriv \triple{p,\subclass,hf}$, and 
    \[
    \mathtt{DefMinEntailmentC}(H,r(H),\dtriple{p,\subclass,hf}) = \mathtt{false} \ .
    \]
\end{itemize}
\nd This latter example shows again the drowning problem (see Remark~\ref{drowning}). In this specific case, since penguins are exceptional birds (they do not fly), in RC they do not inherit any of the typical properties of birds: that is, we cannot conclude that penguins have feathers. 

Now we move to check the behaviour of sub-classes that do not show any exceptional behaviour. From our graph we know that sparrows are birds, and we have no information about any unusual property associated to sparrows. As a consequence, reasoning on the base of the principle of `presumption of typicality' (see Section \ref{sect_semantics}), we would like sparrows to inherit all the typical properties of birds. In fact, we have:
\begin{itemize}
    \item[{\bf Triple $\dtriple{s,\subclass,f}$.}]
    We apply the procedure $\mathtt{DefMinEntailmentC}(H,r(H),\dtriple{s,\subclass,f})$ and get:
    $G'\not\deriv\triple{s,\disjC, s}$ for $i=0$, that implies  $\drnk^p=\{\dtriple{b,\subclass,f}, \dtriple{b,\subclass,hf}\}$. Now, 
    $G^{str}\cup (\drnk^p)^s\deriv \triple{s,\subclass,f}$, since $\triple{b,\subclass,f}\in (\drnk^p)^s$ and $\triple{s,\subclass,b}\in G^{str}$. Therefore,
    \[
    \mathtt{DefMinEntailmentC}(H,r(H),\dtriple{s,\subclass,f}) =  \mathtt{true} \ .
    \]
  
    \item[{\bf Triple $\dtriple{s,\subclass,hf}$.}]
    We apply the procedure $\mathtt{DefMinEntailmentC}(H,r(H),\dtriple{s,\subclass,hf})$ and get:
    $G'\not\deriv\triple{s,\disjC, s}$ for $i=0$.
    $G^{str}\cup (\drnk^p)^s\deriv \triple{s,\subclass,hf}$, since $\triple{b,\subclass,hf}\in (\drnk^p)^s$ and $\triple{s,\subclass,b}\in G^{str}$, it follows that
 \[
 \mathtt{DefMinEntailmentC}(H,r(H),\dtriple{s,\subclass,hf}) =  \mathtt{true} \ .
 \]
\end{itemize}
\nd Finally, we check what happens with an extra exceptional level. Do penguins with jet-packs fly or not?
\begin{itemize}
    \item[{\bf Triple $\dtriple{pj,\subclass,f}$.}]
    We apply the procedure $\mathtt{DefMinEntailmentC}(H,r(H),\dtriple{pj,\subclass,f})$ and have:
    $G'\not\deriv\triple{pj,\disjC, pj}$ for $i=2$, that implies  $\drnk^p=\{\dtriple{pj,\subclass,f}\}$.
    $G^{str}\cup (\drnk^p)^s\deriv \triple{pj,\subclass,f}$, since $(\drnk^p)^s=\{\triple{pj,\subclass,f}\}$, and \[
    \mathtt{DefMinEntailmentC}(H,r(H),\dtriple{pj,\subclass , f}) = \mathtt{true} \ .
    \]
    
    \item[{\bf Triple $\dtriple{pj,\subclass,e}$.}]
    We apply the procedure $\mathtt{DefMinEntailmentC}(H,r(H),\dtriple{pj,\subclass,e})$ and have:
    
    $G^{str}\cup (\drnk^p)^s\not \deriv \triple{pj,\subclass,e}$, and \[
    \mathtt{DefMinEntailmentC}(H,r(H),\dtriple{pj,\subclass,e}) = \mathtt{false} \ .
    \]
\end{itemize}
    
\nd Therefore, correctly RC does not allow penguins with jet-packs to inherit the property of not-flying from typical penguins.

\qed
\end{example}

\nd The next example  shows a case in which we have some information with infinite rank, and how that indicates the presence of some conflict.

\begin{example}\label{ex_decision_2}
Let the graph $L$ contain the following information
\[
L=\{\triple{b,\subclass, ba}, \triple{mb,\subclass, b}, \triple{ba,\disjC,bw} \dtriple{mb,\subclass,bw}\} \ ,
\]
\nd where \emph{ba} is read ``breaths air", \emph{bw} is read ``breaths underwater", and \emph{mb} is the class  ``marsh bird".

When we apply the  $\mathtt{Ranking}(L)$ procedure, as
\[
L^s\deriv \triple{mb,\disjC,mb}
\]
\nd we obtain $\drnk_0=\drnk_1=\drnk_\infty=\{\dtriple{mb,\subclass,bw}\}$. Now, we can check, using the $\mathtt{StrictMinEntailment}$ procedure, that \eg~$L$ minimally entails $\triple{mb,\disjC,mb}$: since $\dtriple{mb,\subclass,bw}\in \drnk_\infty$, we have $\triple{mb,\disjC,mb}\in L'$, that obviously implies
\[
L'\deriv \triple{mb,\disjC,mb} \ .
\]
\nd Therefore, as expected, 
\[
\mathtt{StrictMinEntailment}(L,r(L), \triple{mb,\disjC,mb}) = \mathtt{true} \ .
\]
\qed
\end{example}

\nd The outcomes in Example \ref{ex_decision_2} are reasonable and desirable: despite we are dealing with defeasible information, we are facing an unsolvable conflict: we are informed that birds breath air, $\triple{b,\subclass, ba}$, without exceptions, since it is a strict $\rhodfbot$-triple, while marsh birds usually breath underwater, $\dtriple{mb,\subclass,bw}$. The triple $\triple{b,\subclass, ba}$, not being defeasible, does not allow the existence of birds breathing underwater, and from 
the information at our disposal it is reasonable to conclude that marsh birds cannot exist, that is, $\triple{mb,\disjC,mb}$. 

Triples with infinite rank appear when there is some unsolvable conflict in the graph. Another example comes when we have in the graph pieces of information that are in direct conflict with each other. For example, if we add to the graph $H$ in Example \ref{ex_decision_1} the two defeasible triples $\dtriple{mb,\subclass,f}, \dtriple{mb,\subclass,e}$ (`marsh birds typically fly' and `marsh birds typically do not fly'), that are in direct conflict with each other, we end up again concluding $\triple{mb,\disjC,mb}$.

%
%
%


\subsection{Structural Properties} \label{sect_struct}

\nd RC, like many other non-monotonic approaches, has also been analysed from a `structural properties' point of view~\cite{Makinson1994a}. For example, in the propositional case, given a knowledge base $\K$ of defeasible conditionals $\alpha \fcond  \beta$ (read  `if $\alpha$ holds, typically $\beta$ holds too'), where $\alpha$ and $\beta$ are propositions, RC satisfies a particular form of constrained monotonicity, called \emph{Rational Monotonicity} (RM) \cite{Lehmann92b}:
\begin{center}
  \begin{tabular}{llllll}
      $(RM)$ & {\large $\frac{\K\entail\alpha \fcond\beta,\hspace{0.2cm} \K\not\entail\alpha\fcond\neg\gamma}{\K\entail
      \alpha\land\gamma \fcond \beta}$} 
\end{tabular}   
\end{center}
\nd The intended meaning of an instance of the structural property above is the following: if we know that typical birds fly ($\K\entail \alpha\fcond\beta$), and we are not aware that all typical birds are not black ($\K\not\entail \alpha\fcond\neg\gamma$), then we may conclude that, typically, black birds fly ($\K\entail\alpha\land\gamma\fcond\beta$).

In our framework, a propositional defeasible conditional $\alpha\fcond\beta$ correspond to  defeasible triples of the form $\dtriple{p,\subc,q}$ and $\dtriple{p,\subp,q}$. Therefore, given a graph $G$,  the property (RM) may take a form like~\footnote{Informally, the extension of $\neg r$ is the complementary of the extension of $r$ over $\Delta_{\DC}$.}
\begin{center}
  \begin{tabular}{llllll}
      $(RM)$ & {\large $\frac{G\minentail\dtriple{p,\subc,q},\hspace{0.2cm} G\not\minentail\dtriple{p,\subc,\neg r}}{G\minentail\dtriple{p\land r,\subc,q}}$} 
\end{tabular}   
\end{center}
\nd Such a property is linked to the use of conjunction and negation of terms, which, however, are not supported (so far) in \rhodfbot. 
It does not seem possible to express $(RM)$ in defeasible \rhodfbot.


However, there are a few basic  structural properties that still can be expressed in our framework. One of the simplest one is the property of \emph{Supraclassicality}, as it is called in the propositional setting, that simply indicates that a strict piece of information implies also its own defeasible, weaker formulation, that is, 
\begin{center}
  \begin{tabular}{llllll}

      $(Supra_c)$ & {\large $\frac{G\minentail\triple{p,\subc,q}}{G\minentail\dtriple{p,\subc,q}}$}  & &
      $(Supra_p)$ & {\large $\frac{G\minentail\triple{p,\subp,q}}{G\minentail\dtriple{p,\subp,q}}$} 
\end{tabular}   
\end{center}

\begin{proposition}\label{prop_supra}
$\minentail$ satisfies $(Supra_c)$ and $(Supra_p)$.
\end{proposition}

\begin{proof}
Consider $(Supra_c)$, and let $G\minentail\triple{p,\subc,q}$. That implies that all \rhodfbot-interpretations in the minimal model $\CG$ satisfy $\triple{p,\subc,q}$. Consequently, for every \rhodfbot-interpretation $\I$ in $\cmin(p,\CG)$, $\I\rdfsat \triple{p,\subc,q}$. That is, according to Definition \ref{defsat}, $\CG\rdfsat\dtriple{p,\subc,q}$, that is to $G\rdfent\dtriple{p,\subc,q}$. The proof $(Supra_p)$ is analogous.
\end{proof}

\nd \emph{Reflexivity, Left Logical Equivalence}, and \emph{Right Weakening} are other essential properties. In the propositional case they are of the form:
\begin{center}
  \begin{tabular}{llllll}
  $(Ref)$ & { $\K\entail\alpha\fcond\alpha$} \\
  \ \\
      $(LLE)$ & {\large $\frac{\K\entail\alpha\fcond\beta, \hspace{0.2cm} \entail \alpha\equiv\gamma}{\K\entail\gamma\fcond\beta }$} & &
      $(RW)$ & {\large $\frac{\K\entail\alpha\fcond\beta,\hspace{0.2cm}  \beta\entail\gamma}{\K\entail\alpha\fcond\gamma }$} 
\end{tabular}   
\end{center}

\nd As for \emph{Supraclassicality},  these properties can also be translated in our system in two versions: one for classes and  one for predicates. For $(LLE)$, logical equivalence `$\equiv$' is translated using symmetric pairs of $\subc$- and $\subp$-triples. Specifically, the above axioms are encoded as
\begin{center}
  \begin{tabular}{llllll}
  $(Ref_c)$ & { $G\minentail\dtriple{p,\subc,p}$}\\
    \ \\
      $(LLE_c)$ & {\large 
      $\frac{ G\minentail\dtriple{p,\subc, r},\hspace{0.2cm}
      G\minentail\triple{p,\subc,q},\hspace{0.2cm} G\minentail\triple{q,\subc,p}
      }{G\minentail\dtriple{q,\subc,r}}$} & &
      $(RW_c)$ & {\large $\frac{G\minentail\dtriple{p,\subc,q},\hspace{0.2cm} G\minentail\triple{q,\subc, r}}{G\minentail\dtriple{p,\subc,r}}$} \\ 
      \ \\
%
  $(Ref_p)$ & { $G\minentail\dtriple{p,\subp,p}$}\\
    \ \\
      $(LLE_p)$ & {\large $\frac{
      G\minentail\dtriple{p,\subp, r},\hspace{0.2cm}
      G\minentail\triple{p,\subp,q},\hspace{0.2cm} G\minentail\triple{q,\subp,p}      }{G\minentail\dtriple{q,\subp,r}}$} & &
      $(RW_p)$ & {\large $\frac{G\minentail\dtriple{p,\subp,q},\hspace{0.2cm} G\minentail\triple{q,\subp, r}}{G\minentail\dtriple{p,\subp,r}}$} 
\end{tabular}   
\end{center}

\nd Note that $(Ref_c)$ and $(Ref_p)$ do not hold, as they do not hold in the classical form $G\minentail\triple{p,\subc,p}$ and $G\minentail\triple{p,\subp,p}$. This is the consequence of having considered minimal \rhodf for which reflexivity for $\subc$ and $\subp$ triples does not hold. If we use (full) \rhodf, we will recover reflexivity for the classical triples, and, by Proposition \ref{prop_supra}, we would obtain immediately also $(Ref_c)$ and $(Ref_p)$.

Concerning $(LLE)$ and $(RW)$, they are satisfied.

\begin{restatable}{proposition}{propllerw}\label{prop_lle-rw}
$\minentail$ satisfies $(LLE_c)$, $(RW_c)$, $(LLE_p)$, and $(RW_p)$.
\end{restatable}



    
    


\begin{remark}
Let us note that, despite $(RM)$ seems not to be syntactically expressible in our framework, we have proposed the same kind of construction that is behind propositional RC, that is, we model a kind of defeasible reasoning implementing the \emph{Presumption of Typicality} \cite[p.4]{Lehmann95}: if we are not informed of the contrary, we reason assuming that we are dealing with typical behaviours. 
From a semantical point of view, this ``maximisation of typicality'' has been modelled considering the \emph{minimal models} of a KB: those models in which the entities are associated to the lowest (\ie~most typical) possible rank value, modulo the satisfaction of the knowledge base. This formal solution has been used also elsewhere in order to give a semantic characterisation of RC~\cite{Booth98,Giordano15,Pearl90}.
On the other hand, from the decision procedure point of view, we have defined an algorithm that follows similar ones developed for RC in other formal frameworks; in particular, our decision procedure is built on top of the  decision procedure for the monotonic fragment, and it decides RC through a calculation of exceptionalities and rank values, as it is done \eg~for RC for propositional logic and DLs~\cite{Casini10,CasiniEtAl19,Freund1998}.

\end{remark}








\nd Nevertheless, there are some more properties satisfied by our entailment relation $\minentail$. In the following,  let us consider the closure operation $\closmin$ defined as follows: given a graph $G$, %
\[
\closmin(G):=\{\anytriple{s,p,o}\mid G\minentail\anytriple{s,p,o}\}.
\]
\nd As we may expect, $\closmin$ is not monotonic, that is, given any graph $G$ and triple $\anytriple{s,p,o}$, the following does not necessarily hold:
\[
\closmin(G)\subseteq \closmin(G\cup\{\anytriple{s,p,o}\}) \ .\footnote{Monotonicity at the level of the entailment relation is sometimes called \emph{semi-monotonicity} in the literature on non-monotonic reasoning \cite{Pearl90}.}
\]
\begin{proposition}\label{prop_monot}
$\closmin$ is not monotonic.
\end{proposition}

\begin{proof} 
Consider the graph $F$ in Example~\ref{ex_semantics}, and its subgraph 
\[
F'=\{\triple{p,\subclass,b},\triple{s,\subclass,b},\dtriple{b,\subclass,f},\triple{e,\disjC,f}\} \ ,
\]
\nd obtained eliminating the triple $\triple{p,\subclass, e}$.

It can be easily shown that \wrt~$F'$ penguins are not exceptional, \ie
\[
\mathtt{DefMinEntailmentP}(F',r(F'), \dtriple{p,\subclass,f}) = \mathtt{true} \ ,
\] 
while \wrt~$F$ penguins are  exceptional, \ie
\[
\mathtt{DefMinEntailmentP}(F,r(F), \dtriple{p,\subclass,f})=\mathtt{false} \ .
\]
\end{proof}

\nd Nonetheless, and not surprising, $\closmin(G)$ is monotonic \wrt~the strict part of the information.

\begin{proposition}\label{prop_monot_class}
Let $G$ be any graph, $\triple{s,p,o}$ any \rhodfbot-triple, and $\anytriple{s',p',o'}$ be any triple. If $G\minentail \triple{s,p,o}$, then $G\cup \{\anytriple{s',p',o'}\}\minentail \triple{s,p,o}$.
\end{proposition}

\begin{proof}
This is immediate from Theorem \ref{coroll_decision_class}. Indeed, $G\minentail \triple{s,p,o}$ implies $G'\deriv \triple{s,p,o}$, with $G'$ defined according to the procedure $\mathtt{StrictMinEntailment}$. Let $F:=G\cup \{\anytriple{s',p',o'}\}$, and let $F'$ be defined applying the procedure $\mathtt{StrictMinEntailment}$ to $F$. Clearly $G'\subseteq F'$, and, given that $\deriv$ is monotonic, we can conclude $F'\deriv \triple{s,p,o}$, that is, $F\minentail \triple{s,p,o}$.
\end{proof}

\nd This is clearly a desirable behaviour, since the strict part of our information should be treated as non-defeasible information, and consequently we should reason monotonically about it. 

$\closmin$ satisfies some other desirable properties: namely, \emph{Inclusion, Cumulativity}, and \emph{Idempotence}.

\begin{proposition}\label{prop_incl}
$\closmin$ satisfies inclusion. That is, for any graph $G$
\[
G\subseteq\closmin(G) \ .
\]
\end{proposition}

\begin{proof}
$\closmin(G)$ is determined by the model $\CG$, and, by definition,  $\CG\in\RG$, that is, $\CG$ is a model of $G$, implying $G\subseteq \closmin(G)$.
\end{proof}

\begin{proposition}\label{prop_cumul}
$\closmin$ satisfies \emph{Cumulativity}. That is, for any pair of graphs $G, G'$,
\[
\text{if }G\subseteq G'\subseteq \closmin(G)\text{, then }\closmin(G') = \closmin(G) \ .
\]
\end{proposition}

\begin{proof}
$\closmin(G)$ is determined by the model $\CG$, that is, the minimal model of $G$. Since $G'\subseteq\closmin(G)$, $\CG$ is also a model of $G'$. Therefore, $\CG$ must be the minimal model also for $G'$. Indeed, assume that is not the case, that is, there is a model $R$ of $G'$ \st~$R\preceq\CG$. Since $G\subseteq G'$, $R$ is also a model of $G$, and $\CG$ would not be the minimal model of $G$, against the hypothesis. Hence both $\closmin(G)$ and $\closmin(G')$ are determined by the model $\CG$.
\end{proof}

\nd Cumulativity has two immediate consequences that are well-known, desirable properties: a constrained form of monotonicity, called \emph{Cautious Monotonicity}, and the classical property of \emph{Cut}.

\begin{proposition}\label{prop_cm}
$\closmin$ satisfies \emph{Cautious Monotonicity}. That is, for any graph $G$ and any triple $\anytriple{s,p,o}$,
\[
\text{if }G\minentail\anytriple{s,p,o}\text{ and }G\minentail\anytriple{s',p',o'}\text{, then }G\cup\anytriple{s',p',o'} \minentail\anytriple{s,p,o} \ .
\]
\end{proposition}

\begin{proposition}\label{prop_cut}
$\closmin$ satisfies \emph{Cut}. That is, for any graph $G$ and any triple $\anytriple{s,p,o}$,
\[\text{if }G\cup\anytriple{s',p',o'}\minentail\anytriple{s,p,o}\text{ and }G\minentail\anytriple{s',p',o'}\text{, then }G \minentail\anytriple{s,p,o} \ .
\]
\end{proposition}

\nd Please note that these two properties are here analysed at the level of entailment. As discussed above, in the case of the conditional reasoning \cite{Lehmann92b} these structural properties can be expressed at two levels: at the level of conditionals (see the first formulation of (RM) above), and the meta-level of the entailment relation. Here we are looking at the properties (CM) and (Cut) at the meta-level of the entailment relation. Similarly to the (RM) case, if conjunction among terms would have been allowed, such properties at the level of the language may be expressed as 

\begin{center}
  \begin{tabular}{llllll}
      $(CM)$ & {\large $\frac{G\rdfent\dtriple{p,\subc,q},\hspace{0.2cm} G\rdfent\dtriple{p,\subc, r}}{G\rdfent\dtriple{p\land r,\subc,q}}$}  & &
      $(Cut)$ & {\large $\frac{G\rdfent\dtriple{p\land r,\subc,q},\hspace{0.2cm} G\rdfent\dtriple{p,\subc, r}}{G\rdfent\dtriple{p,\subc,q}}$} 
\end{tabular}   
\end{center}

\nd Finally, $\closmin$ satisfies also \emph{Idempotence}.

\begin{proposition}\label{prop_idemp}
$\closmin$ satisfies \emph{Idempotence}. That is, for any graph $G$, 
\[
\closmin(\closmin(G))=\closmin(G) \ .
\]
\end{proposition}

\begin{proof}
\nd \emph{Idempotence} is an immediate consequence of \emph{Inclusion} and \emph{Cumulativity}. It is sufficient to set $G' = \closmin(G)$ in the Cumulativity property, while Inclusion guarantees that $G \subseteq \closmin(G)$ holds.
\end{proof}

The satisfaction of such properties, defined at the level of entailment and also called \emph{global properties} by Lehmann and Magidor, was already proved for the original formulation of RC in the propositional case \cite[Sect. 5.5]{Lehmann92b}.

\subsection{Computational Complexity} \label{ccomp}

\nd We now address the computational complexity of the previously defined procedures and show that our entailment decision procedures run in polynomial time.

To start with, let us consider $\deriv$. Now, consider a  graph $G$ and a $\rhodfbot$-triple $\triple{s,p,o}$.
An easy way to decide whether $G \deriv \triple{s,p,o}$ holds is to compute  the closure $\clos(G)$ of $G$ and then check whether $\triple{s,p,o} \in \clos(G)$.
Now, as $G$ is ground, like for~\cite{Munoz09}, it is easily verified that, both the time to compute the  closure of $G$ as well as its size are  $O(|G|^2)$ and, thus,

\begin{proposition} \label{complderiv}
For a graph $G$ and a \rhodfbot-triple  $\triple{s,p,o}$, $G \deriv \triple{s,p,o}$ can be decided in time $O(|G|^2)$.
\end{proposition}

\begin{remark}\label{remderiv}
Let us note that \cite[Theorem 21]{Munoz09} provides also an $O(|G| \log |G|)$ time algorithm to decide the ground $\rhodf$ entailment problem.\footnote{It corresponds here to consider  rules (1)-(4).} Whether a similar algorithm  can be extended also to $\rhodfbot$ (so including also rules (5)-(7)) while maintaining the same computational complexity  is still an open problem.
\end{remark}

\nd We next consider the $\mathtt{ExceptionalC}$ and $\mathtt{ExceptionalP}$ procedures. It is immediately verified that, by Proposition~\ref{complderiv}, 
\begin{proposition} \label{complexcept}
For a defeasible graph $G=\Gclass\cup\Gdef$, both  $\mathtt{ExceptionalC}(G)$ as well as $\mathtt{ExceptionalP}(G)$  require at most $|\Gdef|$  checks and, thus, both run in time $O(|\Gdef||G|^2)$.
\end{proposition}

\nd Consider now the case of the $\mathtt{Ranking}$ procedure. It is easily verified that steps 3.-6. may be repeated at most $|\Gdef|$ times and each of which calls once the 
$\mathtt{ExceptionalC}$ and $\mathtt{ExceptionalP}$ procedures. Therefore, by Proposition~\ref{complexcept}, we have easily that 
\begin{proposition} \label{complexrank}
For a defeasible graph $G=\Gclass\cup\Gdef$,  $\mathtt{Ranking}(G)$ runs in time $O(|\Gdef|^2|G|^2)$. Moreover, the number of sets in $\rnk(G)$ is at most $O(|\Gdef|)$.
\end{proposition}

\nd Eventually, let us consider the entailment check procedures in Section~\ref{decproc}. Let us recall Remark~\ref{rnkpre} and, thus, we assume that the ranking has been computed once and for all. The time required to compute $\rnk(G)$ is, by Proposition~\ref{complexrank}, $O(|\Gdef|^2|G|^2)$.

To what concerns $\mathtt{StrictMinEntailment}$, the following result is an immediate consequence from Proposition~\ref{complderiv}.
\begin{proposition} \label{complexstric}
Consider a defeasible graph $G=\Gclass\cup\Gdef$ and a \rhodfbot-triple $\triple{s,p,o}$. Then the procedure $\mathtt{StrictMinEntailment}(G, \rnk(G), \triple{s,p,o})$ runs in time 
$O(|G|^2)$. 
\end{proposition}


\nd Now, consider $\mathtt{DefMinEntailmentC}$. By Proposition~\ref{complexrank}, it is easily verified that the steps 3.-10. may be repeated at most $O(|\Gdef|)$ times and each time we make one $\deriv$ check. By considering also the additional $\deriv$ check in step 12, by Proposition~\ref{complderiv} we have
\begin{proposition} \label{complexminentC}
Consider a defeasible graph $G=\Gclass\cup\Gdef$ and a defeasible triple $\dtriple{p,\subc,q}$. 
Then the procedure $\mathtt{DefMinEntailmentC}(G, \rnk(G), \dtriple{p,\subc,q}))$ runs in time 
$O(|\Gdef||G|^2)$. 
\end{proposition}

\nd The computational complexity of $\mathtt{DefMinEntailmentP}$ is the same as for $\mathtt{DefMinEntailmentC}$ and, thus, we conclude with
\begin{proposition} \label{complexminentP}
Consider a defeasible graph $G=\Gclass\cup\Gdef$ and a defeasible triple $\dtriple{p,\subp,q}$. 
Then the procedure $\mathtt{DefMinEntailmentP}(G, \rnk(G), \dtriple{p,\subc,q}))$ runs in time 
$O(|\Gdef||G|^2)$. 
\end{proposition}

\nd From Propositions~\ref{complderiv} - \ref{complexminentP}, it follows immediately that

\begin{corollary}\label{complexminent}
Consider a defeasible graph $G=\Gclass\cup\Gdef$ and a triple $\anytriple{p,*,q}$.
Then $G\minentail \anytriple{p,*,q}$ can be decided in polynomial time.
\end{corollary}




\section{Inheritance Networks Based Closure} \label{din}

\nd As we have seen in Remark~\ref{drowning} and Example \ref{ex_decision_1}, RC suffers from the so-called drowning problem: if a class is exceptional \wrt~a superclass it does not inherit \emph{any} of the typical property of the superclass (e.g., we can neither derive that ``drug users are usually students" nor that ``penguins usually have feathers"). Such an inferential behaviour is considered a weakness in many applications. In order to overcome such a limit,  some closure operations extending RC have been proposed: for example, the Lexicographic Closure \cite{Casini12,Lehmann95,Lukasiewicz08}, the Defeasible Inheritance-based approach~\cite{CasiniStraccia13}, the Relevant Closure~\cite{CasiniEtAl2014}, and the Multipreference Closure \cite{GiordanoGliozzi2020}.

Here we propose an adaptation of the Defeasible Inheritance-based approach~\cite{CasiniStraccia13}, originally formulated for propositional logic and DLs, to the RDFS framework. We focus  on such an approach for two reasons: it is inspired by inheritance nets~\cite{Touretzky86}, hence it is based on a formalism that is already particularly close to RDFS; and, it is promising from the computational complexity point of view, as we have already seen in~\cite{CasiniStraccia13} and in the case of the DL $\mathcal{EL}_\bot$ \cite[Section 4]{CasiniEtAl19}. 

The approach is based on refining RC by using the graph to identify the axioms taking part in each specific conflict between pieces of information. As we are going to see, such a solution will allow us both to overcome the drowning effect and to preserve computational tractability.\footnote{In~\cite[Appendix A]{CasiniStraccia13} it is also shown that  the inheritance-base closure behaves well, and better than RC \wrt~most of the ``benchmark'' examples.}
We also point out that, at the time of writing, we did not find a tractable procedure to decide defeasible subsumption under other approaches such as Lexicographic Closure, Relevant Closure, or Multipreference closure mentioned above.
However, some tractable results are known for some specific syntactic cases under Lexicographic Closure. In fact, \cite{Eiter00a} shows that, while rational closure and lexicographic entailment share the same complexity for deciding entailment of propositional conditional knowledge bases (they are both $P^{NP}$-complete), when moving to the Horn-case, the complexity of the two approaches diverge in the sense that the former is $P$-complete, while the latter still remains $P^{NP}$-complete. \cite{Eiter00a} also shows that for so-called feedback-free Horn conditional KBs\footnote{Very roughly, feedback-free Horn conditional KBs are such that the consequent of a defeasible Horn rule cannot occur in a strict Horn-rule, and we can define a particular partition, the \emph{default partition}, of the set of the defeasible Horn rules based on the vocabulary (see \cite[Sect. 6.4.2]{Eiter00a} for a detailed explanation).} lexicographic Horn entailment is polynomial. %
%
Nevertheless, we leave the option to restrict syntactically defeasible RDFS graphs under lexicographic closure (or any other before mentioned closure operation) and how these restrictions impact computationally to future work.

\subsection{Decision Procedures for Inheritance Net Based Closure} \label{decprocinh}

\nd  In the Defeasible Inheritance-based approach~\cite{CasiniStraccia13,CasiniEtAl19} the pieces of information in the KB are translated into an inheritance net, which extends the work of Touretzky~\cite{Touretzky86}. Since RDFS is already a graphical formalism, such a translation is not necessary in our framework. 
 Also, in~\cite{CasiniStraccia13,CasiniEtAl19} we have introduced the notion of \emph{duct} on a graph, which is a generalisation of the classical notion of \emph{path} that allows the formalisation of conjunctions and disjunctions, which we do not need here. In fact, we can revert to the notion of \emph{paths} only, defined inductively as follows.\footnote{We  use Greek letters $\pi,\sigma,\ldots$ to denote paths.}

\begin{definition}[path]\label{path}
Given a graph $G=\Gclass\cup\Gdef$, the \emph{paths} in $G$ are defined as follows: 
\begin{enumerate}
\item every triple $\anytriple{p,*,q}$ in $G$ corresponds to a path $\pi=\{ \anytriple{p,*,q}\} $ in $G$ from $p$ to $q$;
\item if $\pi$ is a path from $p$ to $q$ and  $\anytriple{q,*,s}$ is a triple in $G$ that does not already occur in $\pi$, then $\pi'=\pi\cup\{\anytriple{q,*,s}\}$ is a path in $G$ from $p$ to $s$.
\end{enumerate}
\end{definition}


\nd Using the notion of path we  apply the RC procedure locally: that is, if we want to decide whether $\dtriple{p,\subclass,q}$ or $\dtriple{p,\spp,q}$ are entailed by a graph, the exceptionality rankings and the RC are calculated considering only the information in the KB that has some connection to $p$ and $q$ (for more details, we refer the reader to~\cite{CasiniStraccia13}): we consider only the defeasible triples in $\Gdef$ corresponding to triples appearing in the paths  from $p$ to $q$, and we compute the RC of only such a portion of the KB.


%
For the rest of this section, we will assume that we will be working with graphs $G=\Gclass\cup\Gdef$ s.t. $\Gclass$ is closed under  $\mathtt{StrictMinEntailment}$, that is, for every $\rhodfbot$ triple $\triple{s,p,o}$, $\triple{s,p,o}\in\Gclass$ iff $\mathtt{StrictMinEntailment}(G,\mathtt{Ranking}(G),\triple{s,p,o})$ returns $\mathtt{true}$.

 Now, let $G=\Gclass\cup\Gdef$ be a graph, and consider any triple $\dtriple{p,*,q}$ with $*\in\{\subc,\subp\}$. The  procedure for deciding whether $\dtriple{p,*,q}$ is in the closure of a graph $G$ using the Defeasible Inheritance-based approach is as follows~\cite{CasiniStraccia13}.

First, we need to proceed with a first closure operation, that needs to be done once and for all: set $G_{in}\assign G$; then, for every pair of nodes $p,q$ in the graph do

\begin{description}
\item[Step 1.] Compute the set of all paths starting in $p$ and ending into $q$, and let $\Delta_{p,q}$ be the set of all the defeasible triples $\dtriple{t,*,z}$ (with $*\in\{\subclass,\subp\}$) appearing in such paths.
\item[Step 2.] Let $G_{p,q}=\Gclass\cup\Delta_{p,q}$. 
\item[Step 3.] If $G_{p,q}\minentail\dtriple{p,*,q}$, let $G_{in}\assign G_{in}\cup\{\dtriple{p,*,q}\}$.
\end{description}



\nd The procedure $\mathtt{InheritanceCompletion}$ describes the  pseudo-algorithm corresponding to such steps.

\begin{algorithm}\caption{$\mathtt{InheritanceCompletion}(G)$}\label{proc_decision_inh_C}

\algorithmicrequire\  Graph $G=\Gclass\cup\Gdef$ with $\Gclass$  closed under  $\mathtt{StrictMinEntailment}$

\algorithmicensure\ Graph $G_{in}$

\begin{algorithmic}[1]

\STATE $G_{in}\assign G$

\FORALL{$\tuple{p,q}$, s.t. $p,q\in\universe(G)$}

\STATE $\Delta_{p,q}\assign \bigcup\{\sigma\mid \sigma\text{ is a path in $G$ from $p$ to $q$}\}\setminus \Gclass$\;

\STATE $G_{p,q}\assign\Gclass\cup\Delta_{p,q}$\;

\IF{$\mathtt{DefMinEntailmentC}(G_{p,q},  \mathtt{Ranking}(G_{p,q}), \dtriple{p,\subc,q}) = \mathtt{true}$}
{\STATE $G_{in}\assign G_{in}\cup{\dtriple{p,\subc,q}}$\; 
}
\ENDIF 
\IF{$\mathtt{DefMinEntailmentP}(G_{p,q},  \mathtt{Ranking}(G_{p,q}), \dtriple{p,\subp,q}) = \mathtt{true}$}
{\STATE $G_{in}\assign G_{in}\cup{\dtriple{p,\subp,q}}$\;}
\ENDIF


\ENDFOR

\RETURN{$G_{in}$}

\end{algorithmic}
\end{algorithm}

\nd The expansion of the graph $G$ into the graph $G_{in}$ adds to the graph defeasible connections that are not involved into conflicts, but that would have been lost in RC due to the drowning problem. Example \ref{ex_inheritance_1} below will show it.

Once we have avoided the drowning effect, we define the inheritance based entailment (that we will indicate with the symbol $\vdash_{in}$) by closing $G_{in}$ under RC. That is:
\begin{definition}[Inheritance-based Entailment]\label{def_inh_ent}
For any graph $G=\Gclass\cup\Gdef$ and any triple $\anytriple{p,r,q}$,
\begin{equation}\label{eq_inh_ent}
  G \vdash_{in} \anytriple{p,r,q} \text{ iff } G_{in}\minentail \anytriple{p,r,q}  \ .
\end{equation}
\end{definition}

\nd Given Theorems \ref{theor_alg_minentC_correct} and \ref{theor_alg_minentP_correct}, Equation~\ref{eq_inh_ent} can be reformulated as follows:
%
\begin{eqnarray}\label{def_inheritance_1}
  G \vdash_{in} \triple{p,r,q} & \text{ iff }& \mathtt{StrictMinEntailment}(\mathtt{InheritanceCompletion}(G),\nonumber\\ &&\mathtt{Ranking}(\mathtt{InheritanceCompletion}(G)), \triple{p,r,q}) = \mathtt{true};\nonumber\\
  G \vdash_{in} \dtriple{p,\subc,q} & \text{ iff }& \mathtt{DefMinEntailmentC}(\mathtt{InheritanceCompletion}(G),\\ &&\mathtt{Ranking}(\mathtt{InheritanceCompletion}(G)), \dtriple{p,\subc,q}) = \mathtt{true};\nonumber\\
  G \vdash_{in} \dtriple{p,\subp,q} & \text{ iff }& \mathtt{DefMinEntailmentP}(\mathtt{InheritanceCompletion}(G),\nonumber\\ &&\mathtt{Ranking}(\mathtt{InheritanceCompletion}(G)), \dtriple{p,\subp,q}) = \mathtt{true} \ . \nonumber
\end{eqnarray}

\nd That is, given a graph $G$, if we have as query, respectively, a strict triple, a defeasible $\subc$-triple, or a defeasible $\subp$-triple, $\vdash_{in}$ is determined by calling, respectively, $\mathtt{StrictMinEntailment}$,\mbox{ } \newline $\mathtt{DefMinEntailmentC}$, and $\mathtt{DefMinEntailmentP}$.

\begin{example}[Running example cont.]\label{ex_inheritance_1}

We have already seen in Example~\ref{ex_decision_1} that $\mathtt{DefMinEntailmentC}$ and $\mathtt{DefMinEntailmentP}$ suffer from the drowning effect. This also holds for Example~\ref{pexD} (see also Remark~\ref{drowning}). In particular, in Example~\ref{pexD} we have seen that, despite young people are usually happy,  young drug users are atypical young people that are usually unhappy: \ie~$G\minentail\dtriple{yDU, \subclass, uhP}$ and  $G\not\minentail\dtriple{yDU, \subclass, hP}$.
Now, the exceptionality of the class of young drug users \wrt~the class of young people implies that, due to the drowning effect, young drug users do not inherit any of the defeasible properties associated to young people, in particular we will not be able to derive that typically young drug users are students, that is, $G\not\minentail\dtriple{yDU, \subclass, s}$.

We can easily check this. As we have seen in Example \ref{pexD}, the defeasible set associated to the query $\dtriple{yDU, \subclass, uhP}$ is 
\[
\drnkc^{yDU}= \{\dtriple{dU, \subclass, uhP}, \dtriple{dU, \subclass, yP}\} \ . 
\]
\nd As the defeasible set associated to a query triple depends on its first element, $\drnkc^{yDU}$ is associated also to the query $\dtriple{yDU, \subclass, s}$. Now, it is easy to check that 
\[
\Gclass\cup (\drnkc^{yDU})^s\not\deriv \triple{yDU, \subclass, s}
\]
\nd as $\dtriple{yP, \subclass, s}$ is not in $\drnkc^{yDU}$. As a consequence $G\not\minentail\dtriple{yDU, \subclass, s}$.

On the other hand, using $\vdash_{in}$ we avoid the drowning effect, as shown next. Our query triple is $\dtriple{yDU, \subclass, s}$. First of all we have to check whether it is in $\mathtt{InheritanceCompletion}(G)$. Consider all the paths from $yDU$ to $s$: \ie
\begin{eqnarray*}
    \sigma & = &\{\triple{yDU, \subclass, yP}, \dtriple{yP, \subclass, s}\} \\
    \sigma' & = & \{\triple{yDU, \subclass, dU}, \dtriple{dU, \subclass, yP}, \dtriple{yP, \subclass, s}\} \ .
\end{eqnarray*}
\nd That is,
\[
\Delta_{(yDU,S)}\assign (\sigma\cup\sigma')\setminus \Gclass= \{\dtriple{dU, \subclass, yP}, \dtriple{yP, \subclass, s}\}
\]
\nd and
\[
G_{(yDU,s)}=\Gclass\cup \Delta_{(yDU,s)} \ .
\]
\nd Now we have to check whether $G_{(yDU,s)}\minentail \dtriple{yDU, \subclass, s}$. We do so by invoking 
\[
\mathtt{DefMinEntailmentC}(G_{(yDU,s)},  \mathtt{Ranking}(G_{(yDU,s)}), \dtriple{yDU, \subclass, s}) \ .
\]
\nd First of all, we need to execute $\mathtt{Ranking}(G_{(yDU,s)})$. As
\begin{eqnarray*}
(G_{(yDU,s)})^s & \not\deriv & \triple{dU,\disjC, dU}  \\
(G_{(yDU,s)})^s & \not\deriv & \triple{yP,\disjC, yP} \ ,
\end{eqnarray*}
\nd we obtain $r(G_{(yDUr,s)})=\{\drnkc^0, \drnkc^\infty\}$ with 
\[
\drnkc^0= \{\dtriple{yP, \subclass, s}, \dtriple{dU, \subclass, yP}\} \ .
\]
\nd and $\drnkc^\infty=\emptyset$. That is, $G_{(yDU,s)}$ is a subgraph of $G$ without any exceptionality. It is now easy to check that the defeasible set associated to the query triple $\dtriple{yDU, \subclass, s}$ is
\[
\drnkc^{yDU}=\drnkc^0
\]
\nd and we obtain 
\[
\Gclass\cup(\drnkc^0)^s\deriv \triple{yDU, \subclass, s} \ ,
\]
\nd as we have $\triple{yDU, \subclass, yP}$ and $\triple{yP, \subclass, s}$ in $\Gclass\cup(\drnkc^0)^s$. That is,
\[
G_{(yDU,s)}\minentail \dtriple{yDU, \subclass, s} \ .
\]
\nd Therefore, $\dtriple{yDU, \subclass, s}\in\mathtt{InheritanceCompletion}(G)$, that is, $\dtriple{yDU, \subclass, s}\in G_{in}$. Given that $G\vdash_{in} \dtriple{yDU, \subclass, s}$ iff $G_{in} \minentail \dtriple{yDU, \subclass, s}$ or, equivalently, 
\[
\dtriple{yDU, \subclass, s}\in\mathtt{DefMinEntailmentC}(\mathtt{InhertanceCompletion}(G)) \ ,
\]
\nd $\dtriple{yDU, \subclass, s}\in G_{in}$, and $\minentail$ satisfies Inclusion (Proposition \ref{prop_incl}), we can conclude that
\[
G\vdash_{in} \dtriple{yDU, \subclass, s} \ ,
\]
\nd as desired.

On the other hand, it is easy to check that in case of conflicting information the inheritance-based approach and rational closure behave alike.

To see this, we can check whether the triple $\dtriple{yDU,\subclass,uhP}$ follows from $G$, as in Example \ref{pexD}. The paths from $yDU$ to $uhP$ are:
\begin{eqnarray*}
\sigma & = &\{\triple{yDU, \subclass, yP}, \dtriple{yP, \subclass, hP}, \triple{hP, \disjC, uhP}\} \\
\sigma' & = & \{\triple{yDU, \subclass, dU}, \dtriple{dU, \subclass, yP}, \dtriple{yP, \subclass, hP}, \triple{hP, \disjC, uhP}\} \\
\sigma'' & = & \{\triple{yDU, \subclass, dU}, \dtriple{dU, \subclass, uhP}\} \ .
\end{eqnarray*}
\nd Therefore,
\[
\Delta_{(yDU,uhP)}\assign (\sigma\cup\sigma'\cup\sigma'')\setminus \Gclass= \{\dtriple{dU, \subclass, yP}, \dtriple{yP, \subclass, hP}, \dtriple{dU, \subclass, uhP}\}
\]
\nd and
\[
G_{(yDU,uhP)}=\Gclass\cup \Delta_{(yDU,uhP)} \ .
\]
%
\nd Now we check whether $G_{(yDU,uhP)}\minentail\dtriple{yDU,\subclass,uhP}$ holds. At first, we rank $\Delta_{(yDU,uhP)}$ via $\mathtt{Ranking}(G_{(yDU,uhP)})$. We have:
\[
\drnk_0\assign\Delta_{(yDU,uhP)} \ .
\]
\nd We need to check whether $dU,yP$ are `empty classes' in $(G_{(yDU,uhP)})^s$. Indeed, as for $\Gcount$ in Example \ref{pexD}, we obtain $(G_{(yDU,uhP)})^s\deriv\triple{dU,\disjC, dU}$, while $yP$ does not turn out to be empty in $(G_{(yDU,uhP)})^s$. Consequently,  $\mathtt{ExceptionalC}(\Gclass\cup \drnk_0) \assign \{\dtriple{dU, \subclass, uhP},  \dtriple{dU, \subclass, yP}\}$ and also $\mathtt{ExceptionalP}(\Gclass\cup \drnk_0) \assign \emptyset$. Therefore,
\[
\drnk_1\assign\{\dtriple{dU, \subclass, uhP},  \dtriple{dU, \subclass, yP}\} \ .
\]
\nd Next, in $\Gclass\cup \drnk_1$, $dU$ is not exceptional anymore, and we obtain  
\[
\drnk_2\assign\emptyset \ .
\]
\nd That is, 
\[
\drnk_\infty\assign\emptyset
\]
\nd and, thus,  
\[
\rnk(G_{(yDU,uhP)})=\{\drnk_0, \drnk_1,  \drnk_\infty\} \ .
\]
\nd Now we can check whether $G_{(yDU,uhP)}\minentail\dtriple{yDU, \subclass, uhP}$.
To do so, we call 
\[
\mathtt{DefMinEntailmentC}(G_{(yDU,uhP)}, \rnk(G_{(yDU,uhP)}, \dtriple{yDU, \subclass, uhP}) \ .
\]
\nd For $i=0$, we have $G'=(G_{(yDU,uhP)})^s$, and $G'\deriv \triple{yDU, \disjC, yDU}$, since $(G_{(yDU,uhP)})^s\deriv \triple{dU, \disjC, dU}$, that, together with $\triple{yDU, \subclass, dU}$, implies  $\triple{yDU, \disjC, yDU}$ by the rule $(EmptySP')$.


For $i=1$, we have $G'=\Gclass\cup(\drnk_1)^s$, and $G'\not\deriv \triple{yDU, \disjC, yDU}$. Hence we associate the defeasible set $\drnk_1$ to the query $\dtriple{yDU, \subclass, uhP}$.
Following the procedure $\mathtt{DefMinEntailmentC}$ we obtain:
\[
\drnkc^{yDU}\assign \drnk_1\setminus\drnk_\infty= \{\dtriple{dU, \subclass, uhP}, \dtriple{dU, \subclass, yP}\} \ .
\]
\nd Eventually, it is easy to check that:
\[
\Gclass\cup (\drnkc^{yDU})^s\deriv \triple{yDU, \subclass, uhP}
\]
as $\triple{dU, \subclass, uhP}\cup \triple{yDU, \subclass, dU}\subseteq \Gclass\cup (\drnkc^{yDU})^s$.

Therefore, we can conclude $G_{yDU,uhP}\minentail\dtriple{yDU, \subclass, uhP}$. That implies that $\dtriple{yDU, \subclass, uhP}\in G_{in}$, and consequently $G\vdash_{in}\dtriple{yDU, \subclass, uhP}$.

We can verify that actually 
\[
G_{in}\assign G\cup\{\dtriple{yDU, \subclass, uhP},\dtriple{yDU, \subclass, s}\} \ .
\]
\nd For example, $\dtriple{yDU, \subclass, hP}\notin G_{in}$: in fact, we have $\Delta_{(yDU,hP)}=\Delta_{(yDU,uhP)}$, that implies that we have the same ranking and the same set $\drnkc^{yDU}$. Moreover, $\Gclass\cup (\drnkc^{yDU})^s\not\deriv \triple{yDU, \subclass, hP}$, since $\triple{yP, \subclass, hP}\notin \Gclass\cup (\drnkc^{yDU})^s$, which implies that $G_{(yDU,uhP)}\not\minentail\dtriple{yDU, \subclass, hP}$. 

Given $G_{in}$, and in particular checked that $\dtriple{yDU, \subclass, hP}\notin G_{in}$, it is possible to verify that $G_{in}\not\minentail \dtriple{yDU, \subclass, hP}$, that is, $G\not\vdash_{in}\dtriple{yDU, \subclass, hP}$.
\qed
\end{example}

\subsection{Structural Properties} \label{sect_struct_inh}

\nd Here we check that Inheritance-based entailment $\vdash_{in}$ satisfies some of the properties satisfied also by RC. Most of the proofs are straightforward, since $\vdash_{in}$ corresponds to an RC over the graph $G_{in}$ (see Definition \ref{def_inh_ent}). Let $\closinh$ be the closure operation corresponding to $\vdash_{in}$:
\[
\closinh(G):=\{\anytriple{s,p,o}\mid G\vdash_{in}\anytriple{s,p,o}\}.
\]
\nd Some properties are the same as in Section \ref{sect_struct}, simply reformulated for $\vdash_{in}$. 
\begin{center}
  \begin{tabular}{llllll}
      $(Supra_c)$ & {\large $\frac{G\vdash_{in}\triple{p,\subc,q}}{G\vdash_{in}\dtriple{p,\subc,q}}$} & &
      $(Supra_p)$ & {\large $\frac{G\vdash_{in}\triple{p,\subp,q}}{G\vdash_{in}\dtriple{p,\subp,q}}$} \  
\end{tabular}   
\end{center}

\nd The proof of the next proposition proceeds as in Proposition \ref{prop_supra}, just considering the graph $G_{in}$ instead of the graph $G$.

\begin{proposition}\label{prop_supra_inh}
$\rdfent$ satisfies $(Supra_c)$ and $(Supra_p)$.
\end{proposition}

\nd Then we have \emph{Reflexivity, Left Logical Equivalence}, and \emph{Right Weakening}.

\begin{center}
  \begin{tabular}{llllll}
  $(Ref_c)$ & { $G\vdash_{in}\dtriple{p,\subc,p}$} \\ \\
      $(LLE_c)$ & {\large 
      $\frac{ G\vdash_{in}\dtriple{p,\subc, r},\hspace{0.2cm}
      G\vdash_{in}\triple{p,\subc,q},\hspace{0.2cm} G\vdash_{in}\triple{q,\subc,p}
      }{G\vdash_{in}\dtriple{q,\subc,r}}$}  & &
      $(RW_c)$ & {\large $\frac{G\vdash_{in}\dtriple{p,\subc,q},\hspace{0.2cm} G\vdash_{in}\triple{q,\subc, r}}{G\vdash_{in}\dtriple{p,\subc,r}}$} \\ \\
  $(Ref_p)$ & { $G\vdash_{in}\dtriple{p,\subp,p}$} \\ \\
      $(LLE_p)$ & {\large $\frac{
      G\vdash_{in}\dtriple{p,\subp, r}, \hspace{0.2cm}
      G\vdash_{in}\triple{p,\subp,q}, \hspace{0.2cm} G\vdash_{in}\triple{q,\subp,p}      }{G\vdash_{in}\dtriple{q,\subp,r}}$} & &
      $(RW_p)$ & {\large $\frac{G\vdash_{in}\dtriple{p,\subp,q},\hspace{0.2cm} G\vdash_{in}\triple{q,\subp, r}}{G\vdash_{in}\dtriple{p,\subp,r}}$} \ 
\end{tabular}   
\end{center}

\nd As for $\minentail$, $(Ref_c)$ and $(Ref_p)$ do not hold, since we have used  minimal \rhodf~for which reflexivity of $\subc$ and $\subp$ triples does not hold. 

The proof of Proposition~\ref{prop_lle-rw_inh} proceeds as in Proposition~\ref{prop_lle-rw}, just considering the graph $G_{in}$ instead of the graph $G$.

\begin{proposition}\label{prop_lle-rw_inh}
$\vdash_{in}$ satisfies $(LLE_c)$, $(RW_c)$, $(LLE_p)$, and $(RW_p)$.
\end{proposition}
%



\begin{proposition}\label{prop_monot_inh}
$\closinh$ is not monotonic.
\end{proposition}

\nd Once more, it can be proved as for Proposition~\ref{prop_monot}.

\begin{proposition}\label{prop_incl_inh}
$\closinh$ satisfies inclusion. That is, for any graph $G$
\[G\subseteq \closinh(G).\]
\end{proposition}

\begin{proof}
$G\subseteq G_{in}$. $\closmin(G_{in})$ is determined by a model of $G_{in}$, implying $G\subseteq G_{in}\subseteq \closmin(G_{in})$, that is, $G\subseteq \closinh(G)$.
\end{proof}
\nd However, the operator $\closinh$ does not always satisfy \emph{Idempotence}.

\begin{restatable}{proposition}{propidempinh}\label{prop_idemp_inh}
$\closinh$ does not satisfy \emph{Idempotence}. That is, there is a graph $G$ \st 
\[
\closinh(\closinh(G))\neq\closinh(G) \ .
\]
\end{restatable}

\nd  An immediate consequence of Proposition \ref{prop_idemp_inh} is also the failure  of  \emph{Cumulativity}.

\begin{proposition}\label{prop_cumul_inh}
$\closinh$ does not satisfy \emph{Cumulativity}. That is, there are at least two graphs $G$ and $G'$ \st 
\[
G\subseteq G'\subseteq \closinh(G)\text{ and }  \closinh(G')\neq\closinh(G) \ .
\]
\end{proposition}

\nd Note that the same graph used in the proof of Proposition \ref{prop_idemp_inh} proves  also Proposition \ref{prop_cumul_inh}, as it is sufficient to set $G'=\closinh(G)$.

\begin{remark}\label{rem_iteration}
As stated in Proposition \ref{prop_idemp_inh}, $\closinh$ does not satisfy \emph{Idempotence}. However, it is still possible to refine $\closinh$ so that it does satisfy Idempotence. In fact,  it is sufficient to iterate the application of $\mathtt{InheritanceCompletion}$ followed by $\closmin$, until we do reach a fixed point. Given that we work with finite graphs, such a fixed point must be reached after a finite number of iterations.
In particular, given a graph $G$, the procedure $\closinh$ can be iterated at most twice for each pair of nodes $p,q$ in $G$, once to obtain a new triple $\dtriple{p,\subc,q}$, and once to obtain a new triple $\dtriple{p,\subp,q}$. Hence the  new procedure would consist in  at most $2(|G|^2)$ reiterations of the procedure $\closinh$ and, thus, we still remain polynomial to decide defeasible entailment. We have preferred to introduce only one iteration of $\closinh$, since the graph for which Idempotence does not hold appears to be quite artificial and has a complex configuration (see  the example used to prove Proposition \ref{prop_idemp_inh}). In particular, the inferential advantages given by the reiteration of the $\closinh$ operation do not appear to be worth of the extra computational costs. 

However, at the time of writing, we are still missing a proof showing whether by iterating $\closinh$ we  obtain or not \emph{Cumulativity}.
\end{remark}

\subsection{Computational Complexity} \label{ccomp_inh}

\nd In this section we prove that deciding $\vdash_{in}$ is still computationally tractable, \ie~it runs in polynomial time. We start with  $\mathtt{InheritanceCompletion}$.

In line 3 of the $\mathtt{InheritanceCompletion}$ procedure we need to compute all paths between two nodes.

\begin{restatable}{lemma}{lemmacomplexclassclosureDelta}\label{lemma_complexclassclosureDelta}
Consider a defeasible graph $G=\Gclass\cup\Gdef$ and two nodes, $p$ and $q$, in $G$. 
$\Delta_{p,q}$ can be determined in $O(|G|^2)$ time.  
\end{restatable}

\nd Given Lemma \ref{lemma_complexclassclosureDelta}, we can show that:

\begin{restatable}{proposition}{propcomplexinhcompl}\label{prop_complexinhcompl}
Consider a defeasible graph $G=\Gclass\cup\Gdef$. Assume that   $\Gclass$ is closed under $\mathtt{StrictMinEntailment}$. 
Then the procedure $\mathtt{InheritanceCompletion}(G)$ runs in time $O(|\Gdef||G|^4)$. 
\end{restatable}




\nd We conclude with:

\begin{restatable}{corollary}{propcomplexinhcomplinher}\label{prop_complexinhcomplinher}
Consider a defeasible graph $G=\Gclass\cup\Gdef$. Then $G\vdash_{in}\anytriple{p,o,q}$ can be decided in polynomial time.
\end{restatable}

\section{Dealing with Non-Ground Graphs} \label{blanknodes}

\nd So far, we assumed graphs to be ground. However, as pointed out in \eg~\cite{Hogan14}, many $\rhodf$ graphs contain so-called \emph{blank nodes} in triples, \ie~in logical terms existentially quantified variables. We next show how our approach may be extended to non-grounded $\rhodfbot$ graphs as well by preserving the computational complexity of the ground case.

\paragraph{Defeasible $\rhodfbot$}
To start with, again we recap here only the salient notions we will rely on and refer the reader \eg~directly to~\cite{Gutierrez11,Munoz09} for more details. So, let  $\AB$ be a new alphabet, pairwise disjoint to $\AU$ and $\AL$, denoting \emph{blank nodes}, and let us denote with $\AUBL$ the union of these alphabets. 
A (possibly non-ground) $\rhodfbot$ \emph{triple} is now of the form $\tau=\triple{s,p,o} \in \AUBL \times \AU \times \AUBL$, where again $s,o \notin \rhodf$. We will use symbols $x,y,z$  (with optional sub- or super-scripts) to denote blank nodes in triples. 
However, \emph{we still assume that defeasible triples remain ground and, thus, \eg~triples of the form $\dtriple{c,\subclass, d}$ (indicating that ``typically, an instance of  $c$ is also an instance of  $b$'') remain ground}.~\footnote{In fact, so far we do not envisage meaningful cases in which such triples may be not ground. Nevertheless, we will leave  the case in which this type of triples may contain blank nodes as well for future work.} 

We define a \emph{map}  as a function $\mu : \AUBL \to \AUBL$ preserving URIs and literals, \ie $\mu(t) = t$, for all $t \in \AUL$. Given a graph $G$, we define $\mu(G)= \{ \triple{\mu(s), \mu(p), \mu(o)} \mid \triple{s, p, o}\in G \}$. We speak of a map $\mu$ from $G_{1}$ to $G_{2}$, and write $\mu : G_{1} \to G_{2}$, if $\mu$ is such that $\mu(G_{1}) \subseteq G_{2}$.

From a semantical point of view, we extend the notion of interpretation to capture the idea of existentiality of blank nodes by imposing  the following additional condition to an interpretation $\I =\tuple{\Delta_{\DR}, \Delta_{\DP}, \Delta_{\DC},   \Delta_{\DL}, \intP{\cdot}, \intC{\cdot}, \int{\cdot}}$: 

\begin{enumerate}
\setcounter{enumi}{7}
 \item on \AB, $\int{\cdot}$  is a function $\int{\cdot}\colon \AB \to \Delta_{R}$.
\end{enumerate}

\nd The notion of satisfiability (resp.~entailment) is as Definition~\ref{satisfaction} (resp. Definition~\ref{entailment}). 

From a deductive system point of view, as reported in~\cite{Munoz09}, we have to consider some additional rules: namely,

 \begin{enumerate}
 \setcounter{enumi}{7}
  \item Implicit Typing: \\[0.5em]
    \begin{tabular}{llll}
      $(a)$ & $\frac{\triple{A, \dom, B},  \triple{D,  \spp, A}, \triple{X, D, Y}}{\triple{X, \type, B}}$  &
      $(b)$ & $\frac{\triple{A, \range, B},  \triple{D,  \spp, A}, \triple{X, D, Y}}{\triple{Y, \type, B}}$
    \end{tabular}  
    \item Simple': \\[0.5em]
    \begin{tabular}{llll}
        $\frac{G}{G'}$ for a map $\mu:G' \to G$ 
    \end{tabular}
\end{enumerate}    

\nd  with the convention that in all rules (2) - (8) meta-variables may represent now elements in $\AUBL$. Please note that,  rule (9) captures the semantics of blank nodes.

Finally, the notion of \emph{derivation} (Definition~\ref{def:derivation}) is extended in the obvious way:

\begin{definition}[Derivation $\deriv$]\label{def:derivatioGn}
Let $G$ and $H$ be $\rhodfbot$-graphs. $ G\deriv H$ iff there exists a sequence of graphs $P_1,P_2,\ldots, P_k$ with $P_1=G$ and $P_k=H$ and for each $j$ ($2 \leq j \leq k$) one of the following cases hold:

\begin{itemize}
\item there is a map $\mu: P_j\rightarrow P_{j-1}$ (rule (9));
\item $P_j \subseteq P_{j-1}$ (rule (1));
\item there is an instantiation $R/R'$ of one of the rules (2)-(
8), such that
$R \subseteq P_{j-1}$ and $P_j = P_{j-1} \cup R'$.
\end{itemize}

\nd \nd Such a sequence of graphs is called a proof of $G\deriv H$. Whenever
$G\deriv H$, we say that the graph $H$ is derived from the graph $G$. Each pair $(P_{j-1}, P_j)$, $1\leq j \leq k$ is called a step of the proof which
is labeled by the respective instantiation $R/R'$ of the rule applied at the step.
\end{definition}

\nd With $\clos_B(G)$ we denote the closure of a $\rhodfbot$ graph $G$ under the application of rules $(2)-(8)$ (see also~\cite{Munoz09,Gutierrez11}).

\begin{example} \label{exclosureB}
Consider the $\rhodfbot$ graph $G$ containing the triples
\begin{eqnarray*}
G & = & \{\triple{a,\subclass, x}, \triple{x,\subclass, b}, \triple{y,\subclass, c},  \\
&& \triple{c,\disjC, d}, \triple{p,\Dom, y}, \triple{q,\Dom, d} \} \ .
\end{eqnarray*}

\nd Then, it is not difficult to verify that $\clos_B(G)$ contains the following triples:

\begin{eqnarray*}
\clos_B(G) & \supseteq & \{\triple{a,\subclass, b},  \triple{y,\disjC, d}, \triple{p,\disjP, q} \} \ .
\end{eqnarray*}


\qed
\end{example}

\nd Now, likewise for Theorem~\ref{Th:soundcomplete}, by using also~\cite[Theorem 8]{Munoz09} and~\cite[Theorem 10]{Munoz09}, we may prove similarly that

\begin{theorem}\label{Th:soundcompleteB}
Let $G$ and $H$ be possibly non-ground $\rhodfbot$-graphs. 
\begin{enumerate}
\item $G\deriv H$ iff $G\rdfent H$.
\item If $G\deriv H$ then  there is a proof of $H$ from $G$  such that rule (9) is used at most once and at the last step of the proof.
\end{enumerate}
\end{theorem}

\begin{example}[Example~\ref{exclosureB} cont.]\label{exclosureBB}
By referring to Example~\ref{exclosureB}, it is easily verified that indeed 
\begin{eqnarray*}
G & \rdfent & \{\triple{a,\subclass, b},  \triple{y,\disjC, d}, \triple{p,\disjP, q} \} \ .
\end{eqnarray*}

\end{example}



\nd From a computational complexity point of view, from~\cite{Horst05} and Theorem~\ref{Th:soundcompleteB}, point 2, it follows immediately that, as for $\rhodf$, also for $\rhodfbot$ we have that

\begin{theorem}\label{terHorst}
Let $G$ and $H$ be possibly non-ground $\rhodfbot$-graphs. Then 
\begin{enumerate}
    \item Deciding $G\rdfent H$ is an NP-complete problem.
    \item If $H$ is ground, then  $G\rdfent H$ iff
    $H \subseteq \clos_B(G)$.
    \item  The size of $|\clos_B(G)|$ is $O(|G|^2)$.
\end{enumerate}
\end{theorem}

\nd In particular, note that even if $G$ may contain blank nodes, if $H$ is ground, then deciding $G\rdfent H$ can be done in time $O(|G|^2)$.





Please note also the following: by using Theorem~\ref{Th:soundcompleteB}, point 2, and Theorem~\ref{terHorst}, we may also decide  the case $G\rdfent \triple{s,p,o}$, where $s,o$ may be blank nodes, in  time $O(|G|^2)$ by computing $\clos_B(G)$ incrementally and then 
\begin{description}
\item[Case $\triple{x,p,o}, x\in \AB, p,o\in \AUL$:] check whether there is $\triple{s,p,o} \in \clos_B(G)$ with $s\in \AUBL$;
\item[Case $\triple{s,p,y}, y\in \AB, s,p\in \AUL$:] check whether there is $\triple{s,p,o} \in \clos_B(G)$ with $o\in \AUBL$;
\item[Case $\triple{x,p,y}, x,y\in \AB, p\in \AUL$:] check whether there is $\triple{s,p,o} \in \clos_B(G)$ with $s,o\in \AUBL$.
\end{description}

\nd Therefore,

\begin{corollary}\label{blankcompl}
Let $G$ and $H$ be possibly non-ground $\rhodfbot$-graphs. If $H$ is ground or $|H|=1$ then
$G\rdfent H$ can be decided in time $O(|G|^2)$.
\end{corollary}


\nd Next, we turn our attention to defeasible graphs $G$ in which the strict $\rhodfbot$ part may contain non-ground triples. We ask about the computational complexity of deciding defeasible entailment. Now, note the following:
\begin{enumerate}
    \item Consider the $\mathtt{ExceptionalC}$ and $\mathtt{ExceptionalP}$ procedures. In both cases, in Step 3., the query triple is ground, and thus,  by Corollary~\ref{blankcompl}, both procedures run in polynomial time, and Proposition~\ref{complexcept} applies.
    

    \item From the previous point, also the ranking procedure  $\mathtt{Ranking}$ runs in polynomial time and  Proposition~\ref{complexrank} applies.
    
    \item From what was said above and Corollary~\ref{blankcompl}, also the procedure $\mathtt{StrictMinEntailment}$ runs in polynomial time and Proposition~\ref{complexstric} applies.
    
    \item Finally, consider procedures $\mathtt{DefMinEntailmentC}$ and $\mathtt{DefMinEntailmentP}$. Again, as the involved query triples are ground, both procedures run in polynomial time, and Propositions~\ref{complexminentC} and~\ref{complexminentP} apply.
\end{enumerate}

\nd Therefore, the analogue of Corollary~\ref{complexminent} holds. That is,

\begin{theorem} \label{complexminentblankC}
Consider a defeasible graph $G=\Gclass\cup\Gdef$, where the  $\rhodfbot$ part of $\Gclass$ may contain blank nodes. Consider a triple $\anytriple{p,*,q}$. Then $G\minentail \anytriple{p,*,q}$ can be decided in polynomial time.
\end{theorem}




\paragraph{Inheritance Networks Based Closure}
We eventually show how to extend  our decision procedure for inheritance net closure described in Section~\ref{din} also to the non-ground case.

At first, as for Section~\ref{din}, we will assume  a possibly non-ground graph $G=\Gclass\cup\Gdef$ s.t. $\Gclass$ is closed under  $\mathtt{StrictMinEntailment}$. How to do that has been described above.
As next, we extend the notion of \emph{path} given in Definition~\ref{path} naturally to the case in which also non-ground triples may occur in a path. So, for instance, by referring to the graph $G$ in Example~\ref{exclosureB}, 
\[
\pi = \{\triple{a,\subclass, x}, \triple{x,\subclass, b} \}
\]

\nd is a path in $G$ from $a$ to $b$.




Finally,  given a possibly non-ground graph $G$ and  a (ground) defeasible triple $\dtriple{p,*,q}$ with $*\in\{\subc,\subp\}$, the  procedure for deciding whether $\dtriple{p,*,q}$ is in the closure of a graph $G$ using the Defeasible Inheritance-based approach is eactly as the one described in Section~\ref{din}. That is, we apply to $G$ the procedure $\mathtt{InheritanceCompletion}(G)$, obtaining $G_{in}$ and then define (see Definition~\ref{def_inh_ent}), for any triple $\anytriple{p,r,q}$, that
\[
 G \vdash_{in} \anytriple{p,r,q} \text{ iff } G_{in}\minentail \anytriple{p,r,q}  \ .
 \]

\nd From a computational complexity point of view, it is then easily verified that Lemma~\ref{lemma_complexclassclosureDelta} holds for the non-ground case as well. Then, using Corollary~\ref{blankcompl}, we obtain the analogue of Corollary~\ref{prop_complexinhcomplinher}:

\begin{corollary} \label{prop_complexinhcomplinherB}
Consider a possibly non-ground defeasible graph $G=\Gclass\cup\Gdef$. Then $G\vdash_{in}\anytriple{p,r,q}$ can be decided in polynomial time.
\end{corollary}

\nd In summary, even if blank nodes are allowed in $\rhodfbot$, the computational complexity does not change \wrt~the ground case, which concludes this section.

\section{Related Work} \label{relw}

\nd There have been some works in the past about extending RDFS with non-monotonic capabilities, which we briefly summarise below. 

The series of works \cite{Analyti08,Analyti15} essentially deals with \emph{Extended} RDFS (ERDF), where an ERDF ontology consists of two parts: an ERDF graph and an ERDF program containing derivation rules. ERDF augments the expressivity of RDFS in the following way: an ERDF graph allows \emph{negated} triples of the form $\neg\triple{s,p,o}$ indicating informally that 
$\neg p(s,o)$\footnote{More precisely, for a predicate $p$ there is a positive extension of $p$ and a negative extension of $p$, which need not necessarily be disjoint. So, for instance, $\triple{s,p,o}$ enforces  $(s,o)$ to be in the positive extension of $p$, while $\neg \triple{s,p,o}$ enforces  $(s,o)$ to be in the negative extension of $p$.} does hold, while in the body of derivation rules all the classical  connectives 
$\neg, \supset, \land, \lor, \forall, \exists$, plus the weak negation (negation-as-failure) $\sim$ are allowed. Thanks to the latter, we can model non-monotonic reasoning via negation-as-failure semantics. For instance, one may express rules such as 
$
\neg \type(x,EUMember) \leftarrow \sim \type(x,EUMember)$ (a non-EU Member state is one that can not be proven to be a EU member state), or 
$\neg \type(x,EUMember) \leftarrow \type(x,AmericanCountry)
$.

ERDF allows to combine open-world  and closed-world reasoning. In order to define entailment the authors have proposed a \emph{stable model semantics} \cite{Analyti08} and \emph{\#n-stable model semantics} \cite{Analyti15}. In general, under stable model semantics, query answering is undecidable. Decidability can be obtained either by constraining the language or using the \emph{\#n-stable model semantics} \cite{Analyti15}. 
Moreover, from a computational complexity point view, decision problems in ERDF are non-polynomial (see Table 1 in \cite{Analyti15} for details). For instance, deciding model existence,  and, thus, model existence is not guaranteed, ranges from NP to PSPACE, while query answering goes from co-NP to PSPACE, depending on the setting. In our case, model existence is guaranteed, the computational complexity is lower and no rule layer is introduced, remaining, thus, a triple language. Furthermore, while  the authors did enrich RDFS language with two negation operators, we have introduced the disjointness properties $\disjC$ and $\disjP$ only, which are weaker than negation from a logical point of view. 



On a similar line,  \ie~using rule languages on top of RDFS, and borrowing non-monotonic semantics developed within rules languages, are also all other approaches we are aware of, such as \cite{Antoniou02,Billig08,Kontopoulos08} and practical large scale solutions such as \cite{Ianni09,Pham17,Tachmazidis12}. 
Let us also note that there are more general solutions in providing a rule layer on top of  ontology languages such as RDFS and OWL, which can be found in \eg~\cite{Eiter08,Eiter11}.

Strongly related to the present proposal is the introduction of non-monotonic reasoning in the framework of DL (a First-Order Logic that restricts to unary and binary predicates and specific syntactic constructs), that is aimed at modelling defeasible reasoning in the other main ontology formalism, the OWL family of languages. In particular, also in this area some of the most popular proposals are based on semantic solutions associated to rational closure, as mentioned in the introduction.

In~\cite{Casini10}  a form of rational closure for DLs is defined, that has later received a semantic characterisation in terms of ranked DL intepretations \cite{BritzEtAl2020,CasiniEtAl19}. \cite{Casini10} introduces in the language a defeasible  subsumption relation $\dsubs$ between concepts, where $C\dsubs D$ is read as `Typically, an instance of the concept $C$ is also an instance of the concept $D$'.

In \cite{Giordano15} a similar approach is also considered, introducing in the language inclusion axioms of the form ${\bf T}C\subs D$, whose meaning is the same as $C\dsubs D$, and defining on the semantical side modular orderings on the domain, and it has been proved \cite[Proposition 30]{CasiniEtAl19} that the main semantic construction they propose is in fact equivalent to the one later proposed in \cite{BritzEtAl2020,CasiniEtAl19}.  Recently, the work \cite{Bonatti2019} has also investigated how to extend the rational closure construction to DLs that do not satisfy the \emph{disjoint union model property}.

As mentioned above, the entailment relation defined by rational closure suffers from the ``drowning problem'' (see Remark \ref{drowning}), that here we have addressed in Section \ref{din} defining the inheritance-based approach. Also in DLs the same problem has been addressed in various ways, including the original formulation of the inheritance-based closure \cite{CasiniStraccia13}, and other extensions of rational closure \cite{Casini12,CasiniEtAl2014,GiordanoGliozzi2020}.

Finally, another limit of rational closure in the framework of DLs is that it does not allow to make defeasible inferences across role connections: for example, if our ontology contains only $C\dsubs D$ and $\top\subs \exists R.C$, we cannot conclude $\top\dsubs \exists R.D$. Such a limit has been addressed in \eg~\cite{Pensel18}, the aim of which is to extend  rational closure  with defeasible inheritance across role expressions in the description logic~$\mathcal{EL}_{\bot}$. An entailment relation that addresses the same issue, but without using a semantics based on rational closure, is the one proposed in~\cite{BonattiEtAl2015}. In the RDFS framework this type of problem, \ie~ defeasible inheritance across existential role expressions is not present, due to the expressivity constraints of $\rhodf$.

In summary, adopting a rule layer or, in general, a more expressive logic, however, may also come with an increase in computational cost.


\section{Brief Conclusions} \label{concl}

\nd {\bf Contribution.} We have shown how one may integrate RC within RDFS and, thus, obtain a non-monotonic variant of the latter.
To  do so, we started from \rhodf, which is the logic behind RDFS, and then extended it to \rhodfbot, allowing to state that two entities are incompatible. Eventually, we have worked out defeasible \rhodfbot~via a typical RC construction. Furthermore, we have addressed the ``drowning problem'', via a `local', path-based application of RC.

The main and unique features of our approach are summarised as follows: 
\begin{itemize}

\item  defeasible \rhodfbot~remains syntactically a triple language and is a simple  vocabulary extension of \rhodf~by introducing some new predicate symbols, namely  $\disjC$ and $\disjP$, with specific semantics  allowing to state that two terms are incompatible;

\item any RDFS reasoner/store may handle the new types of triples as ordinary ones if it does not want to take account of the extra semantics of the new predicate symbols;

\item  the defeasible entailment  decision procedure is built on top of the \rhodfbot~entailment decision procedure, which in turn is an extension of the one for \rhodf~via some additional inference rules, favouring a potential implementation;

\item each defeasible graph has an unique minimal model;

\item the ``drowning problem" is addressed via a simple, `local', path-based application of RC;

\item defeasible entailment can be decided in polynomial time.

\end{itemize}

\nd We have also analysed our proposal from a `structural properties' point of view with respect to those properties that can be expressed in our language.\footnote{We recall that some can not be expressed in \rhodfbot~as Boolean connectives are missing in RDFS.}


\vspace{1em}
\nd {\bf Future Work.} Concerning future work,  there are still some extensions that are worth to be addressed within our framework. 
In particular, \ii{i}
we would like to investigate how to extend defeasible triples also to other predicates of the \rhodf~vocabulary beyond $\subc$ and $\subp$ and possibly involving blank nodes in case it makes sense to do so; \ii{ii}
another point deals with the computational complexity of our framework. In particular, we would like to see whether an approach similar as described in \cite{Munoz09} can be applied to our context as well, as explained in Remark~\ref{remderiv}, and we want to address the problem of conjunctive query answering; and \ii{iii} last, but not least, we want to implement our proposal. In particular, to scale up in practice, we are  interested in working out parallelisation of computation, a non-trivial problem for non-monotonic reasoning, that has already been investigated for some of the approaches based on the use of a top-layer of rules \cite{TachmazidisEtAl2012}.
\nocite{CormenEtAl01}

\section*{Acknowledgment}
\nd This research was  supported by TAILOR, a project funded by EU Horizon 2020 research and innovation programme under GA No 952215. 

 \bibliographystyle{abbrvnat}
 \bibliography{references,refsEL,mybiblio2,bibliononmonotonerdf,nets}

\appendix

\section{Proofs for Section \ref{sec:preliminaries}}

\sound*

\vspace{0.2cm}
\begin{proof}

$G$ is satisfiable, hence let $\I =\tuple{\Delta_{\DR}, \Delta_{\DP}, \Delta_{\DC},\Delta_{L}, \intP{\cdot}, \intC{\cdot}, \int{\cdot}}$ be a model of $G$ ($\I\rdfsat G$). That is, $\I $ satisfies all the conditions in Definition \ref{satisfaction}. We have to prove that $G\rdfent H$, that is, $\I\rdfsat H$.

We consider only the rules (5)-(7). The theorem has already been proved for groups of rules (1)-(4) in \cite[Lemma 31]{Munoz09}.\footnote{In \cite[Lemma 31]{Munoz09} the authors consider a stronger system in which the predicates $\subclass$ and $\spp$ are reflexive. We drop such properties, hence dropping the corresponding groups (6) and (7) of derivation rules in \cite[Table 1]{Munoz09}. Hence we need to follow the proof of Lemma 31 in \cite{Munoz09} only until the point (4) included.
}
\begin{description}
\item[Rule $(5a)$.]
Let $\triple{c,\disjC,d}\in R$ for some $R\subseteq G$, $R'=R\cup\{\triple{d,\disjC,c}\}$, obtained via the application of rule $(5a)$, and $H=G\cup R'$.
We have that for every model $\I$ of $G$, $\I\rdfsat R$, since $R\subseteq G$.  Therefore, $\I$ satisfies $\triple{c,\disjC,d}$, and, since it is a model of $G$ and is symmetric on $\disjC$ (see Definition \ref{satisfaction}), we have that  $(\int{d},\int{c})\in\intP{\int{\disjC}}$. That is, $\I$ satisfies $\triple{d,\disjC,c}$. Hence, from  $\I\rdfsat R'$ and $\I\rdfsat G$, $\I\rdfsat H$ follows.

\item[Rule $(5b)$.] 
Let $\triple{c,\disjC,d}$ and $\triple{e,\subclass,c}$ be in $R$, for some $R\subseteq G$. Consider $R'=R\cup\{\triple{e,\disjC,d}\}$, obtained via the application of rule $(5b)$, and $H=G\cup R'$.
We have that for every model $\I$ of $G$, $\I\rdfsat R$, since $R\subseteq G$.  $\I$ satisfies $\triple{c,\disjC,d}$ and $\triple{e,\subclass,c}$, and, since it is a model of $G$, by sc-transitivity of $\I$,  $(\int{e},\int{d})\in\intP{\int{\disjP}}$ follows. That is, $\I$ satisfies $\triple{e,\disjC,d}$. Hence, since  $\I\rdfsat R'$ and $\I\rdfsat G$, we have that 
$\I\rdfsat H$.

\item[Rule $(5c)$.]
Let $\triple{c,\disjC,c}\in R$ for some $R\subseteq G$, $R'=R\cup\{\triple{c,\disjC,d}\}$, obtained via the application of rule $(5c)$, and $H=G\cup R'$.
We have that for every model $\I$ of $G$, $\I\rdfsat R$, since $R\subseteq G$.  Therefore, $\I$ satisfies $\triple{c,\disjC,c}$, and, since it is a model of $G$ and is c-exhaustive 
on $\disjC$ (see Definition \ref{satisfaction}), we have that  $(\int{c},\int{d})\in\intP{\int{\disjC}}$. That is, $\I$ satisfies $\triple{c,\disjC,d}$. Hence, from  $\I\rdfsat R'$ and $\I\rdfsat G$, $\I\rdfsat H$ follows.

\item[Rules $(6a), (6b)$ and $(6c)$.] The argument is analogous to rules $(5a), (5b)$ and $(5c)$

\item[Rule $(7a)$.]
Let $\triple{p,\dom,c}$, $\triple{q,\dom,d}$, and $\triple{c,\disjC,d}$ be in $ R$ for some $R\subseteq G$, $R'=R\cup\{\triple{p,\disjP,q}\}$, obtained via the application of rule $(7a)$, and $H=G\cup R'$.
We have that for every model $\I$ of $G$, $\I\rdfsat R$, since $R\subseteq G$.  $\I$ satisfies $\triple{p,\dom,c}$, $\triple{q,\dom,d}$, and $\triple{c,\disjC,d}$, that by condition 1 of Disjointness II implies $\I\rdfsat \triple{p,\disjP,q}$.

\item[Rule $(7b)$.]
As for rule $(7a)$, just by referring to condition 2 of Disjointness II instead of condition 1 of Disjointness II.

\end{description}
\end{proof}

\vspace{0.4cm}

\completefirst*

\vspace{0.2cm}

\begin{proof}
We need to prove that $\I_G$ satisfies the constraints in Definition \ref{satisfaction}. 
At first, note that for the conditions {\bf Simple} to {\bf Typing II}, the proof corresponds to the proof of Lemma 32 in~\cite{Munoz09}\footnote{With the minor difference that in \cite[Lemma32]{Munoz09} the authors impose also reflexivity to the interpretations of the predicates $\subclass$ and $\spp$ and consider the associated derivation rules, while here we do not.
}. 
So, let us verify the remaining conditions.

  \begin{description}

 \item[Disjointness I:] \ 
  \begin{enumerate}
\setlength{\itemindent}{-7mm}
 \item If $(c,d)\in\intP{\intG{\disjC}}$, then $c,d\in\Delta_{\DC}$. This holds by  construction of $\Delta_{\DC}$.
 
 \item $\intP{\intG{\disjC}}$ is symmetric, \emph{sc}-transitive and \emph{c}-exhaustive over $\Delta_{\DC}$.
 
\emph{Symmetry}: Let $(c,d) \in \intP{\intG{ \disjC}} = \intP{ \disjC}$. By construction of $\I_G$ we have that $\triple{c, \disjC, d} \in\clos(G)$, and we also have that
$c,d\in \Delta_\DC$. Due to the closure under rule (5a), $\triple{d, \disjC, c} \in\clos(G)$, and by construction of $\intP{\cdot}$, $(d,c) \in \intP{\intG{ \disjC}}$.
 
\emph{\emph{sc}-transitivity}: Let $(c, d) \in \intP{\intG{ \disjC}} = \intP{ \disjC}$ and $(e, c) \in \intP{\intG{ \subclass}} = \intP{ \subclass}$. By construction of $\I_G$
we have that  $\triple{c, \disjC, d}, \triple{e, \subclass, c} \in\clos(G)$, and we also have that $c,d,e\in \Delta_\DC$. Due to the closure under rule (5b), $\triple{e, \disjC, d} \in\clos(G)$, and by construction of $\intP{\cdot}$, $(e,d) \in \intP{\intG{ \disjC}}$.

\emph{\emph{c}-exhaustive}: Let $(c,c) \in \intP{\intG{ \disjC}} = \intP{ \disjC}$. By construction of $\I_G$ we have that $\triple{c, \disjC, c} \in\clos(G)$ and
$c\in \Delta_\DC$. Due to the closure under rule (5c), $\triple{c, \disjC, d} \in\clos(G)$, and by construction of $\intP{\cdot}$, $(c,d) \in \intP{\intG{ \disjC}}$ with $d\in \Delta_\DC$.

%

\setlength{\itemindent}{-7mm}
 \item If $(p,q)\in\intP{\intG{\disjP}}$, then $p,q\in\Delta_{\DP}$. This holds by  construction of $\Delta_{\DP}$.
 
 \item $\intP{\intG{\disjP}}$ is symmetric, \emph{sp}-transitive and \emph{p}-exhaustive over $\Delta_{\DP}$.
  The proof is a rephrasing of the proof of point 2.
 
%
%
%
%
\end{enumerate}

 \item[Disjointness II:] \ 
  \begin{enumerate}
\setlength{\itemindent}{-7mm}
 \item If $(p, c) \in \intP{\intG{\dom}}$, $(q, d) \in \intP{\intG{\dom}}$, and $(c, d) \in \intP{\intG{\disjC}}$, then $(p, q) \in \intP{\intG{\disjP}}$.
 
Indeed, let $(p, c) \in \intP{\intG{\dom}}$, $(q, d) \in \intP{\intG{\dom}}$, and $(c, d) \in \intP{\intG{\disjC}}$. By construction of $\intP{\cdot}$,
$\triple{p,\dom,c}$, $\triple{q,\dom,d}$, and $\triple{c,\disjC,d}$ are in $\clos(G)$. Moreover, $\clos(G)$ is closed under rule (7a) and, thus, $\triple{p,\disjP,q}\in\clos(G)$, and, by construction of $\intP{\cdot}$, $(p, q) \in \intP{\intG{\disjP}}$.

  \item If $(p, c) \in \intP{\intG{\range}}$, $(q, d) \in \intP{\intG{\range}}$, and $(c, d) \in \intP{\intG{\disjC}}$, then $(p, q) \in \intP{\intG{\disjP}}$.
  
  The proof is analogous to the previous point: we just need to refer to $\range$ and rule (7b) instead of $\dom$ and rule (7a), which concludes.
 
  \end{enumerate}

\end{description}
\end{proof}

\vspace{0.4cm}
\completesecond*
\vspace{0.2cm}

\begin{proof} 
The proof mirrors the proof of Lemma 33 in \cite{Munoz09}. In particular, consider the interpretation 
\[
\I_G =\tuple{\Delta_{\DR}, \Delta_{\DP}, \Delta_{\DC},  \Delta_{\DL}, \intP{\cdot}, \intC{\cdot}, \intG{\cdot}}
\]
\nd as defined in Lemma~\ref{completefirst}.
Therefore, as both $\I_G\rdfsat G$ and $G\rdfent H$ hold, we have $\I_G\rdfsat H$ by Definition \ref{entailment}.
Therefore,  for each $\triple{s,p,o}\in H$, $\intG{p}\in\Delta_\DP$
and $(\intG{s}, \intG{o})\in \intP{\intG{p}}$. Moreover, by construction $\intG{p} = p$, and $\intP{\intG{p}} = \intP{p} = \{(s, o)\mid \triple{s, p, o}\in\clos(G)\}$.
Finally, since $(\intG{s},\intG{o})\in \intP{\intG{p}}$, we have that $\triple{\intG{s}, \intG{p}, \intG{o})}\in\clos(G)$, \ie~$\triple{s, p, o} \in\clos(G)$, for each $\triple{s, p, o}\in H$. Therefore, $H\subseteq \clos(G)$, which concludes.
\end{proof} 

\section{Proofs for Section \ref{defrdfs}}

\propuniqueminimal*
\vspace{0.2cm}

\begin{proof}
Assume to the contrary that $|\min_\preceq (\RG)| > 1$, \ie~there are $\R=(\M,r)$, $\R'=(\M,r')$, with $\R,\R'\in \min_\preceq (\RG)$ and $\R \neq \R'$. 

It cannot be the case that $\R\preceq \R'$ and $\R'\preceq \R$, since that would imply that $\R=\R'$. Hence it must be the case that $\R$ and $\R'$ are incomparable w.r.t. $\preceq$;  that is, there are at least two interpretations $\I,\J$ in $\M$ s.t. $r(\I)< r'(\I)$ and $r'(\J)< r(\J)$.

Now, consider $\R^\star = (\M,r^\star)$, with $r^\star(\I)=\min\{r(\I), r'(\I)\}$ for every $\I\in \M$. Clearly, $\R^\star \prec \R$ and $\R^\star \prec \R'$. 
As next, let us prove that  $\R^\star \rdfsat G$. 
At first, note that $\R^\star \rdfsat \Gclass$ holds as the satisfaction of $\Gclass$ depends only on $\M$ and not on the ranks. 
At second, assume to the contrary that $\R^\star \not\rdfsat \Gdef$, that is, there is a defeasible triple $\dtriple{s,p,o} \in G$ such that $\R^\star\not\rdfsat \dtriple{s,p,o}$ and let's assume that $\dtriple{s,p,o}$ is of the form $\dtriple{c,\subclass,d}$ (the proof for the case $\dtriple{p,\spp,q}$  is similar).
That means that there is $\bar{\I} \in\cmin(c,\R^\star)$ s.t. $\bar{\I}\not\rdfsat \triple{c,\subclass,d}$. 
But, by the definition of $r^\star$, either $\bar{\I} \in\cmin(c,\R)$ or $\bar{\I} \in\cmin(c,\R')$ and, thus, either $\R\not\rdfsat \dtriple{c,\subclass,d}$ or $\R'\not\rdfsat \dtriple{c,\subclass,d}$ holds,
against the hypothesis that both $\R$ and $\R'$ are ranked models of $G$. As a consequence, $\R^\star \rdfsat \Gdef$ has to hold and, thus,  $\R^\star \rdfsat G$. That is, 
 $\R^\star \in \RG$. But then, it can not be the case that $\R,\R'\in \min_\preceq (\RG)$ as there is $\R^\star \in \RG$ with $\R^\star \prec \R$ and $\R^\star \prec \R'$, which is against our hypothesis. Therefore, there can not be two distinct ranked models $\R,\R'$ in $\min_\preceq (\RG)$ and, as $\RG$ is not empty, $|\min_\preceq (\RG)| = 1$ has to hold.
\end{proof}

\vspace{0.4cm}
\propexceptionality*
\vspace{0.2cm}

\begin{proof}
Let $t$ be \ssc-exceptional. 
\begin{description}
\item[$\Rightarrow)$] Immediate from the definition of exceptionality. 

\item[$\Leftarrow)$]  We proceed by contradiction: let $t$ be \ssc-exceptional w.r.t. $\CG$, and assume that there is a model $\R=(\M,r_{\R})\in\RG$ s.t. $t$ is not \ssc-exceptional w.r.t. $\R$.
That is, there is an $\I\in\M$ s.t. $r_{\R}(\I)=0$ and $\I\not\rdfsat \triple{t,\disjC, t}$. In such a case, by Definition \ref{Def_presord} and Proposition \ref{Propuniqueminimal}, we would have that $\I$ has rank $0$ also in $\CG$. Consequently $t$ cannot be \ssc-exceptional w.r.t. $\CG$, which contradicts our assumption.
\end{description}

\nd The proof is analogous if $t$ is \ssp-exceptional.
\end{proof}

\vspace{0.4cm}
\lemmatriplesproofsc*

\vspace{0.2cm}
\begin{proof}
The proof is on induction on the depth $d(T)$ of $T$, where by assumption, $T$ has $\triple{p,\subclass,q}$ as root. 

\begin{description}
    \item[Case $d(T)=0$.] Hence, the tree's only node is $\triple{p,\subclass,q}$, which concludes. 
    \item[Case $d(T)=1$.] In this case, there is only one possible tree, obtained by instantiating rule (3a):
    \[
      \frac{\triple{p,\subclass,r}, \triple{r,\subclass,q}}{\triple{p,\subclass,q}}
     \]
{\bf Remark.} 
\emph{Note that we cannot instantiate rule (2b) in the form
\[
 \frac{\triple{r,  \spp, \subclass},  \triple{p, r, q}}{\triple{p, \subclass, q}} \ , 
\]
\nd as $\triple{r,  \spp, \subclass}$ is not allowed to occur in our language (see Section~\ref{sec:preliminaries}).}
 
 \item[Case $d(T)= n+1$.] 
Let us assume that the lemma holds for all proof trees of depth $m \leq n$, with $n\geq 1$. Let us show that it holds also for the case $d(T)= n+1$ as well.

Note that the tree $T$ of depth $n+1$ with root $\triple{p,\subclass,q}$ can only be built by taking two trees $T_1$ and $T_2$ that have as roots triples of the form $\triple{A_i,\subclass,B_i}$ ($i=1,2$) with $\max(d(T_1), d(T_2))=n$, and applying to their roots rule (3a). 
Therefore, by construction of $T$ and by induction on $T_i$, also the tree $T$ of depth $n+1$ contains only triples of the form $\triple{A,\subclass,B}$, which concludes.
\end{description}
\end{proof}

\vspace{0.4cm}
\lemmatriplesproofbotc*
\vspace{0.2cm}

\begin{proof}
The proof is on induction on the depth $d(T)$ of $T$, where by assumption, $T$ has $\triple{p,\disjC,q}$ as root.

\begin{description}
    \item[Case $d(T)=0$.] Hence, the tree's only node is  $\triple{p,\disjC,q}$, which concludes.
    \item[Case $d(T)=1$.] In this case, there are only three possible trees, obtained by instantiating rules $(5a)$, $(5b)$ or $(5c)$: namely,
\[
\frac{\triple{q,\disjC,p}}{\triple{p,\disjC,q}} \ ,
\frac{\triple{s,\disjC,q}, \triple{p,\subclass,s}}{\triple{p,\disjC,q}}  \text{ \ or \  }
\frac{\triple{p,\disjC,p}}{\triple{p,\disjC,q}}  \ . 
\]
\nd In all three cases the lemma is satisfied, which concludes.

 \item[Case $d(T)= n+1$.] 
Let us assume that the lemma holds for all proof trees of depth $m \leq n$, with $n\geq 1$. Let us show that it holds also for the case $d(T)= n+1$ as well.
 
Note that the tree $T$ of depth $n+1$ with root $\triple{p,\disjC,q}$ can only be built in three ways:
\begin{itemize}
    \item by applying rule (5a) to a tree $T_1$ of depth $n$ having as root a triple of form $\triple{A,\disjC,B}$;
    \item by applying the rule (5b) to two trees $T_2$ and $T_3$, with $\max(d(T_2), d(T_3))=n$, having as root, respectively, a triple of form $\triple{A,\disjC,B}$ and a triple of form $\triple{A,\subclass,B}$;
        \item by applying rule (5c) to a tree $T_4$ of depth $n$ having as root a triple of form $\triple{A,\disjC,A}$;
\end{itemize}

\nd Now, by construction of $T$, by induction hypothesis on $T_1,T_2, T_4$ and by Lemma~\ref{lemma:triples_proof_sc} applied to $T_3$,
also the tree $T$ of depth $n+1$ contains only triples of the form $\triple{A,\subclass,B}$ and $\triple{A,\disjC,B}$, which concludes.
\end{description}
\end{proof}

\vspace{0.4cm}
\lemmaminimalinterpretations*
\vspace{0.2cm}

\begin{proof}
The proof is immediate from the fact that every $\R$ in $\RG$ is a model of $G$ and Definition \ref{defsat}. In fact, if there is  $\I\in\M$ s.t. $r(\I)=0$ and $\I$ does satisfy neither $\triple{p,\subclass,q}$  nor $\triple{p,\disjC,p}$, then $\R\not\rdfsat\dtriple{p,\subclass,q}$ and, thus, $\R$ is not a model of $G$, against the hypothesis.

The proof for $\dtriple{p,\spp,q}\in \Gdef$ is similar.
\end{proof}

\vspace{0.4cm}
\lemmaproofsubclass*

\vspace{0.2cm}

\begin{proof}
We  prove the first half, involving the subclass predicate.

So, assume $\Gcount \deriv \triple{p,\subclass,q}$ and let $\I$ be a $\rhodfbot$ interpretation in $\M$ that has rank $0$.  
The proof is on induction on the depth $d(T)$ of a tree $T$, where $T$ has $\triple{p,\subclass,q}$ as root.

\begin{description}
    \item[Case $d(T)=0$.] Hence, the tree's only node is  $\triple{p,\subclass,q}$. Therefore, either $\triple{p,\subclass,q}\in \Gclass$ or $\dtriple{p,\subclass,q}\in \Gdef$. In the former case $\I\rdfsat\triple{p,\subclass,q}$, as, being $\R$ a model of $G$, every $\I\in\M$ must satisfy $\Gclass$. In the latter case, by Lemma \ref{lemma:minimal_interpretations}, either $\I\rdfsat\triple{p,\subclass,q}$ or $\I\rdfsat\triple{p,\disjC,p}$, which concludes.
    
\item[Case $d(T)=1$.]  In this case, there is only one possible tree, obtained by instantiating rule (3a):
\[
\frac{\triple{p,\subclass,r}, \triple{r,\subclass,q}}{\triple{p,\subclass,q}}
\]
\nd Assume $\I\not\rdfsat\triple{p,\subclass,q}$. As $\subclass$ is a transitive relation, we have two possibilities only:

\begin{description}
    \item[Case $\I\not\rdfsat\triple{p,\subclass,r}$.] Then, since $\triple{p,\subclass,r}\in\Gcount$, it must be the case that  $\dtriple{p,\subclass,r}\in\Gdef$ and, thus, by Lemma~\ref{lemma:minimal_interpretations}, $\I\rdfsat\triple{p,\disjC,p}$.
    \item[Case $\I\rdfsat\triple{p,\subclass,r}$ but $\I\not\rdfsat\triple{r,\subclass,q}$.] Then, since $\triple{r,\subclass,q}\in\Gcount$, it must be the case that  $\dtriple{r,\subclass,q}\in\Gdef$. By Lemma \ref{lemma:minimal_interpretations}, $\I\rdfsat\triple{r,\disjC,r}$. So we have $\I\rdfsat\triple{p,\subclass,r}$ and $\I\rdfsat\triple{r,\disjC,r}$. Given the derivation rule (EmptySC') and the fact that $\deriv$ is sound, we can conclude that  $\I\rdfsat\triple{p,\disjC,p}$.
\end{description}
\nd Therefore, either $\I\rdfsat\triple{p,\subclass,q}$ or $\I\rdfsat\triple{p,\disjC,p}$, which concludes.

\item[Case $d(T)= n+1$.] 
Let us assume that the lemma holds for all proof trees of depth $m \leq n$, with $n\geq 1$. Let us show that it holds also for the case $d(T)= n+1$ as well, where $T$ has $\triple{p,\subclass,q}$ as root. Now, since $d(T)>1$, as for  case $d(T)=1$, the only possibility is that the tree terminates with an instantiation of  rule (3a): that is, 
\[
\frac{\triple{p,\subclass,r}, \triple{r,\subclass,q}}{\triple{p,\subclass,q}}
 \]         

where $\triple{p,\subclass,r}$ and  $\triple{r,\subclass,q}$ are, respectively, the roots of trees $T_1$ and $T_2$, the immediate subtrees of $T$, with $\max(d(T_1), d(T_2))=n$.
By inductive hypothesis on $T_i$, either $\I\rdfsat \triple{p,\subclass,r}$ or $\I\rdfsat \triple{p,\disjC,p}$, and $\I\rdfsat \triple{r,\subclass,q}$ or $\I\rdfsat \triple{r,\disjC,r}$.  

As a consequence, we have three cases:
\begin{description}
    \item[Case $\I\rdfsat \triple{p,\disjC,p}$.] The lemma is satisfied immediately.
    \item[Case $\I\rdfsat \triple{p,\subclass,r}$ and $\I\rdfsat \triple{r,\disjC,r}$.] Then, given the derivation rule (EmptySC') and the fact that $\deriv$ is sound, we have  $\I\rdfsat\triple{p,\disjC,p}$, which concludes.
    \item[Case $\I\rdfsat \triple{p,\subclass,r}$ and $\I\rdfsat  \triple{r,\subclass,q}$.] Then, by the transitivity of the interpretation of $\subclass$, we have $\I\rdfsat \triple{p,\subclass,q}$, which concludes.
\end{description}

\end{description}

\nd For the second half of the lemma, involving the predicate $\spp$, the proof has exactly the same structure, with the only difference that it refers to the rule (2a) instead of the rule (3a).
\end{proof}

\vspace{0.4cm}
\lemmacexceptionality*

\vspace{0.2cm}
\begin{proof}
Let $G$ be a defeasible graph s.t. $\Gcount\deriv \triple{p,\disjC,q}$. Then by Proposition \ref{prop:proof_trees}, there is a $\rhodfbot$ proof tree $T$ from $H$ to 
$\triple{p,\disjC,q}$ for some $H\subseteq \Gcount$. We prove now the lemma by induction on the depth $d(T)$ of $T$. 

Given $\CG=(\M,r)$, we need to prove that $\triple{p,\disjC,q}$ is satisfied by every $\I\in \M$ s.t. $r(\I)=0$. Note that, as $\CG$ is a model of $G$, for every $\I\in \M$, $\I\rdfsat \Gclass$ holds (see Definition \ref{defsat}).

\begin{description}
  \item[Case $d(T)=0$.]  The tree's only node is  $\triple{p,\disjC,q}$, that is in $\Gcount$. Recall  that $\Gcount$ is the union of $\Gclass$ and the strict translation of the defeasible triples in $\Gdef$ (Equation \ref{eq:Gs}). Note that $\triple{p,\disjC,q}$ cannot be the strict form of a triple in $\Gdef$, as the defeasible triples must contain $\subclass$ or $\spp$ as second element (Definition \ref{def:defeasible_triple}), hence $\triple{p,\disjC,q}\in \Gclass$. 
Being $\CG$ a model of $G$, its strict part $\Gclass$ must be satisfied by every $\I\in\M$ and, as a consequence, the triple $\triple{p,\disjC,q}$ must be satisfied by every $\I\in \M$ with rank $0$, which concludes.
    
  \item[Case $d(T)=1$.]  There are only three possible trees of depth 1 with a triple $\triple{p,\disjC,q}$ obtained by instantiating rules $(5a)$, $(5b)$ or $(5c)$:
    

\[
\frac{\triple{q,\disjC,p}}{\triple{p,\disjC,q}} ,
\frac{\triple{s,\disjC,q}, \triple{p,\subclass,s}}{\triple{p,\disjC,q}}  \text{ \ or \  }
\frac{\triple{p,\disjC,p}}{\triple{p,\disjC,q}}  \ .
\]
In the first case,  we can refer to the case $d(T)=0$ and the fact that $\I$ must satisfy the symmetry of $\disjC$. The third case is similar to the first one taking into account that 
$\I$ must be \emph{c}-exhaustive on $\disjC$.  In the second case the tree consists of an instantiation of rule (5b), that has two premises, $\triple{s,\disjC,q}$ and $\triple{p,\subclass,s}$, that must both be in $\Gcount$.  $\triple{s,\disjC,q}$ does not have a correspondent defeasible triple, so $\triple{s,\disjC,q}\in \Gclass$. Since $\Gclass$ must be satisfied by every $\rhodf$-interpretation in a ranked model of $G$, we have $\I\rdfsat \triple{s,\disjC,q}$. 

Concerning the premise $\triple{p,\subclass,s}$, we have two possible cases:

\begin{description}
    \item[Case $\I\rdfsat\triple{p,\subclass,s}$.] In this case, from $\I\rdfsat \triple{s,\disjC,q}$ and the soundness of $\deriv$, we can conclude $\I\rdfsat\triple{p,\disjC,q}$.
    \item[Case $\I\not\rdfsat\triple{p,\subclass,s}$.] In this case,  $\triple{p,\subclass,s}\in \Gcount$ and $\I\not\rdfsat\triple{p,\subclass,s}$ implies that $\triple{p,\subclass,s}\notin \Gclass$, so $\dtriple{p,\subclass,s}\in \Gdef$ must hold. By Lemma \ref{lemma:minimal_interpretations}, $\I\not\rdfsat\triple{p,\subclass,s}$, implies that $\I\rdfsat \triple{p,\disjC,p}$. Now, by rule $(5c)$  and the soundness of $\deriv$, we can conclude $\I\rdfsat \triple{p,\disjC,q}$. 
\end{description}

\item[Case $d(T)=n+1$.] Let us assume that the proposition holds for all the proof trees of depth $m \leq n$, with $n\geq 1$. Let us show that it holds also for the case $d(T)= n+1$ as well, where $T$ has $\triple{p,\disjC,q}$ as root.  
Then, since $d(T)>1$, as for the case  $d(T)=1$ the only three possibilities are that the tree terminates with an application rule $(5a), (5b)$ or $(5c)$: that is, 
%
%
\[ 
\frac{\triple{q,\disjC,p}}{\triple{p,\disjC,q}} ,
\frac{\triple{s,\disjC,q}, \triple{p,\subclass,s}}{\triple{p,\disjC,q}}  \text{ \ or \  }
\frac{\triple{p,\disjC,p}}{\triple{p,\disjC,q}}  \ .
\]
The first case is straightforward, since the immediate subtree of $T$ would be a tree of depth $n$ that has $\triple{q,\disjC,p}$ as root: by inductive hypothesis $\I\rdfsat \triple{q,\disjC,p}$ and, since $\I$ must satisfy the symmetry of $\disjC$, we have $\I\rdfsat \triple{p,\disjC,q}$. The third case can be proven similarly to the first case by relying on fact that $\I$ is \emph{c}-exhaustive over $\Delta_{\DC}$. In the second case   $\triple{s,\disjC,q}$ and  $\triple{p,\subclass,s}$ are, respectively, the roots of trees $T_1$ and $T_2$, the immediate subtrees of $T$, with $\max(d(T_1), d(T_2))=n$.
%
%
By inductive hypothesis on $T_1$, $\I\rdfsat\triple{s,\disjC,q}$. Concerning $T_2$ and its root $\triple{p,\subclass,s}$ we have two possible cases:

\begin{description}
    \item[Case $\I\rdfsat \triple{p,\subclass,s}$.] In this case, by rule (5b) and the soundness of $\deriv$ we conclude $\I\rdfsat\triple{p,\disjC,q}$.
    \item[Case $\I\not\rdfsat \triple{p,\subclass,s}$.] In this case, by the second part of Lemma \ref{lemma:proof_subclass} we have $\I\rdfsat \triple{p,\disjC,p}$. 
    Now, by rule   $(5c)$ and the soundness of $\deriv$, we can conclude $\I\rdfsat \triple{p,\disjC,q}$.
\end{description}
\end{description}
\nd This concludes the proof of then lemma.
\end{proof}

\vspace{0.4cm}
\lemmapexceptionality*

\vspace{0.2cm}
\begin{proof}
Let $G$ be a defeasible graph s.t. $\Gcount\deriv \triple{p,\disjP,q}$. Then by Proposition \ref{prop:proof_trees} there must be a $\rhodfbot$ proof tree $T$ deriving $\triple{p,\disjP,q}$ from some graph $H\subseteq \Gcount$. We prove now the lemma by induction on the depth $d(T)$ of $T$. 

Given $\CG=(\M,r)$, we need to prove that $\triple{p,\disjP,q}$ is satisfied by every $\I\in \M$ s.t. $r(\I)=0$. Note that, as $\CG$ is a model of $G$, for every $\I\in \M$, $\I\rdfsat \Gclass$ holds (see Definition \ref{defsat}).
\begin{description}
    \item[Case $d(T)=0$.] The tree's only node is  $\triple{p,\disjP,q}$, that is in $\Gcount$. Recall  that $\Gcount$ is the union of $\Gclass$ and the strict translation of the defeasible triples in $\Gdef$ (Equation \ref{eq:Gs}). Note that $\triple{p,\disjP,q}$ cannot be the strict form of a triple in $\Gdef$, since the defeasible triples must contain $\subclass$ or $\spp$ as second element (Definition \ref{def:defeasible_triple}), hence $\triple{p,\disjP,q}\in \Gclass$. 
Being $\CG$ a model of $G$, its strict part $\Gclass$ must be satisfied by every $\I\in\M$ (see Definition \ref{defsat}), and as a consequence the triple $\triple{p,\disjP,q}$ must be satisfied by every $\I\in \M$ with rank $0$, which concludes.
    
\item[Case $d(T)=1$.]  There are five possible trees of depth 1 with a triple $\triple{p,\disjP,q}$ as root, obtained by instantiating rule $(6a), (6b), (6c)$, $(7a)$ or $(7b)$: \ie

\begin{tabular}{ll}
     (1)  & {\large $\frac{\triple{q,\disjP,p}}{\triple{p,\disjP,q}}$}  \\ \\
     (2)  & {\large $\frac{\triple{s,\disjP,q}, \triple{p,\spp,s}}{\triple{p,\disjP,q}}$} \\ \\
     (3)  & {\large $\frac{\triple{p,\disjP,p}}{\triple{p,\disjP,q}}$}  \\ \\
     (4)  & {\large $\frac{\triple{p,\dom,r}, \triple{q,\dom,s}, \triple{r,\disjC,s}}{\triple{p,\disjP,q}}$}  \\ \\
     (5)   & {\large $\frac{\triple{p,\range,r}, \triple{q,\range,s}, \triple{r,\disjC,s}}{\triple{p,\disjP,q}}$} \ .
\end{tabular}

\begin{description}
\item[Case (1).] We can refer to the case $d(T)=0$ and the fact that $\I$ must satisfy the symmetry of $\disjP$. 

\item[Case (3).] This case is similar to Case $(1)$ by referring to the fact that $\I$ is \emph{c}-exhaustive.

\item[Case (2).]  The tree consists of an instantiation of rule (6b), that has two premises, $\triple{s,\disjP,q}$ and $\triple{p,\spp,s}$, that must both be in $\Gcount$.  $\triple{s,\disjP,q}$ does not have a correspondent defeasible triple, so $\triple{s,\disjP,q}\in \Gcount$ implies $\triple{s,\disjP,q}\in \Gclass$. Since $\Gclass$ must be satisfied by every $\rhodf$-interpretation in a ranked model of $G$, we have $\I\rdfsat \triple{s,\disjP,q}$.

Concerning the premise $\triple{p,\spp,s}$, we have two possible cases:

\begin{description}
    \item[Case $\I\rdfsat\triple{p,\spp,s}$.]  In this case, from $\I\rdfsat\triple{p,\spp,s}$, $\I\rdfsat \triple{s,\disjP,q}$, and the soundness of $\deriv$, we can conclude $\I\rdfsat\triple{p,\disjP,q}$.
    \item[Case $\I\not\rdfsat\triple{p,\spp,s}$.] In this case, the situation in which $\triple{p,\spp,s}\in \Gcount$ and $\I\not\rdfsat\triple{p,\spp,s}$ is possible only if $\triple{p,\spp,s}\in \Gdef$. By Lemma \ref{lemma:minimal_interpretations}, we have  $\I\rdfsat \triple{p,\disjP,p}$. 
 Now, by rule $(6c)$ and the soundness of $\deriv$ we can conclude $\I\rdfsat \triple{p,\disjP,q}$. 
\end{description}

\item[Case (4).]  $\triple{p,\dom,r}, \triple{q,\dom,s}, \triple{r,\disjC,s}$ are in $\Gcount$, and since they cannot have a defeasible version, they must be in $\Gclass$ too. All the triples in $\Gclass$ must be satisfied by every $\I\in\M$. This, together with the soundness of $\deriv$, guarantees that $\I\rdfsat\triple{p,\disjP,q}$.

\item[Case (5).]  Exactly as Case (4),  just consider the triples of form $\triple{A,\range,B}$ instead of the triples $\triple{A,\dom,B}$.

\end{description}

\item[Case $d(T)=n+1$.] Let us assume that the proposition holds for all the proof trees of depth $m \leq n$, with $n\geq 1$. Let us show that it holds also for the case $d(T)= n+1$ as well, where $T$ has $\triple{p,\disjP,q}$ as root. Then, since $d(T)>1$, as for the case with $d(T)=1$ there are five possibilities, with the tree terminating with an application of the rule $(6a), (6b), (6c)$, $(7a)$ or $(7b)$: namely,

\begin{tabular}{ll}
     (1)  & {\large $\frac{\triple{q,\disjP,p}}{\triple{p,\disjP,q}}$}  \\ \\
     (2)  & {\large $\frac{\triple{s,\disjP,q}, \triple{p,\spp,s}}{\triple{p,\disjP,q}}$} \\ \\
     (3)  & {\large $\frac{\triple{p,\disjP,p}}{\triple{p,\disjP,q}}$}  \\ \\
     (4)  & {\large $\frac{\triple{p,\dom,r}, \triple{q,\dom,s}, \triple{r,\disjC,s}}{\triple{p,\disjP,q}}$}  \\ \\
     (5)   & {\large $\frac{\triple{p,\range,r}, \triple{q,\range,s}, \triple{r,\disjC,s}}{\triple{p,\disjP,q}}$} \ .
\end{tabular}     
          
\begin{description}
\item[Case (1).] Straightforward, as the immediate subtree of $T$ is a tree of depth $n$ that has $\triple{q,\disjP,p}$ as root: by inductive hypothesis $\I\rdfsat \triple{q,\disjP,p}$ and, since $\I$ must satisfy the symmetry of $\disjP$, we have $\I\rdfsat \triple{p,\disjP,q}$. 

\item[Case (3).] The proof is as for Case $(1)$ by referring to the fact that $\I$ is \emph{c}-exhaustive.

\item[Case (2).]   $\triple{s,\disjP,q}$ and  $\triple{p,\spp,s}$ are, respectively, the roots of $T_1$ and $T_2$, the immediate subtrees of $T$, with $\max(d(T_1), d(T_2))=n$.
By inductive hypothesis on $T_1$, $\I\rdfsat\triple{s,\disjP,q}$. Concerning $T_2$ and its root $\triple{p,\spp,s}$, we have two possible cases:
\begin{description}
    \item[Case $\I\rdfsat \triple{p,\spp,s}$.] In this case, from $\I\rdfsat\triple{s,\disjP,q}$ and by the soundness of $\deriv$ we can conclude $\I\rdfsat\triple{p,\disjP,q}$.
    \item[Case $\I\not\rdfsat \triple{p,\spp,s}$.] In this case, by Lemma \ref{lemma:proof_subclass} we have $\I\rdfsat \triple{p,\disjP,p}$.
Now, by rule  $(6c)$ and the soundness of $\deriv$ we can conclude $\I\rdfsat \triple{p,\disjP,q}$.
\end{description}

\item[Case (4).]   Please note that for every pair of terms $p,q$, $\Gcount\deriv \triple{p,\dom,q}$ holds only if $\triple{p,\dom,q}\in \Gcount$: no triple of form $\triple{A,\dom,B}$ can be derived, since there are no rules in \rhodfbot that have triples of form $\triple{A,\dom,B}$ as conclusions. The only possibility could be the rule (2b), by substituting $E$ with $\dom$, but in that case in the premises we would have a triple with $\dom$ in the third position, that is not acceptable in our language (see Section \ref{sec:rdf-syntax}).

Since the triples $\triple{A,\dom,B}$ do not have a defeasible version, $\triple{p,\dom,q}\in \Gcount$ implies that $\triple{p,\dom,q}\in \Gclass$, and consequently $\I\rdfsat\triple{p,\dom,q}$. It follows that in case the tree $T$ terminates with an application of rule (7a), it will have three immediate subtrees: two trees will both have depth $0$ and will consist, respectively, only of the nodes $\triple{p,\dom,r}$ and $\triple{q,\dom,s}$; the third subtree, called  $T'$, will have  $\triple{r,\disjC,s}$ as root. 

We know from Lemma \ref{lemma:c-exceptionality} applied to $T'$ that $\I\rdfsat\triple{r,\disjC,s}$. So we have that $\I\rdfsat\triple{p,\dom,r}$, $\I\rdfsat\triple{q,\dom,s}$, $\I\rdfsat\triple{r,\disjC,s}$, and, by the soundness of $\deriv$, we can conclude $\I\rdfsat\triple{p,\disjP,q}$.

\item[Case (5).]   The proof for this case is analogous to Case (4): it is sufficient to substitute $\dom$ with $\range$.
\end{description}
\end{description}
\nd This concludes the proof of the lemma.
\end{proof}

\vspace{0.4cm}
\propexceptionalitysecond*
\vspace{0.2cm}

\begin{proof}
\begin{description}
\item[$\Rightarrow .)$] Let us show that if a term $t$ is \ssc-exceptional (resp., \ssp-exceptional) w.r.t. a defeasible graph $G$, then $\Gcount\deriv \triple{t,\disjC,t}$ (resp., $\Gcount\deriv \triple{t,\disjP,t}$).

Let $t$ be a term in $G$, and let $t$ be $\ssc$-exceptional (resp., $\ssp$-exceptional) w.r.t. $G$. Then $t$ is $\ssc$-exceptional (resp., $\ssp$-exceptional) w.r.t. $\CG=(\M, r)$, that is, for every $\I\in\M$ s.t. $r(\I)=0$, we have that $\I\rdfsat \triple{t,\disjC, t}$ (resp., $\I\rdfsat \triple{t,\disjP, t}$). Now, we need to prove that $G^s\deriv \triple{t,\disjC,t}$ (resp., $G^s\deriv \triple{t,\disjP,t}$).

Let $\I_{G^s}$ be the canonical model of $G^s$. By construction and  Lemma~\ref{completefirst}, $\I_{G^s}\in \M$. 
It is now sufficient to prove that $r(\I_{G^s})=0$. Let us proceed by contradiction by assuming that $r(\I_{G^s})>0$, and let $\R'=(\M, r')$ be the ranked interpretation obtained from $\CG$, with for every $\I\in\M$, 

\[r'(\I)=
\left\{\begin{array}{cc}
  0   &\text{if } \I=\I_{G^s} \\
  r(\I)   & \text{otherwise} \ .
\end{array}\right.
\]

We can easily check that $\R'$ is  a model of $G=\Gclass\cup\Gdef$:

\begin{itemize}
    \item $\R'$ is still a model of $\Gclass$, since the satisfaction of $\rhodfbot$-triples is not affected by the ranking function.
    \item for every $\dtriple{p,\subclass,q}\in \Gdef$, $\I_{G^s}\rdfsat\triple{p,\subc,q}$ since $\I_{G^s}$ is a model of $G^s$. Now we have two cases: 
    \begin{description}
        \item[$\I_{G^s}\rdfsat\triple{p,\disjC,p}$]. In such a case, by Definition \ref{defsat}, $\I_{G^s}\notin\cmin(p,\R')$ and $\I_{G^s}$ is irrelevant to decide the satisfaction of $\dtriple{p,\subc,q}$ in $\R'$. Consequently $\cmin(p,\R')=\cmin(p,\CG)$, and $\CG\rdfsat \dtriple{p,\subclass,q}$ implies $\R'\rdfsat \dtriple{p,\subclass,q}$.
        \item[$\I_{G^s}\not\rdfsat\triple{p,\disjC,p}$]. Then, since $r'(\I_{G^s})=0$ and $r'(\I)=r(\I)$ for all the other elements $\I$ of $\M$, $\R'\rdfsat\dtriple{p,\subc,q}$. Otherwise  in $\R'$  there should be an $\I'$ with rank $0$ s.t. $\I'\not\rdfsat\triple{p,\subc,q}$ and $\I'\not\rdfsat\triple{p,\disjC,p}$. But then also and $\CG$ should contain such $\I'$ at rank $0$, but that cannot be the case, as otherwise $\CG$ would not satisfy $\dtriple{p,\subclass,q}$, against the fact that $\CG$ is a model of $G$. 
    \end{description}
    Hence $\R'\rdfsat\dtriple{p,\subc,q}$.
    \item We proceed analogously to prove that for every $\dtriple{p,\subp,q}\in \Gdef$, $\R'\rdfsat\dtriple{p,\subp,q}$.
\end{itemize}

Therefore, $\R'$ is a model of $G$, but by Definition \ref{Def_presord} we also have $\R'\prec\CG$, against the definition of $\CG$. Consequently it must be the case that $r(\I_{G^s})=0$, that implies that, for every $t$, $\I_{G^s}\rdfsat\triple{t,\disjC,t}$ (resp., $\I_{G^s}\rdfsat\triple{t,\disjP,t}$).

Since $\I_{G^s}$ is the canonical model for $G^s$, $G^s\deriv \triple{t,\disjC,t}$ (resp., $G^s\deriv \triple{t,\disjP,t}$).

\item[$\Leftarrow .)$] Let us show that given a defeasible graph $G$, if $\Gcount\deriv \triple{t,\disjC,t}$ (resp., $\Gcount\deriv \triple{t,\disjP,t}$), then a term $t$ is \ssc-exceptional (resp., \ssp-exceptional) w.r.t. $G$.

\nd This is an immediate consequence of Corollaries \ref{coroll:c-exceptionality} and \ref{coroll:p-exceptionality}.
\end{description}
\nd This completes the proof.
\end{proof}

\vspace{0.4cm}
\LemmaCGi*

\vspace{0.2cm}
\begin{proof}
$\CG=(\M,r)$ is a model of $G$. Hence all the interpretations in $\M$ are models of $\Gclass$ and, since $\M^i\subseteq \M$, also $\CG^i$ satisfies $\Gclass$. 
Concerning the defeasible triples in $\Gdef_i$, we proceed by contradiction, assuming $\CG^i$ is not a model of $G^i$. So, let $\dtriple{p,\subc,q}\in\Gdef_i$ and  $\CG^i\not \rdfsat\dtriple{p,\subc,q}$, that is, there is a \rhodfbot-interpretation $\I$ s.t. $\I\in\cmin(p,\CG^i)$ and $\I\not \rdfsat\dtriple{p,\subc,q}$. Since $\hc(p)\geq i$, $\cmin(p,\CG^i)=\cmin(p,\CG)$, and consequently $\CG\not \rdfsat\dtriple{p,\subc,q}$, against the assumption that $\CG$ is the minimal model of $G$.  The case $\dtriple{p,\subp,q}\in\Gdef_i$ is proved similarly, which concludes.
\end{proof}

\vspace{0.4cm}
\LemmaminimalmodelCGi*

\vspace{0.2cm}
\begin{proof}
At first, we prove that $\R^*_i$ is a model of $G^i$. So, let $\CG$ be the minimal model of the graph $G$. From the definitions of $\CG$, $\CG^i$ and $\R^*_i$ it is clear that for every $\I\in \M$, $r^*_i(\I)<\infty$ iff $r(\I)<\infty$. Hence $\R^*_i\rdfsat \Gclass$. 
Now, let $\dtriple{p,\subc,q}\in\Gdef_i$. From the construction of $\Gdef_i$, $\CG^i$ and $R^*_i$, we have that for $\dtriple{p,\subc,q}\in\Gdef_i$ it holds that  $\hc(p)\geq i$. Therefore, since for every  $\I\in\M\setminus\M^i$ $r(\I)< i$, it must be the case that  $\I\rdfsat\triple{p,\disjC,p}$ and, thus,
for all the \rhodf-interpretations $\I\in\M\setminus\M^i$, $\I\rdfsat\triple{p,\disjC,p}$.

As a consequence, for every $\dtriple{p,\subc,q}\in\Gdef_i$, $\cmin(p,\CG)=\cmin(p,\CG^i)$, and since $\CG\rdfsat \dtriple{p,\subc,q}$ for every $\dtriple{p,\subc,q}\in\Gdef_i$, $\R^*_i\rdfsat \dtriple{p,\subc,q}$ has to hold too. 

Analogously, for every $\dtriple{p,\subp,q}\in\Gdef_i$, we have that
\begin{itemize}
    \item for all the \rhodf-interpretations $\I\in\M\setminus\M^i$, $\I\rdfsat\triple{p,\disjP,p}$;
    \item $\pmin(p,\CG)=\pmin(p,\CG^i)$; and
    \item for every $\dtriple{p,\subp,q}\in\Gdef_i$, $\R^*_i\rdfsat \dtriple{p,\subp,q}$. 
\end{itemize}

\nd Therefore, $\R^*_i$ is a model of $G^i$.

As next, we have to prove that $\R^*_i$ is in fact the \emph{minimal} model of $G^i$. To do so, we proceed by contradiction, by assuming that this is not the case. Then there is a model $\R'=(\M,r')$ of $G^i$ s.t. for every $\I\in\M$, $r'(\I)\leq r^*(\I)$, and there is an $\I'\in\M$ s.t. $r'(\I')< r^*(\I')$. Note that, since $r^*(\I)=0$ for every $\I\in\M\setminus\M^i$, $\I'\in\M^i$ necessarily. We have to prove that such an $\R'$ cannot exist.

Given $\CG=(\M,r)$ and $\R'=(\M,r')$, we build a ranked interpretation $\R^+=(\M,r^+)$ defining $r^+$ in the following way:
\[
    r^+(\I)=\left\{
    \begin{array}{ll}       
        r'(\I)+i &  \text{if } \I\in \M^i\\
         r(\I)  & \text{otherwise.}
    \end{array}\right.
\]

\nd That is,  

\begin{itemize}
    \item $r^+(\I)=r(\I)$ for every $\I\in\M\setminus\M^i$;
    \item from the definitions of $r^i$, $r'$ and $r^+$ we can conclude that $r^+(\I)\leq r(\I)$ for every $\I\in\M^i$, and there is an $\I'\in\M^i$ s.t. $r^+(\I')< r(\I')$.
\end{itemize}

\nd As a consequence, $\R^+\prec\CG$. Also, $\R^+$ is a model of $G$, because

\begin{itemize}
    \item all the \rhodfbot-interpretations $\I\in\M$ are models of $\Gclass$, hence $\R^+\rdfsat\Gclass$;
    \item for all the defeasible triples $\dtriple{p,\subc,q}\in\Gdef\setminus\Gdef_i$, $\cmin(p,\R^+)=\cmin(p,\CG)$, and, since $\CG\rdfsat\dtriple{p,\subc,q}$, we have $\R^+\rdfsat\dtriple{p,\subc,q}$;
    \item analogously, for all the defeasible triples $\dtriple{p,\subp,q}\in\Gdef\setminus\Gdef_i$, $\pmin(p,\R^+)=\pmin(p,\CG)$, and, since $\CG\rdfsat\dtriple{p,\subp,q}$, we have $\R^+\rdfsat\dtriple{p,\subp,q}$;
    \item for all the defeasible triples $\dtriple{p,\subc,q}\in\Gdef_i$, $\cmin(p,\R^+)=\cmin(p,\R')$, and, since $\R'\rdfsat\dtriple{p,\subc,q}$, we have $\R^+\rdfsat\dtriple{p,\subc,q}$;
    \item analogously, for all the defeasible triples $\dtriple{p,\subp,q}\in\Gdef_i$, $\pmin(p,\R^+)=\pmin(p,\R')$, and, since $\R'\rdfsat\dtriple{p,\subp,q}$, we have $\R^+\rdfsat\dtriple{p,\subp,q}$.
\end{itemize}

Therefore, $\R^+$ is a model of $G$, which is impossible, as $\CG$ is the minimal model of $G$ and, thus, $\R^*_i$ is the minimal model of the subgraph $G^i$.
\end{proof}

\vspace{0.4cm}
\Lemmarenkexceptionality*

\vspace{0.2cm}
\begin{proof}
 Let $i\leq n$, with $\R^*_i=(\M,r^*)$ and $\CG^i=(\M^i,r^i)$ be the models of  $G^i$ built as described above.

For every $\I\in\M_i$, $r^i(\I)=r^*_i(\I)$. If $\I\in\M\setminus\M^i$ we have seen above that $r^*(\I)=0$ and $\I\rdfsat\triple{p,\disjC,p}$ for every term $p$ s.t. $\hc(p)\geq i$.

Given these facts,  the following statements are equivalent:

\begin{itemize}
    \item $p$ is not \ssc-exceptional w.r.t.~$\CG^i$;
    \item there is an $\I\in\M^i$ s.t. $r^i(\I)=0$ and $\I\not\rdfsat\triple{p,\disjC,p}$;
    \item there is an $\I\in\M^i$ s.t. $r^*_i(\I)=0$ and $\I\not\rdfsat\triple{p,\disjC,p}$;
    \item $p$ is not \ssc-exceptional w.r.t.~$\R^*_i$.
\end{itemize}

\nd Analogously, we can prove that if we consider a term $p$ s.t. $\hp(p)\geq i$, the following statements are equivalent:

\begin{itemize}
    \item $p$ is not \ssp-exceptional w.r.t.~$\CG^i$;
    \item there is an $\I\in\M^i$ s.t. $r^i(\I)=0$ and $\I\not\rdfsat\triple{p,\disjP,p}$;
    \item there is an $\I\in\M^i$ s.t. $r^*_i(\I)=0$ and $\I\not\rdfsat\triple{p,\disjP,p}$;
    \item $p$ is not \ssp-exceptional w.r.t.~$\R^*_i$,
\end{itemize}

\nd which concludes the proof.
\end{proof}

\vspace{0.4cm}
\Lemmarankexcept*

\vspace{0.2cm}
\begin{proof}
Let $\dtriple{p,\subc,q}\in\Gdef$. 

For $i< n$, the following statements are equivalent:
\begin{itemize}
    \item $\hc(p)\geq i+1$;
    \item for all  $\I\in\M\setminus\M^{i+1}$, $\I\rdfsat\triple{p,\disjC,p}$;
    \item for all $\I\in\M\setminus\M^{i}$, $\I\rdfsat\triple{p,\disjC,p}$ and $p$ is \ssc-exceptional \wrt~$\CG^i$;
    \item for all $\I\in\M\setminus\M^{i}$, $\I\rdfsat\triple{p,\disjC,p}$ and $p$ is \ssc-exceptional \wrt~$R^*_i$ (by Lemma \ref{Lemma_renk_exceptionality});
    \item  $p$ is \ssc-exceptional \wrt~$G^i$ (by Lemma \ref{Lemma_minimal_model_CGi});
    \item  $\dtriple{p,\subc,q}\in \mathtt{ExceptionalC}(G^i)$ (by Corollary \ref{coroll:exceptionality_funct}).
\end{itemize}

For $i= n$, the following statements are equivalent:
\begin{itemize}
    \item $\hc(p)= \infty$;
    \item for all the $\I\in\M\setminus\M^{\infty}$, $\I\rdfsat\triple{p,\disjC,p}$;
    \item for all the $\I\in\M\setminus\M^{n}$, $\I\rdfsat\triple{p,\disjC,p}$ and $p$ is \ssc-exceptional \wrt~$\CG^n$;
    \item for all the $\I\in\M\setminus\M^{n}$, $\I\rdfsat\triple{p,\disjC,p}$ and $p$ is \ssc-exceptional \wrt~$R^*_i$ (by Lemma \ref{Lemma_renk_exceptionality});
    \item  $p$ is \ssc-exceptional \wrt~$G^n$ (by Lemma \ref{Lemma_minimal_model_CGi});
    \item  $\dtriple{p,\subc,q} \in \mathtt{ExceptionalC}(G^n)$ (by Corollary \ref{coroll:exceptionality_funct}).
\end{itemize}


\nd This proves the first half of the proposition. 

For triples $\dtriple{p,\subp,q} \in \Gdef$ the proof is analogous, which concludes.
\end{proof}

\vspace{0.4cm}
\Lemmaexceptranking*

\vspace{0.2cm}
\begin{proof}
We prove it by induction on the value of $i$. If $i=0$, then $\Gdef_0=\Gdef=\drnk_0$, that is, $G^0=G^\D_0$. Now, assume that $G^i=G^\D_i$ holds for all $i<n$, which implies also $\Gdef_i=\drnk_i$. By Lemma \ref{Lemma_rank_except}, for every $\dtriple{p,\subc,q}\in\Gdef_i$, $\dtriple{p,\subc,q}\in\Gdef_{i+1}$ iff  $\dtriple{p,\subc,q} \in \mathtt{ExceptionalC}(G^i)$ iff $\dtriple{p,\subc,q} \in \mathtt{ExceptionalC}(G^\D_i)$, that in turn is equivalent to $\dtriple{p,\subc,q}\in\drnk_{i+1}$. 

Analogously, by Lemma \ref{Lemma_rank_except}, we have that for every $\dtriple{p,\subp,q}\in\Gdef_i$, $\dtriple{p,\subp,q}\in\Gdef_{i+1}$ iff  $\dtriple{p,\subp,q}\in \mathtt{ExceptionalP}(G^i)$ iff $\dtriple{p,\subp,q} \in \mathtt{ExceptionalP}(G^\D_i)$, that in turn is equivalent to $\dtriple{p,\subp,q}\in\drnk_{i+1}$, which concludes.
\end{proof}

\vspace{0.4cm}
\propdecisionclass*
\vspace{0.2cm}

\begin{proof}
\begin{description}
\item[$\Rightarrow)$] From the construction of $G=\Gclass\cup\Gdef$ and Proposition \ref{Prop_ranking_main}, it is obvious that if $\I\in\MN$ then $\I\rdfsat \Gclass\cup\{\triple{p,\disjC,p}\mid \dtriple{p,\subc,q}\in \drnk_\infty\}\cup\{\triple{p,\disjP,p}\mid \dtriple{p,\subp,q}\in \drnk_\infty\}$.

\item[$\Leftarrow)$] We proceed by contradiction. Assume there is a \rhodfbot-interpretation $\I'\in\M$ s.t. $r(\I')=\infty$ and $\I'\rdfsat \Gclass\cup\{\triple{p,\disjC,p}\mid \dtriple{p,\subc,q}\in \drnk_\infty\}\cup\{\triple{p,\disjP,p}\mid \dtriple{p,\subp,q}\in \drnk_\infty\}$.

Assume  $h(\CG)=n$ (see Definition \ref{Def_concept_predicate_height}), and let $\R'=(\M,r')$ be a ranked interpretation where $r'$ is defined in the following way:
\[
    r'(\I)=\left\{\begin{array}{ll}
        r(\I) & \text{if }\I\in \MN\\
         n+1 &  \text{if } \I=\I'\\
         \infty  & \text{otherwise.}
    \end{array}\right.
\]

\nd Informally, $\R'$ has been obtained from $\CG$ simply by moving $\I'$ from rank $\infty$ to the top of $\MN$. Clearly $\R'\prec\CG$, since for every $\I\in\M$, $r'(\I)\leq r(\I)$, and $r'(\I')< r(\I')$.

It is easy to check that $\R'$ is a model of $G$: every model with a finite rank satisfies $\Gclass$; for every $\dtriple{p,\subc,q}\in\drnk_i$, for $i<\infty$, $\cmin(p,\R')=\cmin(p,\CG)$, and consequently $\R'\rdfsat\dtriple{p,\subc,q}$. Analogously for every $\dtriple{p,\subp,q}\in\drnk_i$, for $i<\infty$. For every $\dtriple{p,\subc,q}\in\drnk_\infty$, every model with a finite rank satisfies $\triple{p,\disjP,p}$; analogously for every $\dtriple{p,\subc,q}\in\drnkp_\infty$.
Hence $\R'$ is a model of $G$ and $\R'\prec\CG$, against the assumption that $\CG$ is the minimal model of $G$. 

As a  consequently for every $\I$, $\I\rdfsat \Gclass\cup\{\triple{p,\disjC,p}\mid \dtriple{p,\subc,q}\in \drnk_\infty\}\cup\{\triple{p,\disjP,p}\mid \dtriple{p,\subp,q}\in \drnk_\infty\}$ implies  $\I\in\MN$.
\end{description}
\end{proof}

\vspace{0.4cm}
\lemmatreescdisjC*
\vspace{0.2cm}

\begin{proof}
The proof proceeds by induction on the depth of the proof tree $T$.

\begin{description}
    \item[Case $d(T)=0$.] In this case, $H=\{\triple{p,\subc,q}\}$. Consequently $\I\rdfsat\triple{p,\disjC,p}$ obviously implies $\I\rdfsat\triple{p,\disjC,p}$.
    
    \item[Case $d(T)=1$.] The only way of deriving a triple of form $\triple{p,\subc, q}$ is using the rule (3a), as shown in the proof of Lemma \ref{lemma:triples_proof_sc}. Hence the proof tree consist of an instantiation of rule (3a) with $\triple{p, \subclass, q}$ as root.

\begin{center}
        \begin{tabular}{ll}
      $(3a)$ & $\frac{\triple{p, \subclass, t},  \triple{t, \subclass, q}}{\triple{p, \subclass, q}}$  
    \end{tabular}
\end{center}   

\nd Now, assume that $\I\rdfsat\triple{A,\disjC,A}$ holds for the antecedent of at least one of the two premises, \ie~$\I\rdfsat\triple{p,\disjC,p}$ or $\I\rdfsat\triple{t,\disjC,t}$ holds. We have to prove that $\I\rdfsat\triple{p,\disjC,p}$.

\begin{description}
    \item[Case (1).] $\I\rdfsat\triple{p,\disjC,p}$. The result is immediate.
    \item[Case (2).] $\I\rdfsat\triple{t,\disjC,t}$. Then together with $\I\rdfsat\triple{p,\subc,t}$ we derive $\I\rdfsat\triple{p,\disjC,p}$, by the soundness of rule (5b).
 \end{description}

    \item[Case $d(T)=n+1$.] Assume that the proposition holds for all the proof trees with depth $m\leq n$, with $n>1$. Let us show that  it holds also for  trees of depth $n+1$. So, let $T$ be a proof tree from $H$ to $\triple{p,\subclass, q}$ with depth $n+1$. The last step in $T$ must correspond to an instantiation of the rule (3a) with $\triple{p, \subclass, q}$ as root:
 
\begin{center}
        \begin{tabular}{ll}
      $(3a)$ & $\frac{\triple{p, \subclass, t},  \triple{t, \subclass, q}}{\triple{p, \subclass, q}}$  
    \end{tabular}
\end{center}   

Hence $T$ has two immediate subtrees: $T'$, having $\triple{p, \subclass, t}$ as root, and $T''$, having $\triple{t, \subclass, q}$ as root; each of them has a depth of at most $n$. Since we assume that the proposition holds for trees of depth at most $n$, if  $\I\rdfsat\triple{s,
\disjC,s}$ for some $\triple{s,\subc, o}\in H$, then by induction hypothesis either $\I\rdfsat\triple{p,\disjC,p}$ (if $\triple{s,\subc, o}$ appears as a leaf in $T'$) or $\I\rdfsat\triple{t,\disjC,t}$ (if $\triple{s,\subc, o}$ appears as a leaf in $T''$). Likewise case $d(T)=1$, in both the cases we can conclude $\I\rdfsat\triple{p,\disjC,p}$, which concludes the proof.
\end{description}
\end{proof}

\vspace{0.4cm}
\lemmaalgminentCcorrect*
\vspace{0.2cm}

\begin{proof}
Consider $G=\Gclass\cup\Gdef$ and  let $\CG=(\M,r)$ be its minimal model. Let $\hc(p)=k$, with $k\leq n$. From Corollary \ref{Coroll_ranking_main} we can conclude that in the the  $\mathtt{DefMinEntailmentC}$ procedure, $G'\not\deriv\triple{p,\disjC,p}$ for $j=k$, and $\drnkc^p = \{\dtriple{r,\subc, s}\mid \dtriple{r,\subc, s}\in \drnk_k\setminus \drnk_{k+1}\}$.

We have to prove that $G\minentail \dtriple{p,\subc,q}$ iff $\Gclass\cup(\drnkc^p)^s\deriv \triple{p,\subc,q}$.

\begin{description}
\item[$\Leftarrow .)$]   Let us show that $\Gclass\cup(\drnkc^p)^s\deriv \triple{p,\subc,q}$ implies $G\minentail \dtriple{p,\subc,q}$.
Assume  $\Gclass\cup(\drnkc^p)^s\deriv \triple{p,\subc,q}$. In order to prove  $G\minentail \dtriple{p,\subc,q}$ we need to show that for every $\I\in\cmin(p,\CG)$, $\I\rdfsat \triple{p,\subc,q}$ holds.

We proceed by contradiction. So, assume that $\I \in \cmin(p,\CG)$ and $\I\not\rdfsat\triple{p,\subc,q}$. $\I\in\cmin(p,\CG)$ implies  $\I\not\rdfsat\triple{p,\disjC,p}$. Also, since $\hc(p)=k$, $r(\I)= k$ follows. 
Since $\Gclass\cup(\drnkc^p)^s\deriv \triple{p,\subc,q}$, $\I\not\rdfsat\triple{p,\disjC,p}$ and $\I\rdfsat\Gclass$ (Lemma \ref{prop_decision_class}), then  there is at least a triple in $(\drnkc^p)^s$ that is not satisfied by $\I$. By Lemma \ref{lemma:triples_proof_sc} such a triple must be of the kind $\triple{s,\subc,t}$. In particular, $\Gclass\cup(\drnkc^p)^s\deriv \triple{p,\subc,q}$ implies that there must be a graph $H\subseteq\Gclass\cup(\drnkc^p)^s$ s.t. there is a proof tree $T$ proving $\triple{p,\subc,q}$ from $H$, and there is some triple $\triple{s,\subc,t}\in H\cap (\drnkc^p)^s$ s.t. $\I\not\rdfsat\triple{s,\subc,t}$.
As $\dtriple{s,\subc,t}\in\drnk_k\setminus \drnk_{k+1}$, by Proposition \ref{Prop_ranking_main} we have $\hc(s)=k$. $\hc(s)=k$ and $r(\I)= k$ imply that  if $\I\not\rdfsat \triple{s,\subc,t}$, then $\I\rdfsat \triple{s,\disjC,s}$ ($\I\notin\cmin(s,\CG)$), otherwise $\CG$ would not be a model of $G$.
However, by Lemma \ref{lemma_tree_sc_disjC}, $\I\rdfsat \triple{s,\disjC,s}$ implies $\I\rdfsat \triple{p,\disjC,p}$, against the hypothesis that $\I\in\cmin(p,\CG)$. Therefore, we must conclude that $\I\rdfsat\triple{p,\subc,q}$.

\item[$\Rightarrow .)$]  Let us show that $G\minentail \dtriple{p,\subc,q}$  implies $\Gclass\cup(\drnkc^p)^s\deriv \triple{p,\subc,q}$.
Let $\hc(p)=k$, with $k\leq n$. $G\minentail \dtriple{p,\subc,q}$ means that, given the minimal model $\CG=(\M,r)$,  for $\I\in\M$ s.t. $r(\I)=k$ and $\I\not\rdfsat\triple{p,\disjC,p}$, $\I\in \cmin(p,\CG)$ and $\I \rdfsat\triple{p,\subc,q}$.

Now, consider the  graph $G^*\assign \Gclass\cup(\drnk_k\setminus\drnk_{k+1})^s\cup\{\triple{r\disjC r}\mid \dtriple{r,\subc,t}\in \drnk_{k+1}\}\cup \{\triple{r\disjP r}\mid \dtriple{r,\subp,t}\in \drnk_{k+1}\}$. That is, $G^*$ contains all the strict triples in $G$, $\Gclass$, all the defeasible triples that are satisfied at height $k$, that is, $\drnk_k\setminus\drnk_{k+1}$, and  for all the triples that are exceptional in $k$, the set $\{\triple{r\disjC r}\mid \dtriple{r,\subc,t}\in \drnk_{k+1}\}\cup \{\triple{r\disjP r}\mid \dtriple{r,\subp,t}\in \drnk_{k+1}\}$.

Let $\I_{G^*}$ be the characteristic \rhodfbot~model of $G^*$, built as defined in Lemma \ref{completefirst}. We know that $\I_{G^*}$ satisfies exactly $\clos(G^*)$. Also, $\I_{G^*}\in\M$, as it can be checked by the definition in Section \ref{sect_semantics} of the ranked interpretations in $\RG$.

Since $\I_{G^*}$ satisfies $\I\rdfsat \Gclass\cup\{\triple{r,\disjC,r}\mid \dtriple{r,\subc,t}\in \drnk_\infty\}\cup\{\triple{r,\disjP,r}\mid \dtriple{r,\subp,t}\in \drnk_\infty\}$, by 
Lemma~ \ref{prop_decision_class} it must be in $\MN$. Moreover, it must hold that $r(\I_{G^*})=k$, since:

\begin{itemize}
    \item for every triple $\dtriple{s,\subc,t}\in\drnk_k\setminus\drnk_{k+1}$  (resp., $\dtriple{s,\subp,t}\in \drnk_k\setminus\drnk_{k+1}$) $\I_{G^*}$ does not satisfy $\triple{s,\disjC,s}$ (resp., $\triple{s,\disjP, s}$), while $\hc(s)=k$ (resp., $\hp(s)=k$). Hence it cannot be $r(\I_{G^*})<k$.
    \item $\I_{G^*}$ is compatible with $r(\I_{G^*})=k$, since it satisfies all the triples in $\drnk_k\setminus\drnk_{k+1}$ and for every $\dtriple{s,\subc,t}\in\drnk_{k+1}$, $\I_{G^*}\rdfsat\triple{s,\disjC,s}$ (resp., for every $\dtriple{s,\subp,t}\in\drnk_{k+1}$, $\I_{G^*}\rdfsat\triple{s,\disjP,s}$).
    \item $\CG$ assigns the minimal rank to every \rhodfbot interpretation, and since $\I_{G^*}$ is compatible with $r(\I_{G^*})=k$ and not with $r(\I_{G^*})<k$, we can conclude $r(\I_{G^*})=k$.
\end{itemize}

$\hc(p)=k$ implies that $G^\D_k\not\rdfent\triple{p,\disjC,p}$ (Corollary \ref{Coroll_ranking_main}). Since $G^\D_k\rdfent G^*$, $G^*\not\rdfent\triple{p,\disjC,p}$. 

In summary, we have the following situation:
\begin{itemize}
    \item $r(\I_{G^*})=k$;
    \item $\hc(p)=k$;
    \item since $G^*\not\rdfent\triple{p,\disjC,p}$ and $\I_{G^*}$ is the characteristic model of $G^*$, $\I_{G^*}\not\rdfsat\triple{p,\disjC,p}$.
\end{itemize}

Hence $\I_{G^*}\in\cmin(p)$. This, together with $G\minentail \dtriple{p,\subc,q}$, implies $\I_{G^*}\rdfsat\triple{p,\subc,q}$. 
Being $\I_{G^*}$ the characteristic model of $G^*$, $\I_{G^*}$ satisfies a triple $\triple{s,\subc,t}$ iff $\triple{s,\subc,t}\in\clos(G^*)$. Hence $\I_{G^*}\rdfsat\triple{p,\subc,q}$ implies $G^*\deriv\triple{p,\subc,q}$.

By Lemma \ref{lemma:triples_proof_sc}, we know that to derive a triple of form $\triple{A,\subc,B}$ from a graph it is sufficient to consider only the triples in the same graph with the same form, and in $G^*$ all such triples are in $\Gclass\cup(\drnk_k\setminus\drnk_{k+1})^s$, that is, $\Gclass\cup(\drnkc^p)^s$. Hence $G^*\deriv\triple{p,\subc,q}$ implies 

\[
\Gclass\cup(\drnkc^p)^s\deriv \triple{p,\subc,q} \ ,
\]

\nd which concludes.

 \end{description}
\end{proof}

\vspace{0.4cm}
\theoralgminentCcorrect*

\vspace{0.2cm}
\begin{proof}
Let $\CG$ be the minimal model of $G$, with $h(\CG)=n$. Given $\dtriple{p,\subc,q}$ we have two possible cases:
\begin{description}
    \item[Case $\hc(p)\leq n$.] The result is guaranteed by Lemma \ref{lemma_alg_minentC_correct}.
    \item[Case $\hc(p)=\infty$.] By Definition \ref{defsat}, $\CG\rdfsat\dtriple{p,\subc,q}$. At the same time, if $\hc(p)=\infty$, then by definition of the procedure, 
    $\mathtt{DefMinEntailmentC}(G, \rnk(G), \dtriple{p,\subc,q})$ must be the case.
\end{description}
\end{proof}

\vspace{0.4cm}
\propllerw*
\vspace{0.2cm}

\begin{proof}
We prove $(LLE_c)$ and $(RW_c)$. The proofs for  $(LLE_p)$ and $(RW_i)$ are analogous.

\begin{itemize}
    \item[$(LLE_c)$.]  Let $G\minentail\triple{p,\subc,q}$, $G\minentail\triple{q,\subc,p}$, and $G\minentail\dtriple{p,\subc, r}$. $G\minentail\triple{p,\subc,q}$ and $G\minentail\triple{q,\subc,p}$ imply that  every $\rhodfbot$-intepretation in $\CG$ satisfies $\triple{p,\subc,q}$ and $\triple{q,\subc,p}$. Since the semantics is sound \wrt~the derivation rules, by rule $(5b)$ we have that for every $\rhodfbot$-intepretation $\I$ in $\CG$, $\I\rdfsat \triple{p,\disjC,p}$ iff $\I\rdfsat \triple{q,\disjC,q}$. Consequently $\cmin(p,\CG)=\cmin(q,\CG)$. 
    
    Let $\I$ be in $\cmin(p,\CG)$. Since $G\minentail\dtriple{p,\subc, r}$, $\I\rdfsat\triple{p,\subc, r}$. We also have $\I\rdfsat\triple{q,\subc,p}$, and, being $\I$ sound w.r.t. rule $(3a)$, we conclude $\I\rdfsat\triple{q,\subc, r}$. $\cmin(p,\CG)=\cmin(q,\CG)$ implies  $\CG\rdfsat\dtriple{q,\subc, r}$, that is, $G\minentail\dtriple{q,\subc,r}$.
    
    \item[$(RW_c)$.] $G\minentail\triple{q,\subc, r}$ implies  that  every $\rhodfbot$-intepretation in $\CG$ satisfies $\triple{q,\subc,r}$. Let $\I$ be in $\cmin(p,\CG)$. Since $G\minentail\dtriple{p,\subc, q}$, $\I\rdfsat\triple{p,\subc, q}$. Hence   $\I\rdfsat\triple{p,\subc, r}$, that implies $\CG\rdfsat\dtriple{p,\subc, r}$, that is, $G\minentail\dtriple{p,\subc,r}$.
\end{itemize}
\end{proof}

\section{Proofs for Section \ref{din}}

\vspace{0.4cm}
\propidempinh*

\vspace{0.2cm}
\begin{proof}
To prove Proposition \ref{prop_idemp_inh} it is sufficient to consider the following example, in which we apply the operator $\closinh$ to a graph $G$, and then we show that applying the inheritance completion to $\closinh(G)$ we obtain new defeasible triples, so $(\closinh(G))_{in}\neq\closinh(G)$, which implies the proposition.


So, consider the following graph $G=\Gclass\cup\Gdef$ with
\begin{eqnarray*}
\Gclass & =& \{\triple{a,\subc,b},\triple{c,\subc,d},\triple{c,\disjC,g},\triple{l,\disjC,m},\triple{e,\disjC,i},\triple{n,\disjC,o}\} \\
\Gdef & =& \{\dtriple{a,\subc,f},\dtriple{f,\subc,g},\dtriple{b,\subc,c},\dtriple{b,\subc,l},\dtriple{c,\subc,m},\dtriple{c,\subc,h},\dtriple{h,\subc,i},\dtriple{d,\subc,e},\\
&& \dtriple{d,\subc,n},\dtriple{e,\subc,o} \ \} \ .
\end{eqnarray*}
\nd The graph is represented in Figure~\ref{figidemp1}. In order for $\Gclass$ to be closed under $\mathtt{StrictMinEntailment}$ it is sufficient to add the symmetric version of the triples $\triple{A,\disjC,B}$, obtained through the application of rule (5a).

At first, we proceed to determine $\closinh(G)$. As a first step we determine $G_{in}$ by applying the procedure $\mathtt{InheritanceCompletion}(G)$, and obtaining the following completion:
\[
G_{in}=G\cup\{\dtriple{a,\subc,l},\dtriple{b,\subc,h},\dtriple{b,\subc,d},\dtriple{b,\subc,n},\dtriple{c,\subc,n}\} \ .
\]
\nd The new triples in $G_{in}$ are represented in orange in Figure~\ref{figidemp2}. Please note that $\dtriple{a,\subclass,e}$ is not in $G_{in}$: $\Delta_{a,e}=\{\dtriple{a,\subc,f},\dtriple{f,\subc,g},\dtriple{b,\subc,c},\dtriple{c,\subc,h},\dtriple{h,\subc,i},\dtriple{d,\subc,e} \}$, and $\Gclass\cup\Delta_{a,e}\not\minentail \dtriple{a,\subclass,e}$.

Determining $\closinh(G)$ corresponds  to determine $\closmin(G_{in})$. The latter, represented in Figure \ref{figidemp3}, is
\[
\closinh(G)=G_{in}\cup\{\dtriple{a,\subc,c},\dtriple{c,\subc,e}\} \ .
\]
\nd $\dtriple{a,\subclass,e}$ is still not derivable, since $G_{in}\not\minentail \dtriple{a,\subclass,e}$.

Now, if $\closinh$ satisfies Idempotence, we would have $\closinh(\closinh(G))=\closinh(G)$, that implies that the inheritance completion of $\closinh(G)$, namely $(\closinh(G))_{in}$, does not add any new triples to $(\closinh(G))$, \ie~$(\closinh(G))_{in}=\closinh(G)$. However, once we determe $(\closinh(G))_{in}$, we obtain a new defeasible triple, as indicated in Figure \ref{figidemp4}:
\[
(\closinh(G))_{in}=\closinh(G)\cup\{\dtriple{a,\subc,e}\} \ ,
\]
\nd which concludes.


\begin{figure}
\begin{center}
\includegraphics[scale=0.45]{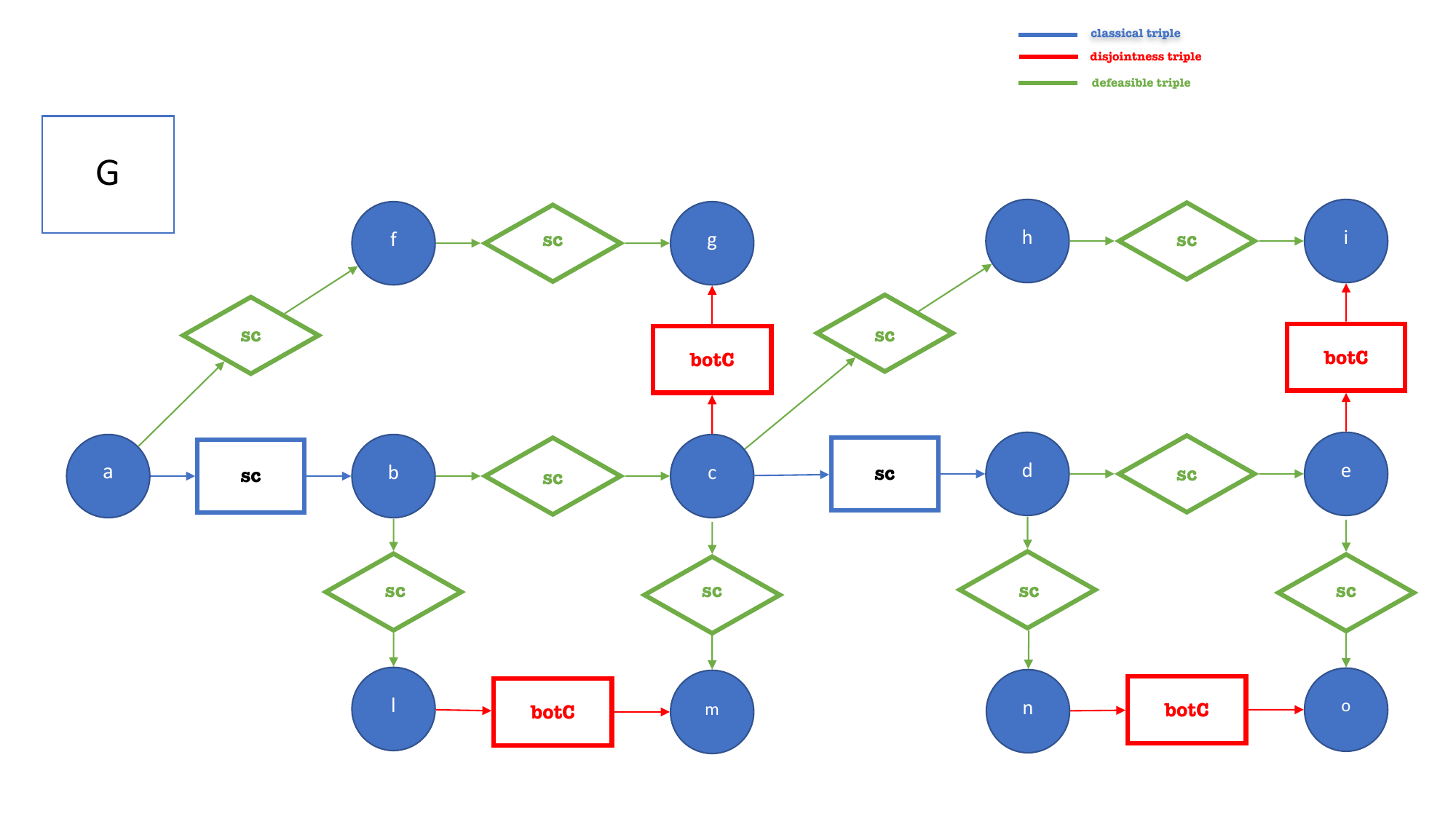}
\end{center}
\caption{The graph $G$. } \label{figidemp1}
\end{figure}

\begin{figure}
\begin{center}
\includegraphics[scale=0.45]{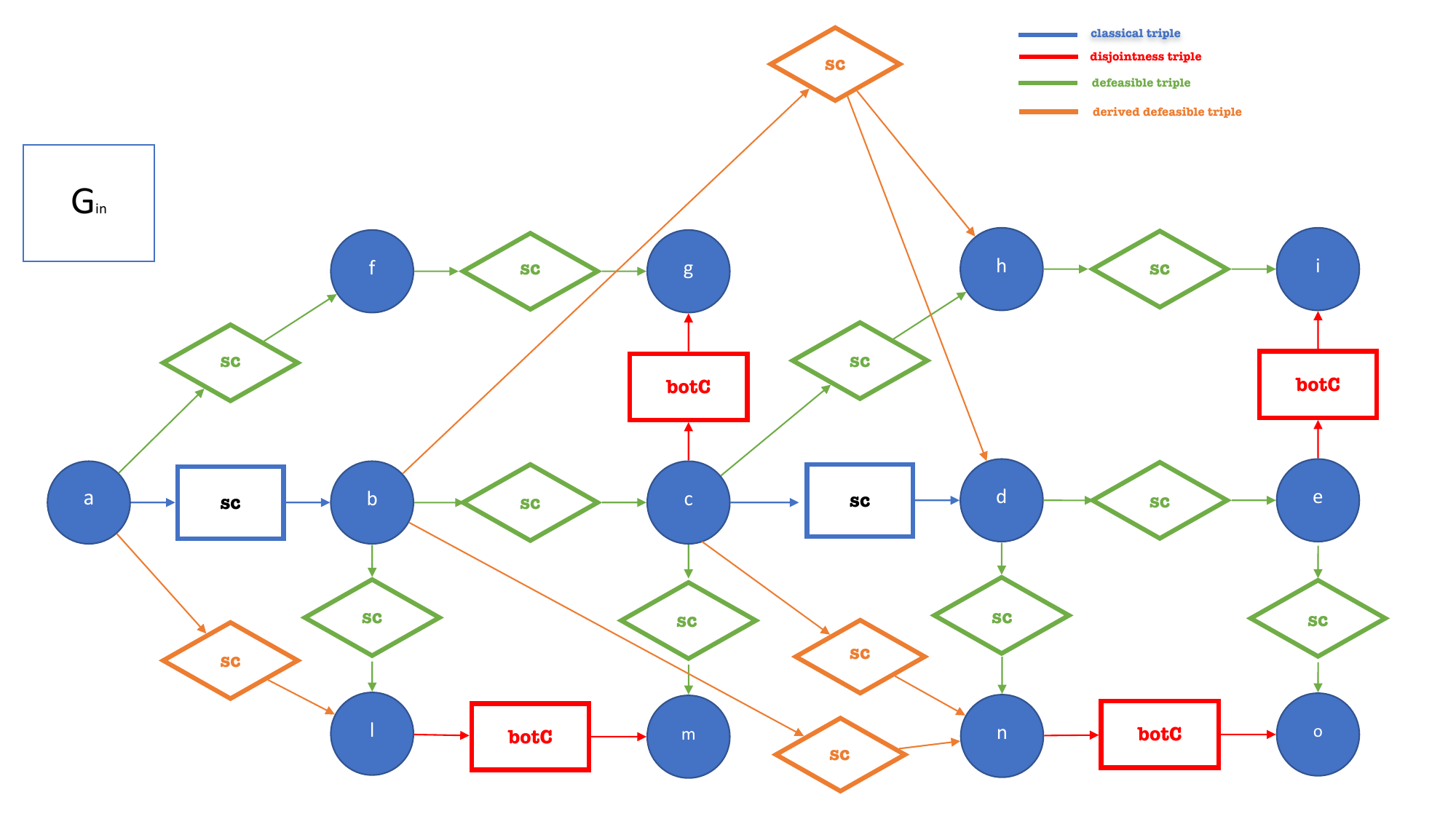}
\end{center}
\caption{The inheritance completion of the graph $G$. } \label{figidemp2}
\end{figure}

\begin{figure}
\begin{center}
\includegraphics[scale=0.45]{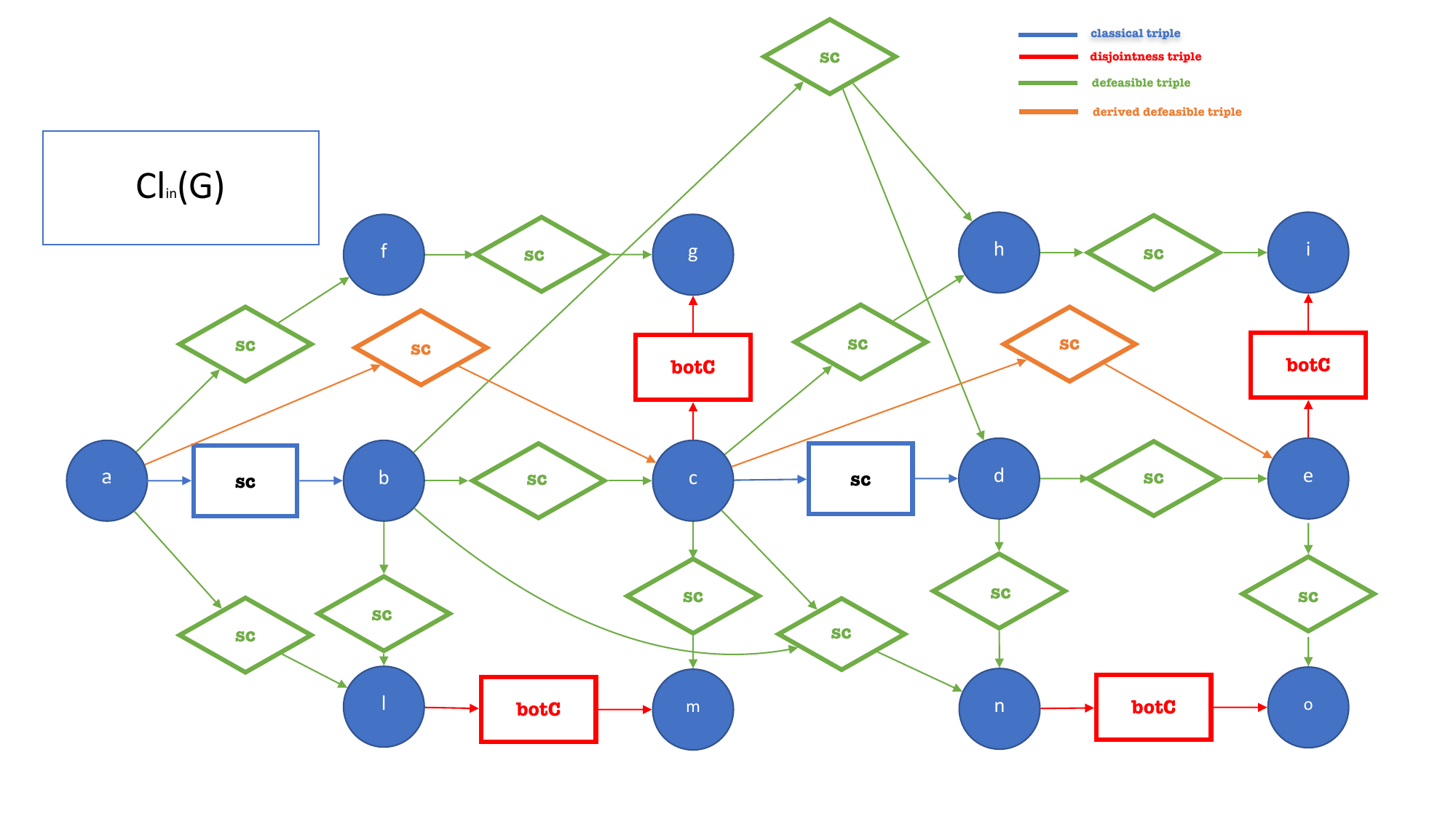}
\end{center}
\caption{The inheritance based closure of the graph $G$. } \label{figidemp3}
\end{figure}

\begin{figure}
\begin{center}
\includegraphics[scale=0.45]{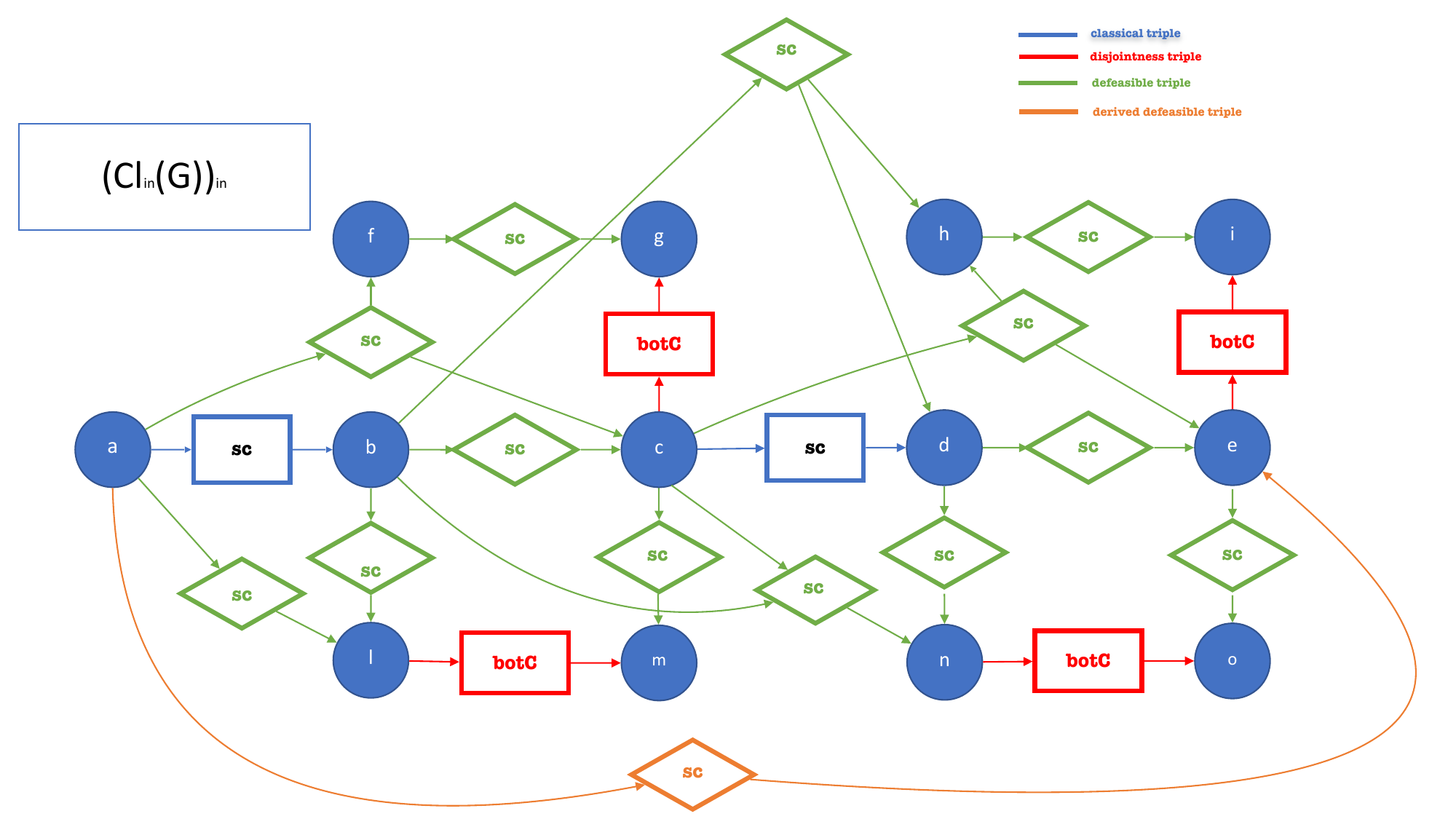}
\end{center}
\caption{An extra inheritance completion of the graph $\closinh(G)$. } \label{figidemp4}
\end{figure}


\end{proof}

\vspace{0.4cm}
\lemmacomplexclassclosureDelta*

\vspace{0.2cm}
\begin{proof}
In order to determine the set $\Delta_{p,q}$, given any two nodes $p,q$ in a graph $G$, we adapt the analogous method presented for ducts in \cite[Section 3.1.6]{CasiniStraccia13} to our case.

Specifically, given that the construction of paths is independent \wrt~the  kind of triples that are involved, we can analyse the problem using simple directed graphs. Given a graph $G$, it is sufficient to define the correspondent directed graph $\mathcal{G}=\tuple{V,E}$ in the following way:
    \begin{itemize}
    \item $V$ is the set of terms that appear in the first or third position in the triples in $G$;
    \item $E$ is a set of directed links $\tuple{s,t}$, with $s,t\in V$, s.t. $\tuple{s,t}\in E$ iff $\anytriple{s,r,t}\in G$, for any $r$.
    \end{itemize}
\nd Once we have defined $\mathcal{G}$, let us recall a well-known result in graph theory saying that in a directed graph, given two nodes $p$ and $q$, determining if there is a path from $p$ to $q$ can be determined in time $\mathcal{O}(|V|+|E|)$, \eg~using  BFS (Breadth First Search)~\cite{CormenEtAl01}. Now, the following argument shows that indeed $\Delta_{p,q}$ can be determined in polynomial time.

At  first, we check if there is a path between $p$ and $q$. If not, then $\Delta_{p,q} = \emptyset$. Otherwise, we call the procedure {\bf Delta($p$)} below:
\begin{description}
\item[Delta($p$):] for each outgoing edge $\tuple{p,x}$ of $p$, such that both $\tuple{p,x}$ and $x$ are not marked, do:
 if there is a path between $x$ and $q$ then mark both $\tuple{p,x}$ and $x$, and recursively, call {\bf Delta($x$)}.
\end{description}
\nd Once finished,  $\Delta_{p,q}$ can immediately be built from the marked edges. 

Note that each edge is marked once and each node is marked (\ie, explored) once and, thus, the algorithm is bounded polynomially by the size of the graph $\mathcal{G}$: the procedure Delta($p$) needs to be called at most $|E|$ times, hence the entire construction of $\Delta_{p,q}$ requires $\mathcal{O}(|E|(|V|+|E|))$.
 Since $|V|\leq 2|E|$ and $|E| \leq |G|$,  $|E|(|V|+|E|) \leq |G|(3|G|)=3|G|^2$. Hence the construction of $\Delta_{p,q}$ runs in time $\mathcal{O}(|G|^2)$.
\end{proof}

\vspace{0.4cm}
\propcomplexinhcompl*

\vspace{0.2cm}
\begin{proof}
First, we analyse the cost of computing the lines 3-8 in $\mathtt{InheritanceCompletion}(G)$. That is, given a graph $G$ and a pair $\tuple{p,q}$ of nodes in $G$, what is the cost of calculating whether $\dtriple{p,\subc,q}$ or $\dtriple{p,\subp,q}$ need to be added in $G_{in}$. We proceed first by calculating $\Delta_{p,q}$ ($\mathcal{O}(|G|^2)$, see Lemma \ref{lemma_complexclassclosureDelta}), followed by $\mathtt{DefMinEntailmentC}$ ($O(|\Gdef||G|^2)$, see Proposition \ref{complexminentC}) and $\mathtt{DefMinEntailmentC}$ ($O(|\Gdef||G|^2)$, see Proposition \ref{complexminentP}). Hence, combining these 3 procedures, these lines can be computed in time $O(|\Gdef||G|^2)$.

We have to repeat such lines for every pair $p,q\in G$. Therefore, $\mathtt{InheritanceCompletion}(G)$ runs in time  $O(|\Gdef||G|^4)$.
\end{proof}

\vspace{0.4cm}
\propcomplexinhcomplinher*

\vspace{0.2cm}
\begin{proof}
The claim follows from the facts that: \ii{i} by Proposition~\ref{complderiv}, the closure of
$\Gclass$ under procedure $\mathtt{StrictMinEntailment}$ can be computed in polynomial time; \ii{ii}
$\mathtt{InheritanceCompletion}(G)$ needs to be run only once for any graph $G$ and, thus, requires polynomial time (see Proposition~\ref{prop_complexinhcompl})); and \ii{iii} once we have obtained $G_{in}$ from $\mathtt{InheritanceCompletion}(G)$, following 
Definition~\ref{def_inh_ent}, deciding whether $G\vdash_{in}\anytriple{p,o,q}$ corresponds to deciding whether $G_{in}\minentail\anytriple{p,o,q}$, which is tractable (see Corollary \ref{complexminent}). Combining the three facts above, $G\vdash_{in}\anytriple{p,o,q}$ can be decided in polynomial time, which concludes.
\end{proof}





\end{document}